\documentclass[journal]{IEEEtran}

\usepackage{tikz}
\newcommand\copyrighttext{%
  \footnotesize This paper is a pre-print. IEEE copyright notice. ``\copyright~2023 IEEE. Personal use of this material is permitted. Permission from IEEE must be obtained for all other uses, in any current or future media, including reprinting/republishing this material for advertising or promotional purposes, creating new collective works, for resale or redistribution to servers or lists, or reuse of any copyrighted component of this work in other works.''}
\newcommand\copyrightnotice{%
\begin{tikzpicture}[remember picture,overlay]
\node[anchor=south,yshift=9pt] at (current page.south) {\fbox{\parbox{\dimexpr\textwidth-\fboxsep-\fboxrule\relax}{\copyrighttext}}};
\end{tikzpicture}
}

\usepackage{times}
\usepackage{epsfig}
\usepackage{graphicx}
\usepackage{amsmath}
\usepackage{amssymb}
\usepackage{array}
\usepackage{multirow}
\usepackage{subcaption}
\usepackage{color}
\usepackage{float}
\usepackage{graphicx}
\usepackage{mathtools}
\usepackage{booktabs}
\usepackage{adjustbox}
\usepackage{hyperref}
\usepackage{cite}
\newcases{Rcases}{$\mskip\thickmuskip$}{%
  \hfil$\m@th{##}$}{$\m@th{##}$\hfil}{.}{\rbrace}
\usepackage{flushend}

\usepackage[ruled,noresetcount,noend]{algorithm2e}
\usepackage{algpseudocode}
\usepackage[font=scriptsize]{caption} 

\makeatletter
\newcommand\HUGES{\@setfontsize\Huge{14}{25}}
\newcommand\HUGESS{\@setfontsize\Huge{24}{45}}
\newcommand\HUGESSS{\@setfontsize\Huge{15}{45}}
\newcommand\HUGESSSS{\@setfontsize\Huge{30}{90}}
\makeatother    

\definecolor{houston_data_Background!}{RGB}{0.0,0.0,127.5}
\definecolor{houston_data_Grass-healthy!}{RGB}{0.0,0.0,204.77}
\definecolor{houston_data_Grass-stressed!}{RGB}{0.0,8.5,255.0}
\definecolor{houston_data_Grass-synthetic!}{RGB}{0.0,76.5,255.0}
\definecolor{houston_data_Tree!}{RGB}{0.0,144.5,255.0}
\definecolor{houston_data_Soil!}{RGB}{0.0,212.5,255.0}
\definecolor{houston_data_Water!}{RGB}{41.13,255.0,205.65}
\definecolor{houston_data_Residential!}{RGB}{95.97,255.0,150.81}
\definecolor{houston_data_Commercial!}{RGB}{150.81,255.0,95.97}
\definecolor{houston_data_Road!}{RGB}{205.65,255.0,41.13}
\definecolor{houston_data_Highway!}{RGB}{255.0,229.81,0.0}
\definecolor{houston_data_Railway!}{RGB}{255.0,166.85,0.0}
\definecolor{houston_data_Parking-lot1!}{RGB}{255.0,103.89,0.0}
\definecolor{houston_data_Parking-lot2!}{RGB}{255.0,40.93,0.0}
\definecolor{houston_data_Tennis-court!}{RGB}{204.77,0.0,0.0}
\definecolor{houston_data_Running-track!}{RGB}{127.5,0.0,0.0}

\definecolor{trento_data_Background!}{RGB}{0.0,0.0,127.5}
\definecolor{trento_data_Apples!}{RGB}{0.0,0.0,204.77}
\definecolor{trento_data_Buildings!}{RGB}{0.0,8.5,255.0}
\definecolor{trento_data_Ground!}{RGB}{0.0,76.5,255.0}
\definecolor{trento_data_Woods!}{RGB}{0.0,144.5,255.0}
\definecolor{trento_data_Vineyard!}{RGB}{0.0,212.5,255.0}
\definecolor{trento_data_Roads!}{RGB}{41.13,255.0,205.65}

\definecolor{MUUFL_data_Background!}{RGB}{0.0,0.0,127.5}
\definecolor{MUUFL_data_Trees!}{RGB}{0.0,0.0,204.77}
\definecolor{MUUFL_data_Grass-Pure!}{RGB}{0.0,8.5,255.0}
\definecolor{MUUFL_data_Grass-Groundsurface!}{RGB}{0.0,76.5,255.0}
\definecolor{MUUFL_data_Dirt-And-Sand!}{RGB}{0.0,144.5,255.0}
\definecolor{MUUFL_data_Road-Materials!}{RGB}{0.0,212.5,255.0}
\definecolor{MUUFL_data_Water!}{RGB}{41.13,255.0,205.65}
\definecolor{MUUFL_data_Buildings'-Shadow!}{RGB}{95.97,255.0,150.81}
\definecolor{MUUFL_data_Buildings!}{RGB}{150.81,255.0,95.97}
\definecolor{MUUFL_data_Sidewalk!}{RGB}{205.65,255.0,41.13}
\definecolor{MUUFL_data_Yellow-Curb!}{RGB}{255.0,229.81,0.0}
\definecolor{MUUFL_data_ClothPanels!}{RGB}{255.0,166.85,0.0}

\definecolor{Augsburg_data_Background!}{RGB}{0.0,0.0,127.5}
\definecolor{Augsburg_data_Forest!}{RGB}{0.0,0.0,204.77}
\definecolor{Augsburg_data_Residential-Area!}{RGB}{0.0,8.5,255.0}
\definecolor{Augsburg_data_Industrial-Area!}{RGB}{0.0,76.5,255.0}
\definecolor{Augsburg_data_Low-Plants!}{RGB}{0.0,144.5,255.0}
\definecolor{Augsburg_data_Allotment!}{RGB}{0.0,212.5,255.0}
\definecolor{Augsburg_data_Commercial-Area!}{RGB}{41.13,255.0,205.65}
\definecolor{Augsburg_data_Water!}{RGB}{95.97,255.0,150.81}

\usepackage[normalem]{ulem}
\usepackage{xcolor}
\definecolor{darkgreen}{rgb}{0,.4,0}
\definecolor{darkcyan}{rgb}{0,.4,.4}
\newcommand{\REMOVE}[1]%
          {{\color{blue}\sout{#1}}}

\newcommand{\COMMENT}[1]%
          {{\color{darkgreen}\textbf{{Ankur:}} {#1}}}
          
\begin{document}
\title{Multimodal Fusion Transformer for Remote Sensing Image Classification}

\author{
        Swalpa Kumar Roy,~\IEEEmembership{Studen Member,~IEEE,}
        Ankur Deria,
        Danfeng Hong,~\IEEEmembership{Senior Member,~IEEE,} \\
        Behnood Rasti,~\IEEEmembership{Senior Member,~IEEE,}
        Antonio Plaza,~\IEEEmembership{Fellow,~IEEE,}
        and Jocelyn Chanussot,~\IEEEmembership{Fellow,~IEEE}
        
\thanks{This work was supported by the National Key Research and Development Program of China under Grant 2022YFB3903401 and also Science and Engineering Research Board, Government of India under Grant SRG/2022/001390.}

\thanks{S. K. Roy is with the Department of Computer Science and Engineering, Jalpaiguri Government Engineering College, Jalpaiguri 735102, West Bengal, India. (e-mail: swalpa@cse.jgec.ac.in)}
\thanks{A. Deria is with the Department of Informatics, Technical University of Munich, 85748 Garching bei München, Germany (e-mail: i.am.ankur.deria@tum.de).}
\thanks{D. Hong is with the Aerospace Information Research Institute, Chinese Academy of Sciences, 100094 Beijing, China. (e-mail: hongdf@aircas.ac.cn)}
\thanks{B. Rasti is with Helmholtz-Zentrum Dresden-Rossendorf, Helmholtz Institute Freiberg for Resource Technology, Freiberg 09599, Germany. (e-mail: b.rasti@hzdr.de)}
\thanks{A. Plaza is with the Hyperspectral Computing Laboratory, Department of Technology of Computers and Communications, Escuela Politécnica, University of Extremadura, 10003 Cáceres, Spain (e-mail: aplaza@unex.es).}
\thanks{J. Chanussot is with the Univ. Grenoble Alpes, CNRS, Grenoble INP, GIPSA-Lab, 38000 Grenoble, France (e-mail: jocelyn@hi.is)}
}
% The paper headers
\markboth{Published in IEEE Transactions on Geoscience and Remote Sensing (DOI: \url{10.1109/TGRS.2023.3286826})}
{Roy \MakeLowercase{\textit{et al.}}: }

\maketitle

\copyrightnotice

\begin{abstract}
\textcolor{black}{Vision transformers (ViTs) have been trending in image classification tasks due to their promising performance when compared to convolutional neural networks (CNNs). As a result, many researchers have tried to incorporate ViTs in hyperspectral image (HSI) classification tasks. To achieve satisfactory performance, close to that of CNNs, transformers need fewer parameters. ViTs and other similar transformers use an external classification (CLS) token which is randomly initialized and often fails to generalize well, whereas other sources of multimodal datasets, such as light detection and ranging (LiDAR) offer the potential to improve these models by means of a \texttt{CLS}. %\REMOVE{However, the use of a feature embedding derived from other sources of multimodal datasets, such as light detection and ranging (LiDAR), offers the potential to improve those models by means of a \texttt{CLS}. We also introduce a new attention mechanism. However, the introduction of a \texttt{CLS} token which is extracted from complementary information (e.g., LiDAR) in the attention mechanism \REPLACE{for}{leads to the} improvement in exchange of information between HSI tokens and the \texttt{CLS} \REMOVE{(e.g., LiDAR)} token.} 
In this paper, we introduce a new multimodal fusion transformer (MFT) network which comprises a multihead cross patch attention (\texttt{mCrossPA}) for HSI land-cover classification. Our \texttt{mCrossPA} utilizes other sources of complementary information in addition to the HSI in the transformer encoder to achieve better generalization. The concept of tokenization is used to generate \texttt{CLS} and HSI patch tokens, helping to learn a {distinctive representation} in a reduced and hierarchical feature space. Extensive experiments are carried out on {widely used benchmark} datasets {i.e.,} the University of Houston, Trento, University of Southern Mississippi Gulfpark (MUUFL), and Augsburg.  We compare the results of the proposed MFT model with other state-of-the-art transformers, classical CNNs, and conventional classifiers models. The superior performance achieved by the proposed model is due to the use of multihead cross patch attention. The source code will be made available publicly at \url{https://github.com/AnkurDeria/MFT}.}
\end{abstract}

\graphicspath{{figures/}}


\section{Introduction and Contributions}

\IEEEPARstart{P}{henomena} such as climate change or desertification has led to a drastic growth in the popularity of Earth Observation (EO) via remote sensing (RS). These tasks include (but are not limited to) land cover classification~\cite{ahmad2021hyperspectral, bartholome2005glc2000, roy2021revisiting}, forestry~\cite{koetz2008multi}, mineral exploration and mapping, object/target detection~\cite{wu2019fourier,wu2023uiu}, environmental monitoring~\cite{ustin2004manual}, urban planning~\cite{chen2020classification}, biodiversity conservation, and disaster response and management. All of these tasks have been explored in the past few decades using data coming from single EO sensors, i.e., hyperspectral imaging (HSI) instruments, which can simultaneously provide rich spectral and spatial information~\cite{hong2021interpretable}. However, such single-sensor data tend not sufficient to identify and recognize objects of interest.

Recent advances in RS technology have increased the availability of multi-sensor data, allowing for multiple representations of the same geographical region. Depending on the sensors' characteristics, the captured data can provide information with different characteristics for the same observed land-cover region. For example, synthetic-aperture radar (SAR) data provide the amplitude and phase geometrical information, while light detection and ranging (LiDAR) collects depth and intensity information, measuring the elevation of an object on the Earth's surface. Multispectral (MS) sensors measure the reflected light within specific wavelength ranges across the electromagnetic spectrum. Integrating these data with different modalities provides unique complementary information, allowing us to further obtain complete feature representations \cite{ghamisi2019multisource}.

However, from the perspective of an RS imaging system, the compatibility of spatial and spectral resolution is often questionable \cite{hong2019augmented}. The very high spectral resolution of HSI data often limits spatial resolution. As a consequence, spectral aliasing becomes a phenomenon that adversely affects land-cover classification in complex scenes \cite{ghamisi2015land}. In contrast, LiDAR provides elevation information that allows  distinguishing land-cover objects with identical spectral signatures but different elevations, such as roads and roofs with built-in cement. For instance, HSI data can accurately differentiate between water and grass. But, for instance, roads and roofs built using the same materials cannot be differentiated. Therefore, the complementary elevation information obtained from LiDAR can be greatly beneficial for classification purposes \cite{heiden2012urban}. Khodadadzadeh \textit{et. al} extracted geometric representations from LiDAR data for land-cover classification~\cite{khodadadzadeh2015fusion}. Many works have explored to fully exploit the complementary information between HSI and other sources of multimodal data such as LiDAR, SAR, MS or digital surface models (DSMs) \cite{SubFus,ghamisi2019multisource}. In this context, researchers used extended morphological profiles (EPs) and attribute profiles (APs) for joint feature extraction and classification of HSI and LiDAR data \cite{dalla2010morphological, liao2013graph, ghamisi2015land}. Rasti \textit{et al.} improved the joint extraction of EPs by applying total variation component analysis for feature fusion \cite{rasti2017hyperspectral}. Merentitis \textit{et al.} introduced an automatic fusion technique using the random forest classifier for joint HSI and LiDAR data classification \cite{merentitis2014automatic}.

To adequately exploit the information in both HSI and other sources of multimodal data like SAR, DSM, LiDAR, and DSM, we propose a new yet efficient multimodal fusion transformer (MFT) network  which uses multihead cross patch attention mechanism to fuse other sources of complementary information and HSI patch tokens for land cover classification. Our newly developed MFT incorporates the best properties of transformers encoder and  multihead cross patch attention for the fusion of RS classification tasks without introducing much computational overhead. The major contributions of this paper can be summarized as follows:
\begin{itemize}
    \item We propose a new multimodal fusion transformer (MFT) network to improve RS data fusion for land cover classification, where HSIs are used along with other multimodal data sources (e.g., LiDAR, MSI, SAR and DSM) in a transformer network to enhance the classification performance.
      
    \item We introduce a novel multihead cross patch attention (\texttt{mCrossPA}) mechanism for RS data fusion. The class token, which also contains supplementary information, is derived from multimodal data (e.g., LiDAR, MSI, SAR, and DSM), which are fed to the transformer network along with HSI patch tokens.
      
    \item The newly developed \texttt{mCrossPA} utilizes the widely used attention mechanism that can efficiently fuse the information from HSI patch tokens and existing \texttt{CLS} tokens into a new token that integrates multimodal features.
       
    \item We conduct extensive experiments on four public hyperspectral datasets, i.e., Houston, Trento, University of Southern Mississippi Gulfpark (MUUFL), and Augsburg. These experiments reveal the effectiveness of the proposed method. To illustrate the advantages of our approach, we use HSI data alone and also combine it with other sources of multimodal data.
\end{itemize}

The remaining of paper is organized as follows. Section~\ref{sec:related} discusses the traditional methods, classic deep learning-based methods, and transformer-based methods used for HSI classification. The pre-processing mechanisms for HSI and other sources of multimodal data and the components of the proposed MFT model, including a novel LiDAR and HSI cross-patch attention mechanism for transformer-based deep feature fusion and image classification are described in Section~\ref{sec:method}. Extensive experiments are then conducted, with an analysis of hyperparameter sensitivity and a discussion of the obtained results in Section~\ref{sec:exp}. The paper concludes with some remarks and hints at plausible future research lines in Section~\ref{sec:conc}.

\section{{Related Works}}
\label{sec:related}

\subsection{{Traditional Methods}}

Conventional methods have been widely used for HSI classification, even with limited training samples \cite{roy2020attention,roy2021morphological, hong2021spectralformer}. Generally speaking, these methods incorporate two steps. First, they represent the HSI data in feature space to reduce the dimensionality and extract a few highly informative features. Then, the extracted features are sent to a spectral classifier \cite{ahmad2021hyperspectral, roy2019hybridsn, ahmad2020fast, hong2021multimodalgan,hong2021joint}. Among conventional methods, support vector machines (SVMs) with non-linear kernels are very popular, especially when the training data are limited \cite{melgani2004classification}. The extreme learning machine (ELM) has also been widely used for unbalanced feature classification, and it was used in HSI classification. It has been shown that ELMs were able to achieve better performance when compared to SVMs. The random forest has been used due to its discriminative power when classifying unlabeled HSIs \cite{hang2020classification}. However, conventional methods encounter a performance bottleneck when the training data becomes complex, due to their limitations in terms of data fitting and representation ability. The aforementioned spectral classifiers consider HSIs as a collection of spectral measurements without considering their spatial arrangement. Spatial-spectral classifiers considerably boost the performance of spectral classifiers by incorporating spatial information such as the shape and size of different objects and textures. It is also worth mentioning that the spectral classifiers are less robust to noise compared to the spatial-spectral classifiers~\cite{ahmad2021hyperspectral,rs10030482}.

\subsection{\textcolor{black}{Conventional Deep Learning Methods}}

In recent years, multimodal data integration for RS data classification using deep learning (DL) methods has attracted significant attention~\cite{wang2022multi}. DL methods can efficiently learn spectral and spatial features from vast amounts of fragmented data. In the case of HSI-LiDAR data integration, both supervised and unsupervised DL methods have shown efficient performance in deep feature representation. Hong \textit{et al.} \cite{hong2022deep} developed a new technique for classifying multi-source RS data (HSI and LiDAR), which can effectively extract compact feature representations of multimodal RS data. Very recently, Hong \textit{et al.} first proposed a general and unified multimodal deep learning framework for RS image classification \cite{hong2021more} which provides an effective solution with respect to multi-source and multimodal RS products.

Unlike unsupervised methods, which learn the feature representation using observed data, supervised methods rely on ground reference training samples. Therefore, DL-based supervised techniques often demonstrate superiority in RS image classification tasks. One-dimensional convolution (CNN1D) \cite{hong2021graph}, two-dimensional convolution (CNN2D) \cite{makantasis2015deep}, and three-dimensional convolution (CNN3D) \cite{hamida2018deep} have shown success in HSI and LiDAR data classification. Recent studies show that DL methods can balance algorithm accuracy and robustness. On the other hand, shallow learning methods are dependent on the prior information obtained from training or observed data. He \textit{et al.} proposed a residual network (ResNet) with minimum information loss after each convolutional operation to redeem the vanishing gradient problem \cite{he2016deep}. Zhong \textit{et al.} designed a spectral-spatial residual network (SSRN) that can better utilize both spectral and spatial information for enhanced classification \cite{zhong2017spectral}. Roy \textit{et. al} explored the light-weighted paradigm by modeling spectral and spatial features extracted through the squeeze-and-excitation residual network, which can be combined with bag-of-features learning for accurate land-use and land-cover classification \cite{roy2020lightweight,roy2020fusenet}. To boost the performance of SSRNs, Zhu \textit{et al.} included another spatial and channel attention layer in the SSRN architecture to extract discriminative representation \cite{zhu2020residual}. To take advantage of the residual behavior of the network, one can extend it to form an even more complex model, i.e., the lightweight spectral-spatial squeeze-and-excitation attention \cite{roy2020lightweight} with adaptive kernels \cite{roy2020attention} and pyramidal residual networks \cite{paoletti2018deep}. Lightweight heterogeneous kernel convolution \cite{roy2021lightweight}, rotation equivariant CNN \cite{paoletti2020rotation} and gradient centralized convolution \cite{roy2021revisiting, roy2021new} enable efficient feature extraction and classification. On the other hand, the generative adversarial network also provides a solution for the classification of HSIs containing a class-wise imbalanced number of samples \cite{zhu2018generative, roy2021generative}.  

The networks mentioned above (and their variants) are generally insufficient for detecting subtle discrepancies among spectral dimensions. CNNs, which have shown their capacity in the task of extracting spatial-contextual information from HSIs, barely capture the sequence attributes, especially the middle and long-term dependencies. That leads to a decrease in performance, especially when the data comprise many classes with similar spectral signatures, creating problems in extracting diagnostic spectral attributes. On the other hand, recursive neural networks (RNNs) can accurately model spectral signatures from HSIs by accumulating them band by band in an orderly manner. As such, the order of spectral bands is essential in learning long-term dependencies and preventing gradient vanishing problems \cite{bengio1994learning}, which might further complicate the interpretation of spectrally salient changes. This is because HSIs contains limited samples, and RNNs cannot train the model concurrently, which limits the classification performance. In this work, we rethink HSI data classification using transformers to address the aforementioned limitations.

\subsection{{\textcolor{black}{Transformer-based Methods}}}

Unlike CNNs and RNNs, transformers are one of the most cutting-edge backbone networks due to their adoption of self-attention techniques, that are efficient for processing and analyzing sequential (or time series) data \cite{khan2021transformers}. Inspired by the success of transformers in computer vision \cite{dosovitskiy2020image}, several new transformer models have been developed in the last few years \cite{aleissaee2022transformers}. Although the transformer excels at capturing the information containing in spectral signatures, it cannot equally characterize local semantic elements and fails to make adequate use of spatial information. The original transformer \cite{vaswani2017attention} is a model based on the self-attention mechanism that is mainly used in natural language processing (NLP). A series of tokens are used as model’s input and multi-head attention is employed to draw global correlations in the input character sequence. Hong et al. recently introduced SpectralFormer \cite{hong2021spectralformer}, which can learn spectral representation information using neighboring bands and can construct a cross-layer transformer encoder module. The spectral-spatial feature tokenization (SSFT) transformer model \cite{sun2022spectral} was suggested to make use of the transformer’s capacity to extract local spatial semantic information and represent the link between neighboring sequences. The major drawback of this model is the number of parameters that are used. SSFT performs well with HSI data; however, it cannot include other multimodal data sources during the classification task.  Ding \textit{et al.} introduced a global–local transformer network (GLT-Net) to fully exploit the potential of transformer architectures for modelling the long-range dependencies as well as the convolution operator, which characterizes locally correlated features in joint classification of HSI and LiDAR data \cite{ding2022global}. To fuse multisource heterogeneous information and enhance the joint classification performance, a novel dual-branch network was presented by Zhao \textit{et al.} that consists of a transformer network and a hierarchical CNN \cite{zhao2022joint}. Yu \textit{et al.} created a cross-context capsule vision transformer (CapViT) to take advantage of long-range global feature interactions at various context scales for land cover classification with MS-LiDAR data \cite{yu2022capvit}. In order to capture the relationship between HSI and its complementary information such as LiDAR, as well as to fuse the multisource features and prevent the generation of redundant information, the local information interaction transformer (LIIT) model was introduced for land cover classification using a combination of HSI and LiDAR data \cite{zhang2022local}. A multi-attentive hierarchical fusion network (MAHiDFNet) was also developed in  \cite{wang2021multi} to implement feature-level fusion and classification of HSIs using LiDAR data. The connection between the structure and shape information of the HSI token and the CLS token was enhanced via spectral and spatial morphological convolution procedures in conjunction with a self-attention mechanism (morphFormer) for land cover classification in \cite{roy2023spectral}.
%\REMOVE{Although the transformer excels at capturing the information contained in spectral signatures, it cannot equally characterize local semantic elements and fails to make adequate use of spatial information. The original transformer \cite{vaswani2017attention} is a model based on the self-attention mechanism that is used in natural language processing (NLP). A series of tokens is used as the model’s input. Multi-head attention is employed to draw global correlations in the input character sequence. The spectral-spatial feature tokenization (SSFT) transformer model \cite{sun2022spectral} was suggested to make use of the transformer’s capacity to extract local spatial semantic information and represent the link between neighboring sequences. The major drawback of this model is the number of parameters that it uses. SSFT performs well with HSI data; however, it cannot include other multimodal data sources during the classification task.} 

%%%%%%%%%%%%%%%%%%%%%%%%%%%%%%%%%%%%%%%%%%%%%%%%%
\begin{figure*}[!t]
    \centering
    \includegraphics[clip=true, trim = 02 02 02 02, width=0.9\textwidth]{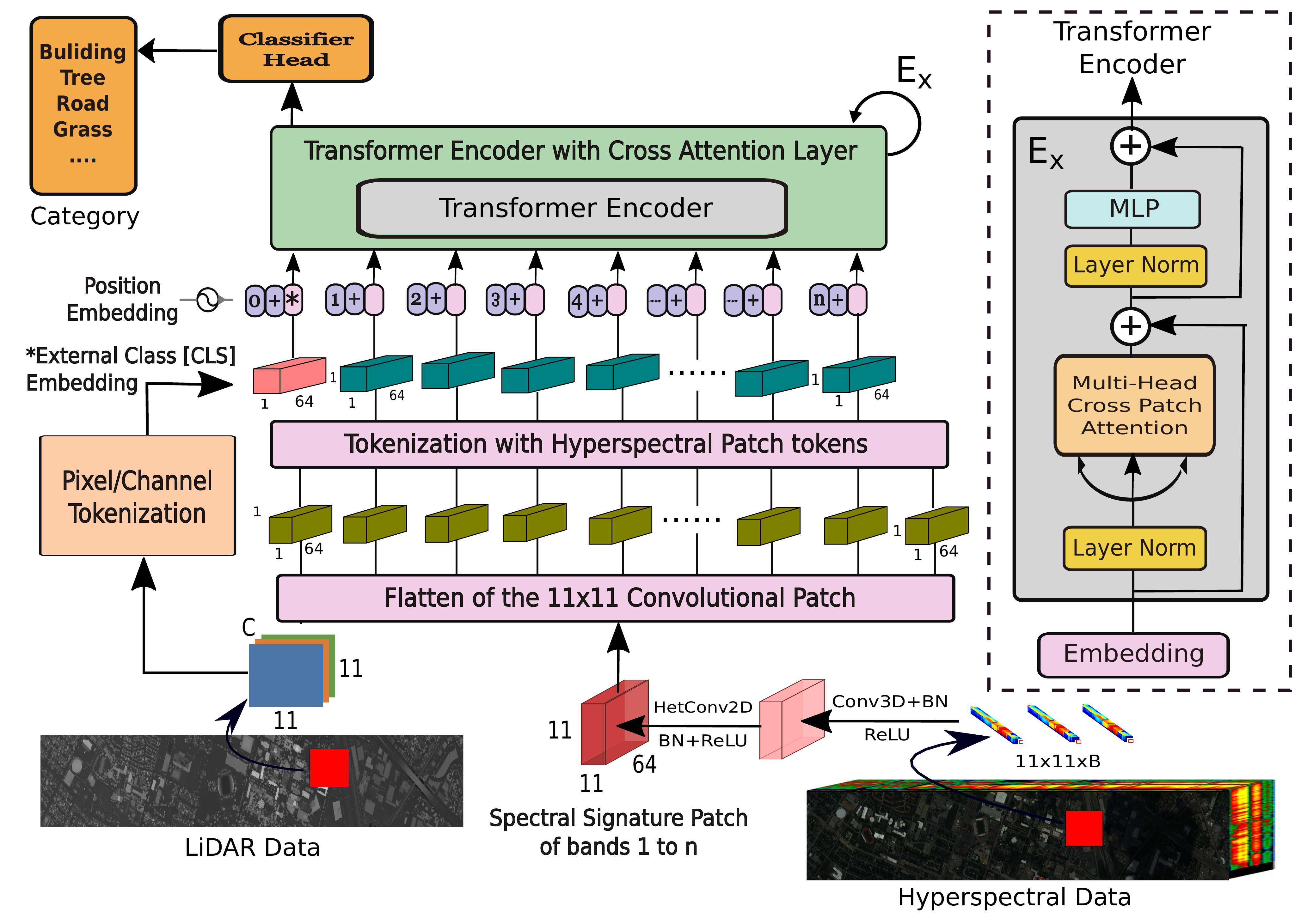}%trim - left, bottom, right, top
    \caption{\textcolor{black}{Graphical representation of the proposed multimodal fusion transformer network for HSI and LiDAR data fusion.}}
    \label{fig:propFormer}
\end{figure*}
%%%%%%%%%%%%%%%%%%%%%%%%%%%%%%%%%%%%%%%%%%%%%%%%%

\section{Methodology}
\label{sec:method}

%%%%%%%%%%%%%%%%%%%%%%%%%%%%%%%%%%%%%%%%%%%%%%%%%
\subsection{Pre-processing of HSI and LiDAR Data}
\label{sec:pre_proc}

Let us denote an HSI as $\mathbf{X_{H}} \in \mathcal{R}^{M\times{N}\times{B}}$ and a corresponding LiDAR image as $\mathbf{X_{L}} \in \mathcal{R}^{M\times{N}}$ where $M$ and $N$ respectively represent two spatial dimensions and $B$ refers to the number of spectral bands of the HSI. All the pixels are classified into $C$ land-cover classes denoted by $C = (y_1, y_2, \dots y_{c})$. A single pixel can be defined as $\mathbf{x}_{i,j} \in \mathbf{X_{H}}$, where $i =1, \dots M$ and $j =1, \dots N$ with $\mathbf{x}_{i,j} = [x_{i,j,1}, \dots x_{i,j,B}] \in \mathcal{R}^{B}$ containing $B$ spectral bands. \textcolor{black}{Patch extraction is performed in a pre-processing step, where a spectral-spatial cube $\mathbf{x}^{i,j}_{h} \in \mathcal{R}^{k\times{k\times B}}$ is obtained from the normalized HSI data $\mathbf{X_{H}}$ and a spatial image $\mathbf{x}^{i,j}_{l} \in \mathcal{R}^{k\times{k}}$ is obtained from the LiDAR data $\mathbf{X_{L}}$, with neighboring regions of size $(k\times{k})$ and centered at pixel $({i,j})$.}

By jointly exploiting the spectral-spatial information, we can increase the discriminative power of the feature learning network. For this reason, spectral-spatial cubes $\mathbf{x}^{(i,j)}_{h}$ are extracted from the original HSIs and stacked into $X_{H}$ prior to the feature extraction process. Similarly, for the LiDAR data, spatial patches of the same size $\mathbf{x}^{(i,j)}_{l}$ are extracted and stacked into $X_{L}$. Finally, the training and test samples in each class for $X_{H}$ and $X_{L}$ can be represented by

\begin{equation}
\begin{aligned}
D^{train} &= \{(x_h, x_l), y^{(i)}|i = 1, \dots P\} \\
D^{test} &= \{(x_h, x_l), y^{(i)}|i = 1, \dots Q\}
\end{aligned}
\end{equation}
where $x_h \in X_{H}$ and $x_l \in X_{L}$ are randomly chosen from the HSI and LiDAR data, $P$ and $Q$ represent the number of training and test samples, and $y^{(i)}$ is the actual class for the $i^{th}$ land-cover pixel. %\hl{($N$ has already been used as $\mathbf{X_{H}} \in \mathcal{R}^{M\times{N}\times{B}}$}

%%%%%%%%%%%%%%%%%%%%%%%%%%%%%%%%%%%%%%%%%%%%%%%%%

\subsection{Proposed Multimodal Fusion Transformer}
\label{sec:prop}

Fusion is crucial for effectively learning multimodal feature representations of RS data since complementary information of a model can enhance the learning of discriminative features. If we incorporate this technique into conventional transformer models, we may exponentially increase the number of parameters. For instance, in conventional ViT models, if we use the HSI data as input, we increase the linear projection's complexity due to the fact that the input has numerous spectral bands, which may lead to chances of over-fitting. If we assume that other multimodal data (e.g., LiDAR, SAR, or DSM) are concatenated to the HSIs to learn complementary information, the aforementioned problem is aggravated due to the increase in the number of bands. Hence, a transformer model can effectively learn multimodal information without increasing computational overhead of linear projection.

To address the aforementioned challenges, we introduce a multimodal fusion transformer (MFT) where the input patches are taken from the HSIs. Additionally, the \texttt{CLS} token is generated from a patch that describes the same spatial region as the HSI patch, taken from the corresponding LiDAR image (or other modalities like SAR, MSI, DSM etc). In addition, we propose a simple yet effective multi-head cross patch attention (\texttt{mCrossPA}) module to fuse the LiDAR (as \texttt{CLS}) token and the HSI patch tokens. Fig.~\ref{fig:propFormer} depicts the proposed multimodal fusion transformer (MFT) network architecture for multimodal RS image fusion to improve the land cover classification performance. For the sake of simplicity, only LiDAR data from the University of Houston dataset are considered, but the concept can be extended to other multimodal data as well. The results for different kinds of multimodal data (i.e., MS, SAR, and DSM) are reported in the results section. The objective of the proposed multimodal fusion transformer model is to learn the spectral-spatial patch embeddings instead of the band-wise embeddings of the input HSIs, in addition to enriching the abstract description of the \texttt{CLS} token by considering LiDAR as an external class embedding without introducing any computational overhead. The components of the proposed multimodal fusion transformer are discussed step by step as follows.

\subsubsection{HSI and LiDAR Feature Learning via CNNs}

% Let the spectral-spatial patch $\mathbf{X_{H}} \in \mathcal{R}^{M\times{N}\times{B}}$

The ability of automatic contextual modeling among features helps CNNs to make strong inferences and exhibit promising performance in HSI classification tasks. The availability of a large number of spectral bands in HSIs enables us to utilize the benefits of CNNs and control the depth of the output feature maps. It has already been shown that CNNs can extract high-level abstract features, which are also invariant to the source of data modality. Hence, the CNNs adopted in our proposed model extract high-level abstract features that are used as input in the transformer and reduce the HSI bands down to a suitable number. 

In order to extract robust and discriminative features from raw HSIs, sequential layers of Conv3D and \texttt{HetConv2D} are used. Fig. \ref{fig:propFormer} shows a graphical representation of the \texttt{Conv3D} and \texttt{HetConv2D} network for HSIs. The HSI cube $\textbf{X}_{H}$ of size ($11\times{11}\times{B}$) is first unsqueezed into shape (${1}\times11\times{11}\times{B}$), passed through the first \texttt{Conv3D} layer to produce $X_{in} = Conv3D(X_H)$ and the size of the considered kernels of that layer is chosen to be (${3}\times3\times{9}$) while the padding is (${1}\times1\times{0}$) to keep the spatial height and width of the output image equal to those of the input. \textcolor{black}{On the other hand, the \texttt{HetConv2D} \cite{singh2019hetconv} utilizes two \texttt{Conv2D} layers in parallel, one of which performs a group-wise convolution (with kernel size = 3, groups = 4 and padding = 1) and the other one performs a point-wise convolution (with kernel size = 1, groups = 1 and padding = 0).} The outputs from both these convolutions are added in element-wise ($\oplus$) fashion and returned from the \texttt{HetConv2D} layer. These can be defined as follows: 
\begin{equation}
  X_{out} = 
  \begin{cases}
  Conv2D(X_{in}, k=(3,3), s=1, g=4, p=1) \\ 
  ~~~~~~~~~~~~~~\oplus \\
  Conv2D(X_{in}, k=(1,1), s=1, g=1, p=1) 
  \end{cases}
\end{equation}

The shapes of the output feature maps after the \texttt{Conv3D} layer and the \texttt{HetConv2D} layer are (${8}\times11\times{11}\times{(B-8)}$) and (${11}\times{11}\times{64}$), respectively. Both of these layers are followed by their respective batch normalization (BN) \cite{ioffe2015batch} and ReLU activation layers. The BN layer is used to address the issue of overfitting (due to the small number of training samples) and also to accelerate the training performance. \textcolor{black}{ReLU plays an important role by introducing non-linearities into the output features maps in order to achieve efficient and smooth propagation of the loss gradient.}

%%%%%%%%%%%%%%%%%%%%%%%%%%%%%%%%%%%%%%%%%%%%%%%%%
\begin{figure}[!t]
    \centering
    \includegraphics[clip=true, trim = 20 410 130 10, width=0.99\columnwidth]{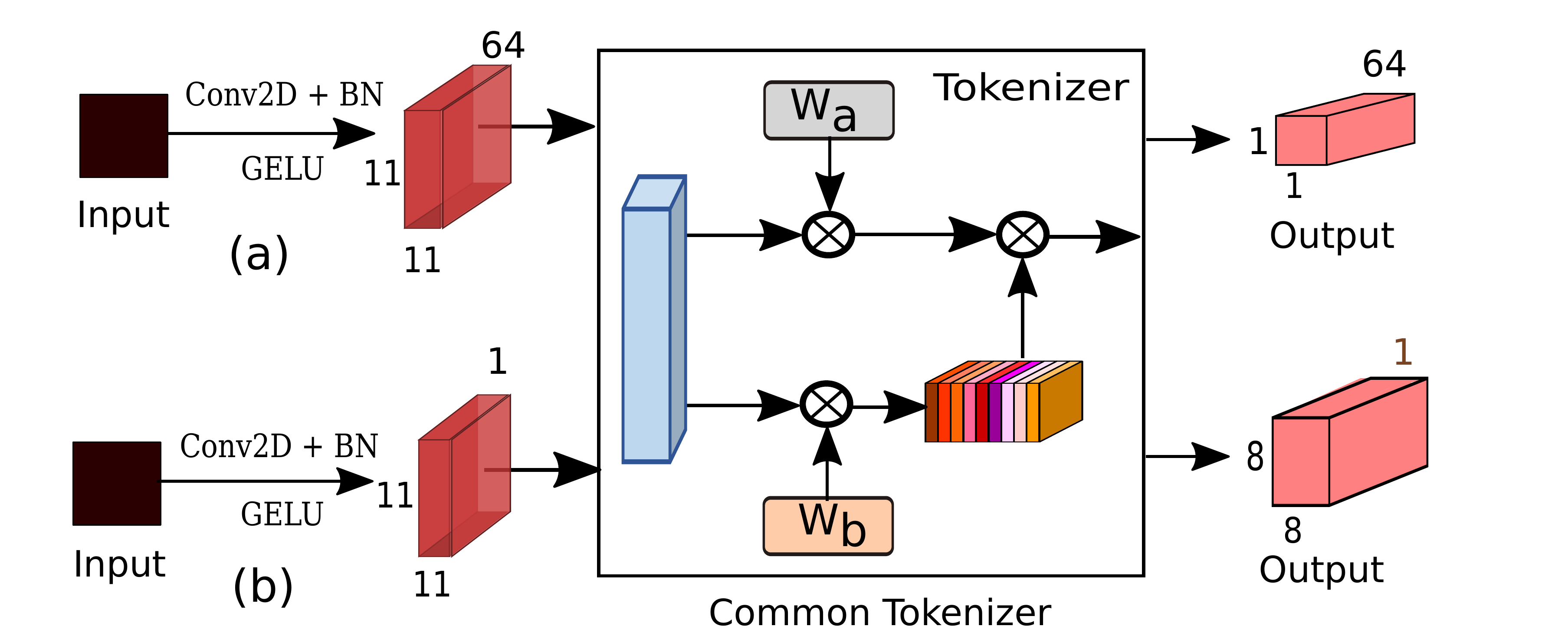}%trim - left, bottom, right, top
    \caption{Graphical representation of: (a) channel and (b) pixel tokenizer.}
    \label{fig:token}
\end{figure}
%%%%%%%%%%%%%%%%%%%%%%%%%%%%%%%%%%%%%%%%%%%%%%%%%

\subsubsection{HSI and LiDAR Tokenization}

Spectral and spatial features contain highly discriminative information, leading to higher accuracy in HSI classification tasks. HSI cubes of size \{$({11}\times{11})\times{64}$\} are flattened into feature patch tokens with shape (${1}\times{64}$) each. $n$ such patches are selected from $121$ patches using tokenization \cite{sun2022spectral}. The tokenization operation uses two learnable weights, i.e., $W_{a}$ and $W_{b}$, which are multiplied by the input to extract the key features. The following equation summarizes the process:
\begin{equation}
\begin{aligned}
X^{H}_{flat} &= T({\rm{Flatten}}(X_{out})) \\
X^{H}_{W_{a}} &= {\rm{softmax}}(T(X^{H}_{flat}.W_{aH})) \\
X^{H}_{W_{b}} &= X^{H}_{flat}.W_{bH} \\
X^{H}_{patch} &= X^{H}_{W_{a}}.X^{H}_{W_{b}}
\end{aligned}
\label{equ:hsi_patch_tokenization}
\end{equation}
where $T(.)$ is a transpose function, $X^{H}_{flat}\in \mathcal{R}^{121\times{64}}$, $W_{a}\in \mathcal{R}^{64\times{n}}$ and $W_{b}\in \mathcal{R}^{64\times{64}}$, $X^{H}_{W_{a}}\in \mathcal{R}^{n\times{121}}$ and $X^{H}_{W_{b}}\in \mathcal{R}^{121\times{64}}$ and $X^{H}_{patch}\in \mathcal{R}^{n\times{64}}$.

Focusing on the LiDAR data, a patch of size ($11\times{11}\times{C}$) extracted from a LiDAR image is first passed through either a pixel tokenization or a channel tokenization operation shown in Fig. \ref{fig:propFormer}. Both operations output a \texttt{CLS} token of shape ($1\times{64}$) generated from the input LiDAR patch by passing the it through a \texttt{Conv2D} layer, followed by a BN layer, a \texttt{GELU} activation function and a tokenizer, respectively. Fig. \ref{fig:token} illustrates the difference between the pixel and the channel tokenization operations. The \texttt{Conv2D} layer may reduce the number of channels to $1$ or increase it to $64$ depending on the variation used. The convolutional layer extracts different spatial elevation information of height and shape like roads and roof built-in cement having different heights, which facilitates the exchange of complementary information among the HSI patch tokens ($X^{H}_{patch}$). The LiDAR data may have only one channel to begin with, but other multimodal types of data like SAR, DSM, MSI, etc. may have several channels which need to be reduced to one, since the aim of the network is to use the LiDAR data as the \texttt{CLS} token which should have only one channel. The input and output spatial heights and widths are kept the same through the convolutional layer using a kernel of size 3 and padding of size 1. The LiDAR feature embeddings are now suitable to be used as the external classification (\texttt{CLS}) token. The whole step can be summarized as follows: 
\begin{equation}
\begin{aligned}
X^{L}_{conv} &= T({\rm{Flatten}}(Conv2D(X^{L}_{patch}))) \\
X^{L}_{W_{a}} &= {\rm{softmax}}(T(X^{L}_{conv}.W_{aL})) \\
X^{L}_{W_{b}} &= X^{L}_{conv}.W_{bL} \\
X^{L}_{cls} &= X^{L}_{W_{a}}.X^{L}_{W_{b}}
\end{aligned}
\label{equ:cls_embed}
\end{equation}
where $X^{L}_{patch}\in \mathcal{R}^{{11}\times{11}\times{C}}$, $T(.)$ is a transpose function and, in the case of the pixel tokenizer, $X^{L}_{conv}\in \mathcal{R}^{121\times{1}}$, $W_{aL}\in \mathcal{R}^{1\times{1}}$ and $W_{bL}\in \mathcal{R}^{1\times{64}}$. In the case of the channel tokenizer, $X^{L}_{conv}\in \mathcal{R}^{121\times{64}}$, $W_{aL}\in \mathcal{R}^{64\times{1}}$ and $W_{bL}\in \mathcal{R}^{64\times{64}}$. In both cases,  $X^{L}_{W_{a}}\in \mathcal{R}^{1\times{121}}$ and $X^{L}_{W_{b}}\in \mathcal{R}^{121\times{64}}$ and $X^{L}_{cls}\in \mathcal{R}^{1\times{64}}$.

The \texttt{CLS} token, i.e., the LiDAR feature embeddings are then concatenated to the $n$ HSI patch embeddings, making a total of $(n + 1)$ patches from $n$ as shown in Eq. (\ref{equ:HLCLS}). A size of $64$ is used because it is a power of $2$, which helps in calculating the head dimension.
\begin{equation}
  \widehat{X}^{HL}  = [X^{L}_{cls} ~\parallel~X^{H}_{patch} ]
  \label{equ:HLCLS}
\end{equation}
To retain the positional information, trainable position embeddings are added to the patch embeddings to preserve the semantic textural information of the image cube within the image patch tokens. For visual understanding, we refer to Fig. \ref{fig:propFormer} where the position embeddings are added in element-wise fashion to all the patches ($1$ to $n + 1$). This process is followed by a dropout layer which helps in reducing the effects of the overfitting of the network. The above process can be summarized by the following equation:
\begin{equation}
\begin{aligned}
  X^{HL}  &= \mathcal{DP}({X}^{L}_{cls} \oplus \mathcal{PE}~\parallel~{X}^{H}_{patch} \oplus \mathcal{PE}) \\
  &= (\widehat{X}^{L}_{cls}~\parallel~\widehat{X}^{H}_{patch})
  \label{equ:HLCLS2}
 \end{aligned} 
\end{equation}
where $\mathcal{DP}$ is a dropout layer with a value of 0.1 and $\mathcal{PE}$ is a learnable position embedding.

%%%%%%%%%%%%%%%%%%%%%%%%%%%%%%%%%%%%%%%%%%%%%%%%%
\begin{figure}[!t]
    \centering
    \includegraphics[clip=true, trim = 05 02 05 02, width=0.99\columnwidth]{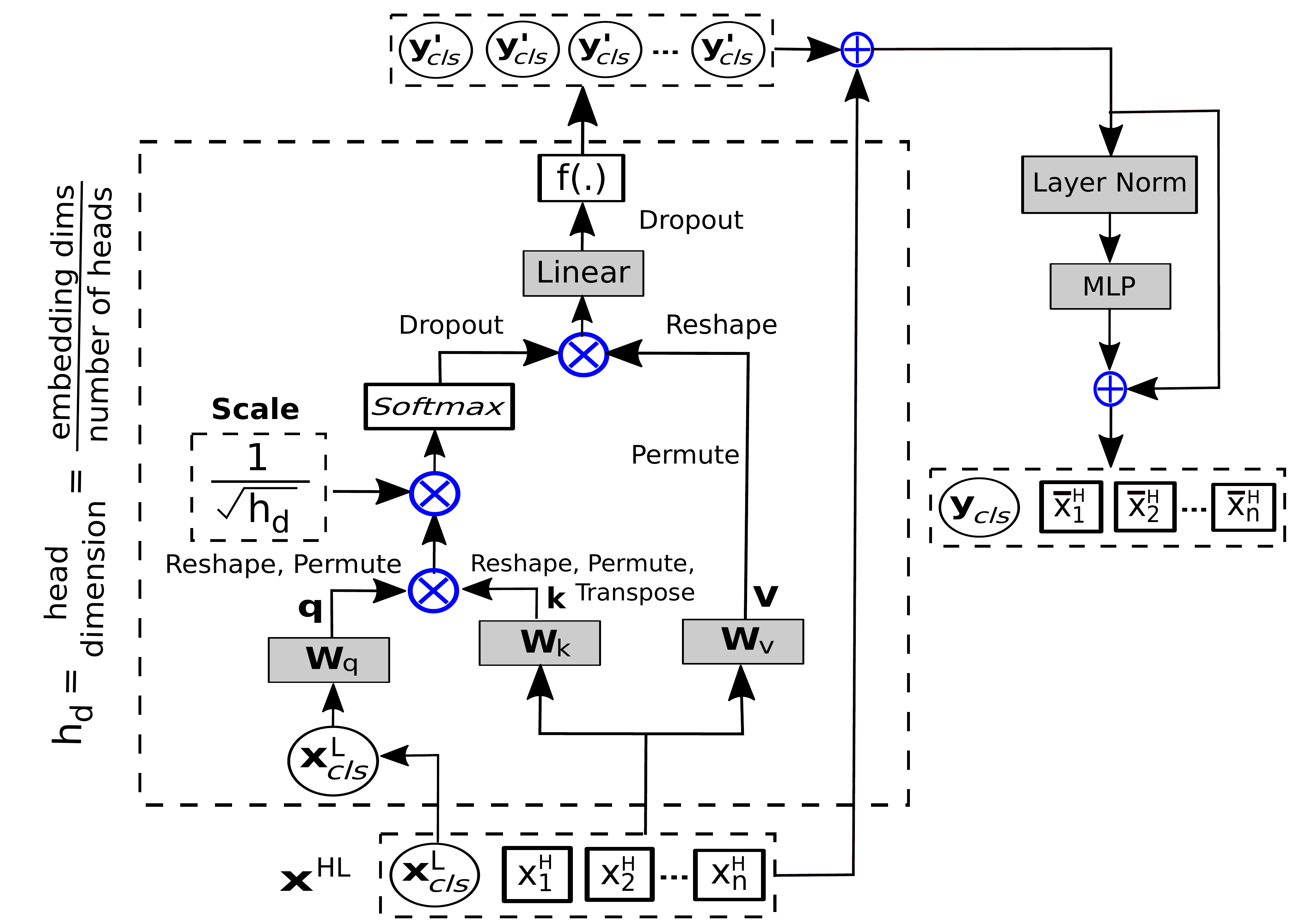} %trim - left, bottom, right, top
    \caption{\textcolor{black}{Multihead cross patch attention taking example of LiDAR and HSI data.}} 
    \label{fig:propAttn}
\end{figure}
%%%%%%%%%%%%%%%%%%%%%%%%%%%%%%%%%%%%%%%%%%%%%%%%%

\begin{algorithm}[htbp]\footnotesize\scriptsize
\SetAlgoLined
\vspace{.cm}
\KwIn{\vspace{.cm}$\mathbf{X^{HL}}$, $\mathbf{X^{L}_{cls}}$, $\mathbf{X^{HL}_{patch}}$, $h_d$}
\vspace{.cm}
\KwOut{$\mathbf{y_{cls}}$}
\vspace{.cm}
\textbf{Multihead Cross Patch Attention (Begin)}\\
\vspace{.cm}
\hskip 10pt {\bf Step 1.} $\mathbf{Q={X}^{L}_{cls}W_q,\hskip 10pt K=X^{HL}W_k,\hskip 10pt V=X^{HL}W_v}$,\\
\vspace{.cm}
% \hskip 42pt $ \mathbf{Q' = \textup{reshape}(Q), K' = \textup{reshape}(K)}$,\\
% \vspace{.cm}
% \hskip 42pt $ \mathbf{V' =\textup{reshape}(V)}$\\
\vspace{.cm}
\hskip 10pt {\bf Step 2.} $\mathbf{Z} = \texttt{softmax}\mathbf{(QK^T/\sqrt{h_d})}$, (\ref{equ:7a})\\
\vspace{.cm}
\hskip 45pt      $\texttt{mCrossPA}(\mathbf{X^{HL}}) = \mathcal{DP}(\mathbf{\mathbf{W_l}ZV})$ (Eq.~(\ref{equ:7b}))\\
\vspace{.cm}
{\bf Multihead Cross Patch Attention (End)}\\
\vspace{.cm}
{\bf Step 3.} $\mathbf{y'_{cls}} = \texttt{mCrossPA}(\mathbf{X^{HL}}))$ (Eq.~(\ref{equ:8a}))\\
\vspace{.cm}
{\bf Step 4.} $\mathbf{y_k} = f(y'_{cls}) \oplus (\mathbf{[\widehat{X}^{L}_{cls}~\parallel~\widehat{X}^{H}_{patch}]})$ (Eq.~(\ref{equ:8b})), \\
\vspace{.cm}
\hskip 32pt      $\mathbf{X^{HL}_{k}} = \mathbf{y_k} + \texttt{MLP}(\texttt{LN}(\mathbf{y_k}))$ (Eq.~(\ref{equ:9})), \\
\vspace{.cm}
\hskip 32pt      $\mathbf{y_{cls}} = \mathbf{X^{HL}_k}(1,:)$ \\
\caption{Transformer Encoder with \texttt{mCrossPA}}
\label{alg:MHSPA}
\end{algorithm}

\subsubsection{LiDAR and HSI Cross Patch Attention}

\textcolor{black}{The \texttt{CLS} ($X^{L}_{cls}$) token plays a vital role in learning the abstract representation of the entire HSI patch by exchanging the information between the patch tokens and itself. This entire process takes place in the transformer encoder blocks, where each transformer encoder block comprises a residual multihead cross patch attention block and a residual \texttt{MLP} block, respectively. The residual attention block starts with a layer normalization (LN) operation followed by a self-attention layer, after which the output is added in element-wise ($\oplus$) fashion to the input of the LN (shown in Fig.~\ref{fig:propFormer}). Similar to the attention block, the \texttt{MLP} block also starts with an LN operation, but the input of the LN is added element-wise ($\oplus$) to the output of \texttt{MLP} layer, as shown in Fig.~\ref{fig:propFormer}.}

\textcolor{black}{Fig.~\ref{fig:propAttn} illustrates the cross patch attention module for the LiDAR data (as a \texttt{CLS}) and for the HSI data (as a patch token). More specifically, the generated patch tokens from the HSI feature maps are collected and then concatenated with the external \texttt{CLS} tokens from the LiDAR data from the same spatial region corresponding to the HSI patch, followed by the combination of position embeddings and dropout, as shown in Eqs.~(\ref{equ:HLCLS}) and (\ref{equ:HLCLS2}) and illustrated in Fig.~\ref{fig:propFormer}. After that, the module performs cross patch attention (\texttt{CrossPA}) between $X^{L}_{cls}$ and $X^{HL}_{patch}$, where query ($\mathbf{Q} = \mathbf{X}^{L}_{cls}\mathbf{W}_q \hskip 10pt \rm{where} \hskip 10pt W_q\in \mathcal{R}^{64\times{64}}$) along with the key ($\mathbf{K} = \mathbf{X}^{HL}\mathbf{W}_k \hskip 10pt \rm{where} \hskip 10pt W_k\in \mathcal{R}^{64\times{64}}$) and the value ($\mathbf{V} = \mathbf{X}^{HL}\mathbf{W}_v \hskip 10pt \rm{where} \hskip 10pt W_v\in \mathcal{R}^{64\times{64}}$) are fused to generate the \texttt{CLS} token ($y_{cls}$) at the end of the \texttt{CrossPA} block. A linear projection layer ($\mathbf{W}_l\in \mathcal{R}^{64\times{64}}$) is applied to the final fused \texttt{CLS} token, which precedes a dropout layer ($\mathcal{DP}$) with a value of $0.1$. Mathematically, the cross attention can be formulated as follows:}
\begin{subequations}
\begin{align}
    Z = \rm{softmax}(\frac{\mathbf{QK}^{T}}{\sqrt{h_d}}) \label{equ:7a} \\ 
    {\rm{CrossPA}(X^{HL})} = \mathcal{DP}(\mathbf{W}_l\mathbf{ZV}) \label{equ:7b}
\end{align}    
\end{subequations}
\textcolor{black}{where ${Z}\in \mathcal{R}^{1\times{64}}$ and $h_d$ = embedding dimension/number of heads. As shown in the self-attention module, if the number of heads is more than one, then the \texttt{CrossPA} will become a multi-head cross attention and, upon doing so, it can be represented as \texttt{mCrossPA}. The connections between the various patch tokens and CLS token are further strengthened by applying cross patch attention with numerous heads. The output $y_{k}$ of the \texttt{mCrossPA} module for a given embedding $X^{HL}_{k-1}$ with layer normalization (LN) and residual shortcut, where $k$ is the $k^{th}$ transformer encoder block, is defined below:
\begin{subequations}
\begin{align}
    y'_{cls} &= {\rm{mCrossPA}(LN}(X^{HL}_{k-1})) \label{equ:8a}\\
    y_k &= {f}(y'_{cls}) \oplus [\widehat{X}^{L}_{cls}~\parallel~\widehat{X}^{H}_{patch}] \label{equ:8b}
\end{align}    
\end{subequations}
where $y_{k}\in\mathcal{R}^{64\times{64}}$ and $f(.)$ is a function for broadcasting the dimension of $y'_{cls}$ that is used to make the dimension the same as that of input $X^{HL}_{k-1}$, as shown in Fig.~\ref{fig:propAttn}. This output $y_{k}$ is then passed through multi-layer perceptron (\texttt{MLP}) block after layer normalization, followed by a residual shortcut to generate $X^{HL}_{k}$, which can be treated as input to $(k+1)$th transformer encoder block, as shown in Fig.~\ref{fig:propAttn}, and can be defined as follows:
\begin{equation}
\begin{aligned}
    X^{HL}_{k} &= y_{k} \oplus {\rm{MLP}(LN}(y_{k}))
    \label{equ:9}
\end{aligned}    
\end{equation}
The number of heads $h_d$ is set to $8$ in the proposed model. Finally, the output from the $k$th transformer encoder blocks $y_{cls} = X^{HL}_{k}(1,:)$ is fed into the classifier head to get the final classification results.}

\section{Experiments}
\label{sec:exp}

%%%%%%%%%%%%%%%%%%%%%%%%%%%%%%%%%%%%%%%%%%%%%%%%%
\subsection{HSI Datasets}
\label{subsec:hsi_data}
In this section, we consider four different HSIs and their associated multimodal sources of data (LiDAR, MS, SAR, and DSM) to evaluate the performance of the proposed multimodal fusion transformer network. The experimental datasets include the University of Houston (UH), Trento, MUUFL Gulfport, and Augsburg scenes. 

\begin{figure}[!t]
\centering
        \begin{subfigure}{0.18\columnwidth}
            \centering
	        \includegraphics[width=1.5cm,height=7cm]{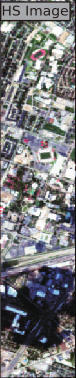}
		    \caption{Pseudo-color Map}
	       % \label{Fig.3A}
        \end{subfigure}
        \begin{subfigure}{0.18\columnwidth}
	    \centering
	    \includegraphics[width=1.5cm,height=7cm]{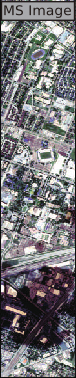}
    	\caption{MSI Map} 
% 		\label{Fig.3B}
	    \end{subfigure}
	    \begin{subfigure}{0.18\columnwidth}
	    \centering
	    \includegraphics[width=1.5cm,height=7cm]{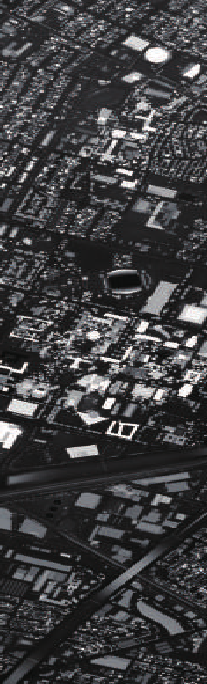}
    		\caption{LiDAR Map} 
		  %  \label{Fig.3B}
	    \end{subfigure}
        \begin{subfigure}{0.18\columnwidth}
            \centering
	        \includegraphics[width=1.5cm,height=7cm]{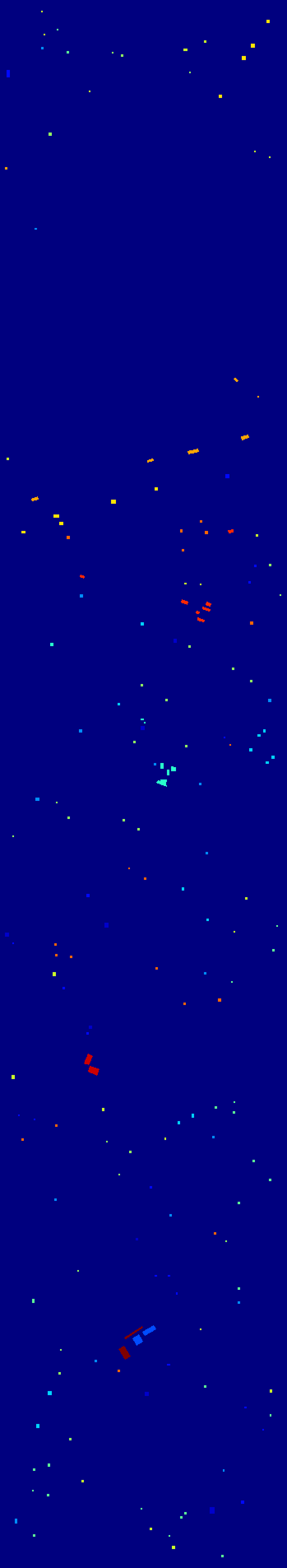}
		    \caption{Training Samples}
	       % \label{Fig.3C}
        \end{subfigure}
	    \begin{subfigure}{0.18\columnwidth}
	        \centering
		    \includegraphics[width=1.5cm,height=7cm]{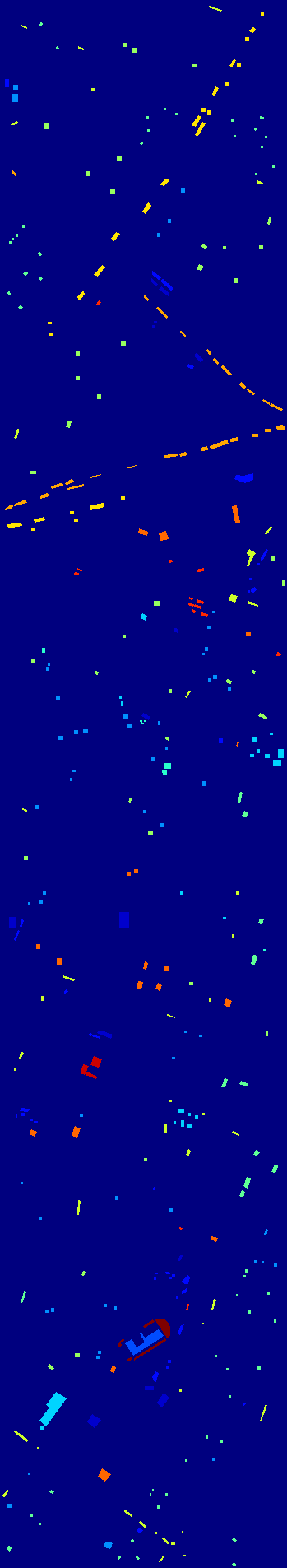}
    		\caption{Test Samples} 
		  %  \label{Fig.3D}
	    \end{subfigure}
        \vspace{0.3cm}

    \resizebox{0.8\linewidth}{!}{\begin{tabular}{cccc||cccc} \hline
    Color & Land cover & Train & Test & Color & Land cover & Train & Test \\ \hline
    \fboxsep=1mm \fboxrule=1mm
    \fcolorbox{houston_data_Background!}{houston_data_Background!}{\null} & Background & 662013 & 652648 & \fboxsep=1mm \fboxrule=1mm \fcolorbox{houston_data_Grass-healthy!}{houston_data_Grass-healthy!}{\null} & Grass-healthy & 198 & 1053 \\
    \fboxsep=1mm \fboxrule=1mm
    
    \fcolorbox{houston_data_Grass-stressed!}{houston_data_Grass-stressed!}{\null} & Grass-stressed & 190 & 1064 & \fboxsep=1mm \fboxrule=1mm
    \fcolorbox{houston_data_Grass-synthetic!}{houston_data_Grass-synthetic!}{\null} & Grass-synthetic & 192 & 505\\
    \fboxsep=1mm \fboxrule=1mm
    
    \fcolorbox{houston_data_Tree!}{houston_data_Tree!}{\null} & Tree & 188 & 1056 & \fboxsep=1mm \fboxrule=1mm
    \fcolorbox{houston_data_Soil!}{houston_data_Soil!}{\null} & Soil & 186 & 1056\\
    \fboxsep=1mm \fboxrule=1mm
    
    \fcolorbox{houston_data_Water!}{houston_data_Water!}{\null} & Water & 182 & 143 & \fboxsep=1mm \fboxrule=1mm
    \fcolorbox{houston_data_Residential!}{houston_data_Residential!}{\null} & Residential & 196 & 1072\\
    \fboxsep=1mm \fboxrule=1mm
    
    \fcolorbox{houston_data_Commercial!}{houston_data_Commercial!}{\null} & Commercial & 191 & 1053 & \fboxsep=1mm \fboxrule=1mm
    \fcolorbox{houston_data_Road!}{houston_data_Road!}{\null} & Road & 193 & 1059\\
    \fboxsep=1mm \fboxrule=1mm
    
    \fcolorbox{houston_data_Highway!}{houston_data_Highway!}{\null} & Highway & 191 & 1036 & \fboxsep=1mm \fboxrule=1mm
    \fcolorbox{houston_data_Railway!}{houston_data_Railway!}{\null} & Railway & 181 & 1054\\
    \fboxsep=1mm \fboxrule=1mm
    
    \fcolorbox{houston_data_Parking-lot1!}{houston_data_Parking-lot1!}{\null} & Parking-lot1 & 192 & 1041 & \fboxsep=1mm \fboxrule=1mm
    \fcolorbox{houston_data_Parking-lot2!}{houston_data_Parking-lot2!}{\null} & Parking-lot2 & 184 & 285\\
    \fboxsep=1mm \fboxrule=1mm
    
    \fcolorbox{houston_data_Tennis-court!}{houston_data_Tennis-court!}{\null} & Tennis-court & 181 & 247 & \fboxsep=1mm \fboxrule=1mm
    \fcolorbox{houston_data_Running-track!}{houston_data_Running-track!}{\null} & Running-track & 187 & 473 
    \fboxsep=1mm \fboxrule=1mm \\ \hline
    % & Total samples & 664845 & 664845 \\\hline
    \end{tabular}}
    \caption{the University of Houston (UH) scene. (a) Pseudo-color image from the HSI data using bands 64, 43, and 22, respectively. (b) Grayscale image from the MSI data (c) Grayscale image from the LiDAR data, (d) Ground truth of disjoint training samples, and (e) Ground truth of disjoint test samples. The table represents class-specific land-cover types and the number of disjoint training and test samples.}
    \label{fig:UH}
\end{figure}

$\bullet$ The \textbf{University of Houston} dataset was collected by the Compact Airborne Spectrographic Imager (CASI) and was provided by IEEE Geoscience and Remote Sensing Society in 2013 as part of its Data Fusion Contest. The dataset comprises an HSI, an MS image and a LiDAR image. All the images are made up of $340 \times 1905$ pixels and the HSI has $144$ bands, whereas the MS image has $8$ spectral bands. This dataset has a spatial resolution of $2.5$ meters per pixel and wavelength ranging between $0.38-1.05 ~\mu{m}$. The ground truth has $15$ different land-cover and land-use classes. Furthermore, the samples of the $15$ land-cover classes are divided into fixed size training and testing samples. Fig. \ref{fig:UH} represents the $15$ different categories of land-cover and land-use and the training and test samples associated with each.

\begin{figure}[!t]
% \begin{tabular}{ l l}
\centering
\begin{subfigure}{0.22\columnwidth}
    \centering
    \includegraphics[width=2cm, height=4cm]{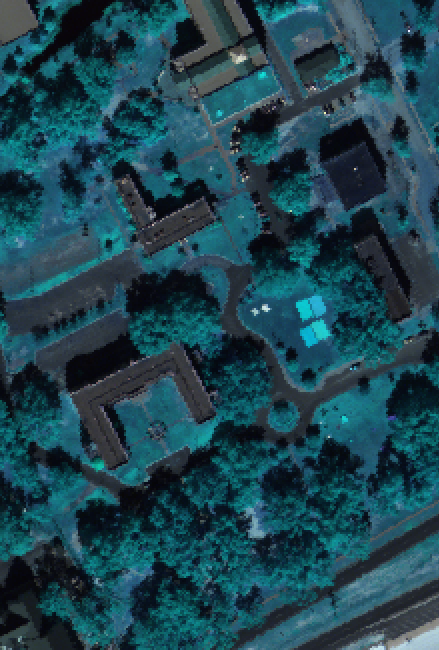}
	\caption{Pseudo-color Map}
	\label{Fig.4A}
\end{subfigure}
% &
\begin{subfigure}{0.22\columnwidth}
    \centering
	\includegraphics[width=2cm, height=4cm]{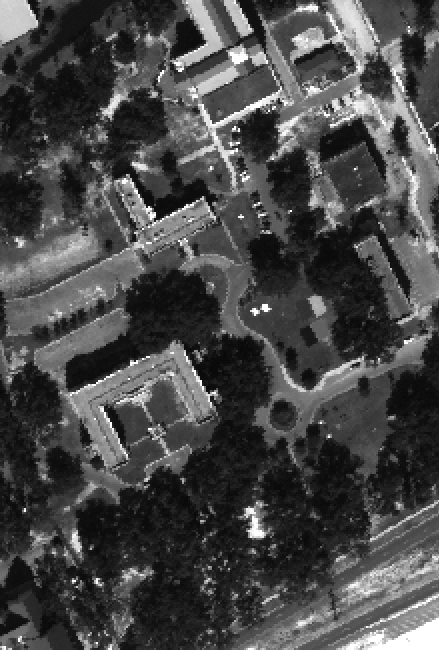}
    \caption{LiDAR Map} 
	\label{Fig.4B}
\end{subfigure}
% &
\begin{subfigure}{0.22\columnwidth}
   \centering
   \includegraphics[width=2cm, height=4cm]{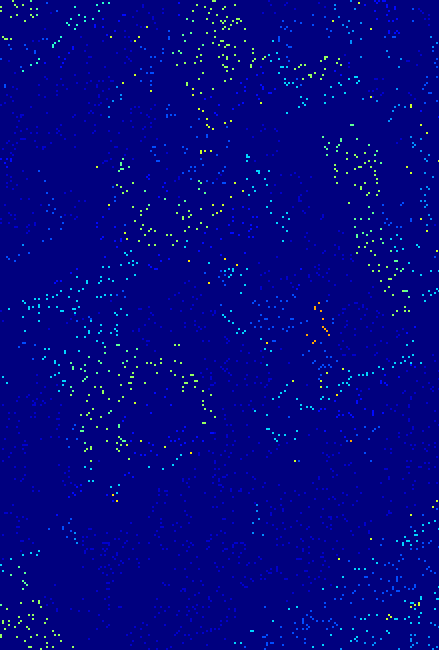}
	\caption{Training Samples}
    \label{Fig.4C}
\end{subfigure}
% &
\begin{subfigure}{0.22\columnwidth}
   \centering
   \includegraphics[width=2cm, height=4cm]{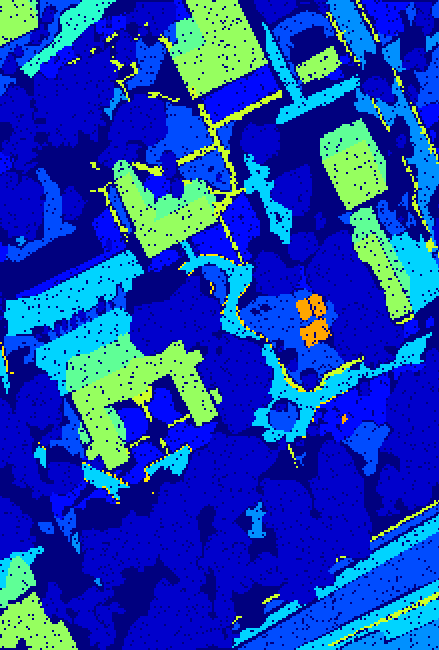}
	\caption{Test Samples}
    \label{Fig.4D}
\end{subfigure}
% \end{tabular}
\vspace{0.3cm}
\resizebox{0.8\linewidth}{!}{\begin{tabular}{cccc||cccc}\hline
    Color & Land cover & Train & Test & Color & Land cover & Train & Test \\ \hline
    \fboxsep=1mm \fboxrule=1mm
    
    \fcolorbox{MUUFL_data_Background!}{MUUFL_data_Background!}{\null} & Background & 68817   & 20496 & \fboxsep=1mm \fboxrule=1mm
    \fcolorbox{MUUFL_data_Trees!}{MUUFL_data_Trees!}{\null} & Trees & 1162   & 22084\\
    \fboxsep=1mm \fboxrule=1mm
    
    \fcolorbox{MUUFL_data_Grass-Pure!}{MUUFL_data_Grass-Pure!}{\null} & Grass-Pure & 214   & 4056 & \fboxsep=1mm \fboxrule=1mm
    \fcolorbox{MUUFL_data_Grass-Groundsurface!}{MUUFL_data_Grass-Groundsurface!}{\null} & Grass-Groundsurface & 344   & 6538\\
    \fboxsep=1mm \fboxrule=1mm
    
    \fcolorbox{MUUFL_data_Dirt-And-Sand!}{MUUFL_data_Dirt-And-Sand!}{\null} & Dirt-And-Sand & 91   & 1735 & \fboxsep=1mm \fboxrule=1mm
    \fcolorbox{MUUFL_data_Road-Materials!}{MUUFL_data_Road-Materials!}{\null} & Road-Materials & 334   & 6353\\
    \fboxsep=1mm \fboxrule=1mm
    
    \fcolorbox{MUUFL_data_Water!}{MUUFL_data_Water!}{\null} & Water & 23   & 443 & \fboxsep=1mm \fboxrule=1mm
    \fcolorbox{MUUFL_data_Buildings'-Shadow!}{MUUFL_data_Buildings'-Shadow!}{\null} & Buildings'-Shadow & 112   & 2121\\
    \fboxsep=1mm \fboxrule=1mm
    
    \fcolorbox{MUUFL_data_Buildings!}{MUUFL_data_Buildings!}{\null} & Buildings & 312   & 5928 & \fboxsep=1mm \fboxrule=1mm
    \fcolorbox{MUUFL_data_Sidewalk!}{MUUFL_data_Sidewalk!}{\null} & Sidewalk & 69   & 1316\\
    \fboxsep=1mm \fboxrule=1mm
    
    \fcolorbox{MUUFL_data_Yellow-Curb!}{MUUFL_data_Yellow-Curb!}{\null} & Yellow-Curb & 9   & 174 & \fboxsep=1mm \fboxrule=1mm
    \fcolorbox{MUUFL_data_ClothPanels!}{MUUFL_data_ClothPanels!}{\null} & ClothPanels & 13   & 256 
    \fboxsep=1mm \fboxrule=1mm \\\hline
    % &   Total samples & 71500   & 71500  \\\hline
    \end{tabular}}
    \caption{MUUFL scene. (a) True-color image from the HSI data using bands 40, 20, and 10, respectively (b) Grayscale image from the LiDAR data and (c) Ground truth of MUUFL scene. The table represents class-specific land-cover types and the number of randomly selected 5\% training and the remaining 95\% test samples.}
\label{fig:MUUFL}
\end{figure}

$\bullet$ The \textbf{MUUFL Gulfport} scene was collected over the campus of the University of Southern Mississippi in November 2010 using the Reflective Optics System Imaging Spectrometer (ROSIS) sensor \cite{gader2013muufl, du2017scene}. There are $325\times 220$ pixels with $72$ spectral bands in the HSI of this dataset. The LiDAR image of this dataset contains elevation data of $2$ rasters. The $8$ initial and final bands were removed due to noise, giving a total of $64$ bands. The data depicts $11$ urban land-cover classes containing $53687$ ground truth pixels. Fig. \ref{fig:MUUFL} displays the distribution of $5\% $ of the samples randomly selected from every class.
%%%%%%%%%%%%%%%%%%%%%%%%%%%%%%%%%%%%%%%%%%%%%%%%%
%%%%%%%%%%%%%%%%%%%%%%%%%%%%%%%%%%%%%%%%%%%%%%%%%

\begin{figure}[!t]
\begin{subfigure}{0.22\columnwidth}
    \centering
    \includegraphics[width=2cm, height=7cm]{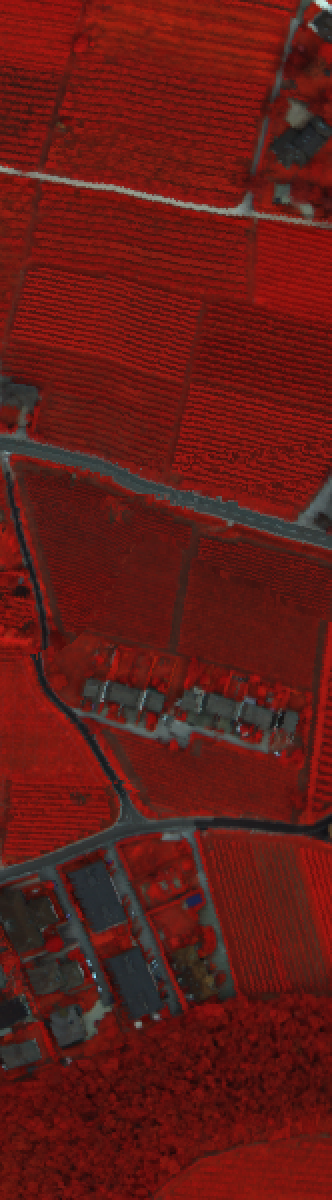}
    \caption{Pseudo-color Map}
    \label{Fig.5A}
\end{subfigure}
\begin{subfigure}{0.22\columnwidth}
    \centering
    \includegraphics[width=2cm, height=7cm]{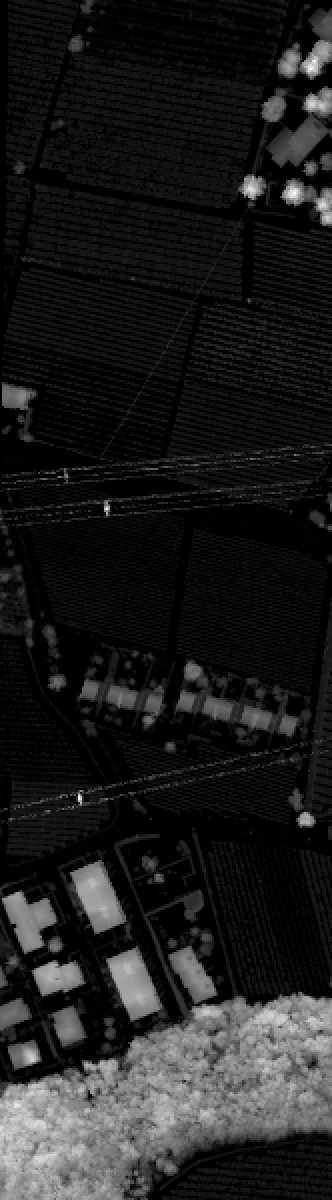}
	\caption{LiDAR Map} 
    \label{Fig.5B}
\end{subfigure}
\begin{subfigure}{0.22\columnwidth}
    \centering
    \includegraphics[width=2cm, height=7cm]{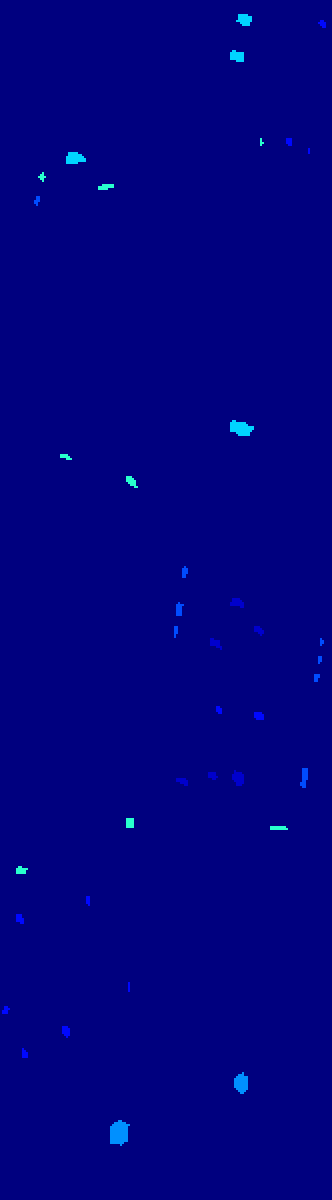}
    \caption{Training Samples}
    \label{Fig.5C}
\end{subfigure}
\begin{subfigure}{0.22\columnwidth}
    \centering
    \includegraphics[width=2cm, height=7cm]{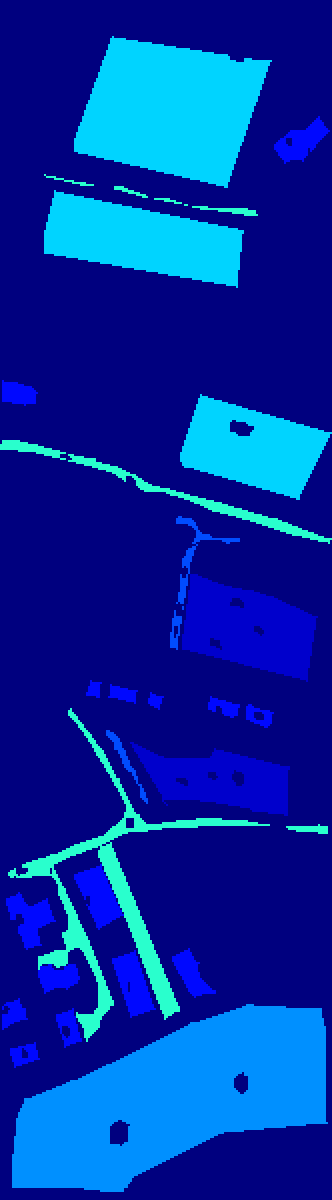}
	\caption{Test Samples} 
    \label{Fig.5D}
\end{subfigure}
\vspace{0.3cm}
\centering
\resizebox{0.90\linewidth}{!}{\begin{tabular}{cccc||cccc}\hline
    Color & Land cover & Train & Test & Color & Land cover & Train & Test \\ \hline
    \fboxsep=1mm \fboxrule=1mm
    
    \fcolorbox{trento_data_Background!}{trento_data_Background!}{\null} & Background & 98781 & 70205 & \fboxsep=1mm \fboxrule=1mm
    \fcolorbox{trento_data_Apples!}{trento_data_Apples!}{\null} & Apples & 129 & 3905\\
    \fboxsep=1mm \fboxrule=1mm
    
    \fcolorbox{trento_data_Buildings!}{trento_data_Buildings!}{\null} & Buildings & 125  & 2778 & \fboxsep=1mm \fboxrule=1mm
    \fcolorbox{trento_data_Ground!}{trento_data_Ground!}{\null} & Ground & 105   & 374\\
    \fboxsep=1mm \fboxrule=1mm
    
    \fcolorbox{trento_data_Woods!}{trento_data_Woods!}{\null} & Woods & 154   & 8969 & \fboxsep=1mm \fboxrule=1mm
    \fcolorbox{trento_data_Vineyard!}{trento_data_Vineyard!}{\null} & Vineyard & 184   & 10317\\
    \fboxsep=1mm \fboxrule=1mm
    
    \fcolorbox{trento_data_Roads!}{trento_data_Roads!}{\null} & Roads & 122   & 3052 
    \fboxsep=1mm \fboxrule=1mm \\\hline
    % &   Total samples & 99600   & 99600 \\ \hline
    \end{tabular}}

\caption{Trento data. (a) True color image from the HSI using bands 40, 20, and 10, respectively. (b) Grayscale image from the LiDAR data. (c) Ground truth of disjoint training samples. (d) Ground truth of disjoint test samples. The table represents class-specific land-cover types and the number of disjoint training and test samples.}
\label{fig:Trento}
\end{figure}

$\bullet$ AISA Eagle sensors were used to collect HSI data over rural regions in the south of \textbf{Trento}, Italy, where the Optech ALTM 3100EA sensors collected LiDAR data. There are $63$ bands in each HSI with wavelength ranging from $0.42-0.99 \mu{m}$, and $1$ raster in the LiDAR data that provides elevation information. The spectral resolution is $9.2~nm$, and the spatial resolution is $1$ meters per pixel. The scene comprises $6$ vegetation land-cover classes that are mutually exclusive and a pixel count of $600 \times 166$. Moreover, the training and test samples are disjoint. Fig. \ref{fig:Trento} provides information about the training and testing samples for each class.

%%%%%%%%%%%%%%%%%%%%%%%%%%%%%%%%%%%%%%%%%%%%%%%%%

%%%%%%%%%%%%%%%%%%%%%%%%%%%%%%%%%%%%%%%%%%%%%%%%%

%%%%%%%%%%%%%%%%%%%%%%%%%%%%%%%%%%%%%%%%%%%%%%%%%

\begin{figure*}[!t]
% \begin{tabular}{ c c}
\centering
\begin{subfigure}{0.35\columnwidth}
    \centering
    \includegraphics[width=0.99\columnwidth]{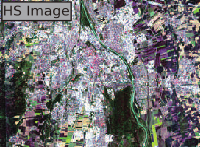}
	\caption{HS image}
% 	\label{Fig.4A}
\end{subfigure}
\begin{subfigure}{0.35\columnwidth}
    \centering
	\includegraphics[width=0.99\columnwidth]{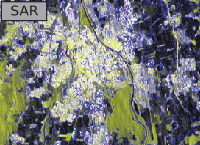}
    \caption{SAR Image} 
% 	\label{Fig.4B}
\end{subfigure}
\begin{subfigure}{0.35\columnwidth}
    \centering
	\includegraphics[width=0.99\columnwidth]{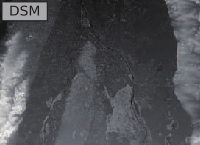}
    \caption{DSM Image} 
% 	\label{Fig.4B}
\end{subfigure}
\begin{subfigure}{0.37\columnwidth}
   \centering
   \includegraphics[width=0.99\columnwidth]{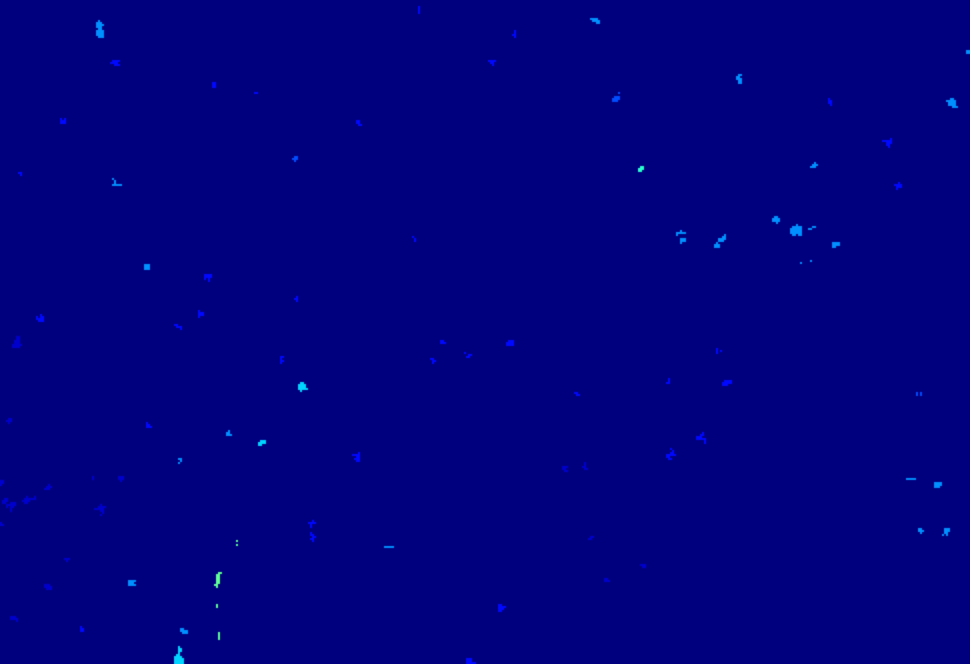}
	\caption{Training Image}
%   \label{Fig.4C}
\end{subfigure}
\begin{subfigure}{0.37\columnwidth}
   \centering
   \includegraphics[width=0.99\columnwidth]{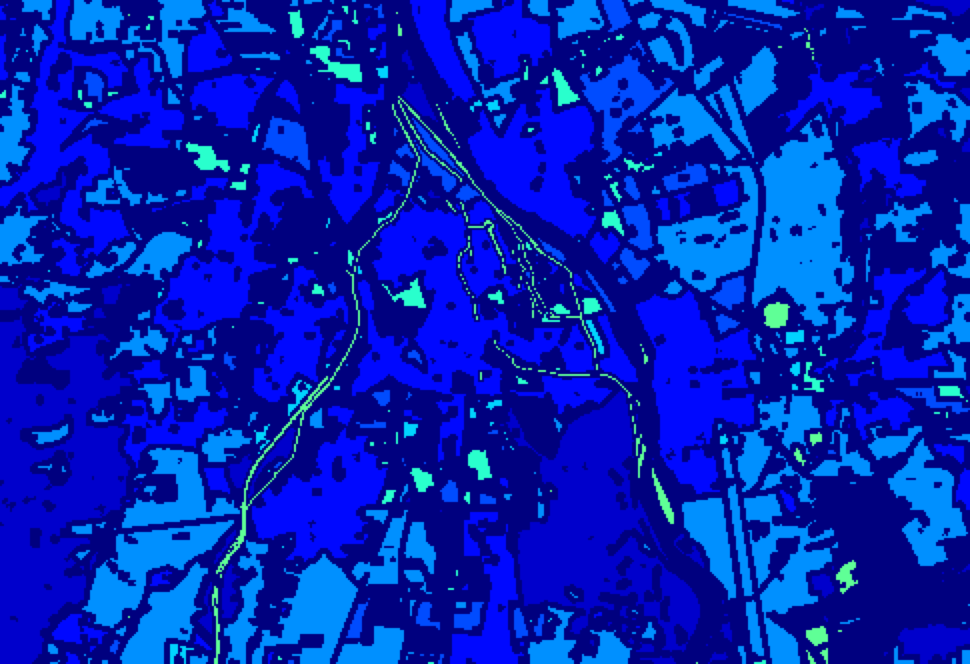}
	\caption{Test Image}
%    \label{Fig.4C}
\end{subfigure}
% \end{tabular}
\vspace{0.5cm}
\centering
\resizebox{0.8\linewidth}{!}{\begin{tabular}{cccc||cccc}\hline
    Color & Land cover & Train & Test & Color & Land cover & Train & Test \\ \hline
    \fboxsep=1mm \fboxrule=1mm
    
    \fcolorbox{Augsburg_data_Background!}{Augsburg_data_Background!}{\null} & Background & 68816   & 20497 & \fboxsep=1mm \fboxrule=1mm
    \fcolorbox{Augsburg_data_Forest!}{Augsburg_data_Forest!}{\null} & Forest & 1162   & 22084 \\
    \fboxsep=1mm \fboxrule=1mm
    
    \fcolorbox{Augsburg_data_Residential-Area!}{Augsburg_data_Residential-Area!}{\null} & Residential-Area & 344   & 6538 & \fboxsep=1mm \fboxrule=1mm
    \fcolorbox{Augsburg_data_Industrial-Area!}{Augsburg_data_Industrial-Area!}{\null} & Industrial-Area & 91   & 1735 \\
    \fboxsep=1mm \fboxrule=1mm
    
    \fcolorbox{Augsburg_data_Low-Plants!}{Augsburg_data_Low-Plants!}{\null} & Low-Plants & 334   & 6353 & \fboxsep=1mm \fboxrule=1mm
    \fcolorbox{Augsburg_data_Allotment!}{Augsburg_data_Allotment!}{\null} & Allotment & 23   & 443 \\
    \fboxsep=1mm \fboxrule=1mm
    
    \fcolorbox{Augsburg_data_Commercial-Area!}{Augsburg_data_Commercial-Area!}{\null} & Commercial-Area  & 112   & 2121 & \fboxsep=1mm \fboxrule=1mm
    \fcolorbox{Augsburg_data_Water!}{Augsburg_data_Water!}{\null} & Water & 312   & 5928\\\hline
    
    % &   Total samples & 71500   & 71500  \\\hline
    \end{tabular}}
    \caption{Visualization of the Augsburg scene. (a) True-color image for the HS image data over bands 40, 20, and 10, respectively (b) Grayscale image for the LiDAR data and (c) Ground truth of Augsburg scene. The table represents class-specific land-cover types and the number of randomly selected 5\% training and remaining 95\% test samples.}
\label{fig:Augsburg}
\end{figure*}

$\bullet$ There are three types of data in \textbf{Augsburg scene} which include an HSI, a dual-Pol SAR image, and a DSM image~\cite{hong2021multimodal}. SAR data are collected from the Sentinel-1 platform, while HS and DSM data are captured by DAS-EOC, DLR over the city of Augsburg, Germany. The collection is done by the HySpex sensor~\cite{baumgartner2012characterisation}, the Sentinel-1 sensor, and the DLR-3 K system~\cite{kurz2011real}, respectively. The spatial resolutions of all images are down-sampled to a unified spatial resolution of $30~m$ ground sampling distance (GSD) for adequately managing the multimodal fusion. For the HSI, there are $332\times 485$ pixels and $180$ spectral bands ranging between $0.4-2.5~\mu{m}$. The DSM image has a single band, whereas the SAR image has $4$ bands. The four bands indicate VV intensity, VH intensity, the real component, and the imaginary component of the PolSAR covariance matrix’s off-diagonal element. There are $15$ distinct land-cover classes in the ground truth. Fig. \ref{fig:Augsburg} shows detailed information on the train and test sets.

%%%%%%%%%%%%%%%%%%%%%%%%%%%%%%%%%%%%%%%%%%%%%%%%%
%%%%%%%%%%%%%%%%%%%%%%%%%%%%%%%%%%%%%%%%%%%%%%%%%

%%%%%%%%%%%%%%%%%%%%%%%%%%%%%%%%%%%%%%%%%%%%%%%%%

%%%%%%%%%%%%%%%%%%%%%%%%%%%%%%%%%%%%%%%%%%%%%%%%%

A class-specific summary of the UH, MUUFL, and Trento scenes is shown in Figs. \ref{fig:UH}, \ref{fig:MUUFL}, \ref{fig:Trento} and \ref{fig:Augsburg}, respectively. Each dataset includes its corresponding ground truth, the type associated with the land-cover classes, and the number of available labeled samples per class.

\subsection{Experimental Setup}
In order to study the effectiveness of the proposed multimodal fusion transformer model, we have performed extensive experiments and compared it with traditional as well as state-of-the-art methods. The compared methods include traditional classifiers i.e., KNN~\cite{rasti2020feature}, RF \cite{ahmad2021hyperspectral}, and SVM~\cite{melgani2004classification}, as well as classical CNN models i.e., CNN1D \cite{hong2021graph}, CNN2D \cite{makantasis2015deep}, CNN3D \cite{hamida2018deep} and RNN~\cite{cho2014properties}. We also considered a few state-of-the-art transformer models, i.e., ViT~\cite{dosovitskiy2020image} and SpectralFormer~\cite{hong2021spectralformer}. The experiments have been performed using both HSI data and other sources of multimodal data i.e., LiDAR, MSI, SAR and DSM, combined. \textcolor{black}{It can be noted that, for other compared models, the HSI and the other complementary data were concatenated band-wise before feeding the data to the models.}

\textbf{Configuration:} All the tests have been performed on a Red Hat Enterprise Server (Release $7.6$) with a CPU having \textbf{ppc64le} architecture and a total of $40$ cores with $4$ threads per core and $377$ GB of RAM. The GPU used is a single \textbf{Nvidia Tesla V100} with $32510$ MB of VRAM. The number of HSI patch tokens $(n)$ that is received from the tokenization operation is set to 4 in our experiments.

A \textbf{batch size} of $64$ and $500$ has been used for training  and testing the performance of the considered models, where patches of size $11\times{11}\times{B}$ are extracted from the HSI and $11\times{11}\times{C}$ from other sources of multimodal data. All the models except KNN, RF, SVM and RNN have been trained with \textbf{Adam} optimizer~\cite{adam, dubey2019diffgrad} with \textbf{learning rate} set to $5e^{-4}$ and \textbf{weight decay} of $5e^{-3}$. For the RNN, no weight decay was used and a higher learning rate of $1e^{-3}$ was adopted. These models (including the RNN) also utilized a \textbf{step scheduler} with step size $= 50$ and gamma $= 0.9$, while training has been conducted using $500$ \textbf{epochs}. Each experiment has been repeated 3 times and the average and standard deviations are reported. The source code of the proposed multimodal fusion transformer was implemented using PyTorch 1.5.0 and Python 3.7.7. The parameters and computational complexities for the considered models with respect to the dataset are shown in Fig. \ref{fig:complexity}.

%It can be clearly seen that using other multimodal data gives an overall improvement in performance.

\textbf{Evaluation Matrix:} Several widely used quantitative measures, such as overall accuracy (OA), average accuracy (AA) and statistical Kappa ($\kappa$) coefficients, are used to assess the performance of the proposed network and compare it with other methods. OA represents the proportion of correctly classified test samples versus all test samples, whereas AA represents the average of class-wise accuracy. The kappa value reflects the degree of agreement between the generated classification maps of the considered model and the provided ground truth.

\textcolor{black}{The experimental process has been conducted in three different settings, 1) Disjointed (spatially and spectrally) training and test samples, i.e., the intersection between the training samples and testing samples remains empty. 2) Different percentages of training samples have been randomly selected and used to validate the performance of the proposed network. 3) Different variations of the proposed model have been compared on the same disjointed datasets.}

%%%%%%%%%%%%%%%%%%%%%%%%%%%%%%%%%%%%%%%%%%%%%%%%%
\begin{table*}[!t]
\centering
\caption{OA, AA and Kappa values on the University of Houston dataset (in \%) by considering HSI data only.}
\resizebox{\linewidth}{!}{\color{black}
% Please add the following required packages to your document preamble:
% \usepackage{multirow}
\begin{tabular}{c||ccc|cccc||ccc}
\hline 
\multirow{2}{*}{\begin{tabular}[c]{@{}c@{}}\textbf{Class}\\ \textbf{No.}\end{tabular}} 
& \multicolumn{3}{c|}{\textbf{Conventional Classifiers}}            
& \multicolumn{4}{c||}{\textbf{Classical Convolutional Networks}}
& \multicolumn{3}{c}{\textbf{Transformer Networks}}  \\ \cline{2-11}
& \multicolumn{1}{c|}{\textbf{KNN}} & \multicolumn{1}{c|}{\textbf{RF}} & \textbf{SVM} 
& \multicolumn{1}{c|}{\textbf{CNN1D}} 
& \multicolumn{1}{c|}{\textbf{CNN2D}} 
& \multicolumn{1}{l|}{\textbf{CNN3D}} 
& \textbf{RNN} 
& \multicolumn{1}{c|}{\textbf{ViT}} & \multicolumn{1}{c|}{\textbf{SpectralFormer}} 
& \textbf{MFT} \\ \hline \hline
1 &77.87    &82.81 $ \pm $ 00.08    &79.77    &81.10 $ \pm $ 00.43    &80.53 $ \pm $ 00.23    &81.70 $ \pm $ 00.38    &80.22 $ \pm $ 02.80    &82.40 $ \pm $ 00.39    &82.49 $ \pm $ 00.25    &\textbf{82.53 $ \pm $ 00.13}    \\ %\hline
2 &77.44    &82.86 $ \pm $ 00.36    &82.42    &80.23 $ \pm $ 01.51    &83.90 $ \pm $ 00.09    &80.55 $ \pm $ 01.84    &78.51 $ \pm $ 00.19    &80.29 $ \pm $ 00.79    &\textbf{89.13 $ \pm $ 06.36}    &85.03 $ \pm $ 00.04    \\
3 &96.83    &63.10 $ \pm $ 01.71    &59.41    &53.73 $ \pm $ 00.09    &57.49 $ \pm $ 02.10    &96.57 $ \pm $ 00.09    &52.94 $ \pm $ 17.72    &97.43 $ \pm $ 01.32    &69.77 $ \pm $ 12.29    &\textbf{98.55 $ \pm $ 01.08}    \\
4 &75.28    &91.95 $ \pm $ 00.15    &83.81    &83.74 $ \pm $ 00.71    &89.46 $ \pm $ 00.24    &78.54 $ \pm $ 02.82    &83.81 $ \pm $ 07.71    &90.40 $ \pm $ 00.25    &91.73 $ \pm $ 03.42    &\textbf{94.60 $ \pm $ 02.53}    \\
5 &90.72    &99.78 $ \pm $ 00.12    &95.27    &87.06 $ \pm $ 01.51    &92.36 $ \pm $ 01.20    &98.48 $ \pm $ 00.28    &85.61 $ \pm $ 09.73    &99.24 $ \pm $ 00.13    &96.78 $ \pm $ 00.76    &\textbf{99.84 $ \pm $ 00.16}    \\
6 &66.43    &96.97 $ \pm $ 00.33    &67.13    &52.45 $ \pm $ 01.14    &64.10 $ \pm $ 02.38    &73.89 $ \pm $ 00.66    &70.16 $ \pm $ 07.18    &91.38 $ \pm $ 00.87    &85.31 $ \pm $ 07.99    &\textbf{95.10 $ \pm $ 00.57}    \\
7 &76.96    &85.23 $ \pm $ 00.50    &83.21    &71.42 $ \pm $ 01.46    &71.39 $ \pm $ 04.04    &82.77 $ \pm $ 01.45    &73.01 $ \pm $ 01.31    &\textbf{86.10 $ \pm $ 01.54}    &80.25 $ \pm $ 02.23    &85.79 $ \pm $ 00.94    \\
8 &30.96    &42.58 $ \pm $ 00.36    &29.53    &41.12 $ \pm $ 01.70    &44.95 $ \pm $ 06.09    &38.30 $ \pm $ 01.09    &43.84 $ \pm $ 08.75    &73.95 $ \pm $ 00.29    &62.74 $ \pm $ 11.68    &\textbf{79.87 $ \pm $ 09.29}    \\
9 &69.50    &85.36 $ \pm $ 00.13    &75.45    &60.25 $ \pm $ 00.08    &62.45 $ \pm $ 02.04    &65.94 $ \pm $ 02.86    &68.84 $ \pm $ 02.54    &85.33 $ \pm $ 02.24    &70.57 $ \pm $ 01.62    &\textbf{89.05 $ \pm $ 02.47}    \\
10 &42.95    &35.81 $ \pm $ 00.70    &46.62    &39.12 $ \pm $ 02.03    &49.94 $ \pm $ 01.66    &43.28 $ \pm $ 07.46    &37.52 $ \pm $ 01.15    &50.42 $ \pm $ 05.59    &48.17 $ \pm $ 05.04    &\textbf{62.45 $ \pm $ 02.95}    \\
11 &56.17    &63.03 $ \pm $ 00.39    &45.07    &42.06 $ \pm $ 00.78    &44.53 $ \pm $ 00.78    &33.59 $ \pm $ 02.72    &49.65 $ \pm $ 09.61    &80.80 $ \pm $ 03.06    &62.75 $ \pm $ 04.12    &\textbf{98.04 $ \pm $ 01.45}    \\
12 &75.79    &66.63 $ \pm $ 00.50    &70.03    &62.98 $ \pm $ 04.58    &53.92 $ \pm $ 06.25    &67.85 $ \pm $ 02.10    &64.07 $ \pm $ 00.28    &81.91 $ \pm $ 01.77    &79.09 $ \pm $ 02.16    &\textbf{95.84 $ \pm $ 00.48}    \\
13 &60.35    &87.60 $ \pm $ 00.17    &68.42    &42.11 $ \pm $ 01.52    &47.13 $ \pm $ 02.33    &77.54 $ \pm $ 01.98    &53.92 $ \pm $ 08.72    &89.47 $ \pm $ 01.74    &63.63 $ \pm $ 05.56    &\textbf{91.81 $ \pm $ 00.33}    \\
14 &76.92    &99.73 $ \pm $ 00.19    &75.30    &83.94 $ \pm $ 00.38    &82.46 $ \pm $ 03.07    &92.58 $ \pm $ 01.06    &81.38 $ \pm $ 12.63    &99.33 $ \pm $ 00.95    &93.66 $ \pm $ 04.30    &\textbf{99.73 $ \pm $ 00.19}    \\
15 &88.37    &85.62 $ \pm $ 01.38    &49.89    &34.46 $ \pm $ 02.04    &42.92 $ \pm $ 05.11    &93.52 $ \pm $ 01.55    &44.12 $ \pm $ 13.23    &\textbf{99.72 $ \pm $ 00.26}    &77.24 $ \pm $ 07.34    &92.60 $ \pm $ 02.42    \\ \hline \hline
OA   &69.48    &74.87 $ \pm $ 00.09    &68.13    &63.04 $ \pm $ 00.08    &65.85 $ \pm $ 00.09    &70.26 $ \pm $ 00.14    &65.20 $ \pm $ 02.74    &83.23 $ \pm $ 00.47    &76.35 $ \pm $ 01.97    &\textbf{88.45 $ \pm $ 00.28}    \\ %\hline
AA  &70.84    &77.94 $ \pm $ 00.14    &67.42    &61.05 $ \pm $ 00.10    &64.50 $ \pm $ 00.11    &73.67 $ \pm $ 00.17    &64.51 $ \pm $ 02.77    &85.88 $ \pm $ 00.33    &76.89 $ \pm $ 02.31    &\textbf{90.05 $ \pm $ 00.29}    \\ %\hline
$\kappa (\times 100)$   &67.08    &72.93 $ \pm $ 00.09    &65.56    &60.01 $ \pm $ 00.08    &63.04 $ \pm $ 00.09    &67.91 $ \pm $ 00.16    &62.43 $ \pm $ 02.87    &81.88 $ \pm $ 00.51    &74.42 $ \pm $ 02.14    &\textbf{87.46 $ \pm $ 00.30}    \\ \hline
\end{tabular}}
\label{tab:UH_HS}
\end{table*}

%%%%%%%%%%%%%%%%%%%%%%%%%%%%%%%%%%%%%%%%%%%%%%%%%

%%%%%%%%%%%%%%%%%%%%%%%%%%%%%%%%%%%%%%%%%%%%%%%%%
\begin{table*}[!t]
\centering
\caption{OA, AA and Kappa values on the University of Houston dataset (in \%) by considering HSI and LiDAR data.}
\resizebox{\linewidth}{!}{\color{black}
% Please add the following required packages to your document preamble:
% \usepackage{multirow}
\begin{tabular}{c||ccc|cccc||ccc}
\hline 
\multirow{2}{*}{\begin{tabular}[c]{@{}c@{}}\textbf{Class}\\ \textbf{No.}\end{tabular}} 
& \multicolumn{3}{c|}{\textbf{Conventional Classifiers}}            
& \multicolumn{4}{c||}{\textbf{Classical Convolutional Networks}}
& \multicolumn{3}{c}{\textbf{Transformer Networks}}  \\ \cline{2-11}
& \multicolumn{1}{c|}{\textbf{KNN}} & \multicolumn{1}{c|}{\textbf{RF}} & \textbf{SVM} 
& \multicolumn{1}{c|}{\textbf{CNN1D}} 
& \multicolumn{1}{c|}{\textbf{CNN2D}} 
& \multicolumn{1}{l|}{\textbf{CNN3D}} 
& \textbf{RNN} 
& \multicolumn{1}{c|}{\textbf{ViT}} & \multicolumn{1}{c|}{\textbf{SpectralFormer}} 
& \textbf{MFT} \\ \hline \hline
1 &77.30    &79.33 $ \pm $ 00.38    &79.96    &81.32 $ \pm $ 00.16    &82.62 $ \pm $ 00.08    &82.30 $ \pm $ 00.29    &81.80 $ \pm $ 00.62    &82.59 $ \pm $ 00.18    &\textbf{82.65 $ \pm $ 00.24}    &82.34 $ \pm $ 00.47    \\ %\hline
2 &81.58    &71.49 $ \pm $ 01.69    &82.89    &81.92 $ \pm $ 00.19    &81.83 $ \pm $ 00.66    &78.70 $ \pm $ 00.98    &71.40 $ \pm $ 05.21    &82.33 $ \pm $ 01.46    &83.33 $ \pm $ 00.29    &\textbf{88.78 $ \pm $ 05.27}    \\
3 &97.82    &98.15 $ \pm $ 00.25    &60.79    &57.49 $ \pm $ 00.61    &62.90 $ \pm $ 00.37    &96.96 $ \pm $ 00.47    &76.04 $ \pm $ 14.49    &97.43 $ \pm $ 00.70    &75.78 $ \pm $ 12.48    &\textbf{98.15 $ \pm $ 01.08}    \\
4 &80.59    &78.09 $ \pm $ 00.47    &86.65    &86.74 $ \pm $ 00.23    &88.57 $ \pm $ 00.04    &80.49 $ \pm $ 01.57    &88.51 $ \pm $ 02.14    &92.93 $ \pm $ 01.72    &91.10 $ \pm $ 01.29    &\textbf{94.35 $ \pm $ 02.40}    \\
5 &91.86    &89.68 $ \pm $ 00.08    &95.36    &95.93 $ \pm $ 00.27    &97.19 $ \pm $ 00.19    &98.11 $ \pm $ 00.54    &85.76 $ \pm $ 04.87    &\textbf{99.84 $ \pm $ 00.04}    &98.30 $ \pm $ 01.12    &99.12 $ \pm $ 01.25    \\
6 &78.32    &93.71 $ \pm $ 00.57    &69.23    &58.04 $ \pm $ 00.57    &65.03 $ \pm $ 01.98    &73.89 $ \pm $ 01.84    &85.78 $ \pm $ 02.87    &84.15 $ \pm $ 03.30    &89.04 $ \pm $ 02.70    &\textbf{99.30 $ \pm $ 00.00}    \\
7 &80.97    &79.32 $ \pm $ 01.03    &85.26    &77.95 $ \pm $ 00.88    &80.22 $ \pm $ 01.42    &81.09 $ \pm $ 00.75    &82.77 $ \pm $ 01.71    &87.84 $ \pm $ 01.49    &81.72 $ \pm $ 01.19    &\textbf{88.56 $ \pm $ 01.16}    \\
8 &42.45    &53.37 $ \pm $ 02.26    &49.38    &54.42 $ \pm $ 02.33    &60.81 $ \pm $ 04.32    &44.63 $ \pm $ 06.52    &61.44 $ \pm $ 07.10    &79.93 $ \pm $ 00.16    &67.81 $ \pm $ 09.00    &\textbf{86.89 $ \pm $ 05.33}    \\
9 &69.22    &76.27 $ \pm $ 00.46    &78.28    &66.13 $ \pm $ 00.19    &67.74 $ \pm $ 01.08    &74.76 $ \pm $ 03.84    &67.42 $ \pm $ 07.89    &82.94 $ \pm $ 00.85    &74.47 $ \pm $ 01.89    &\textbf{87.91 $ \pm $ 03.90}    \\
10 &43.44    &38.42 $ \pm $ 01.07    &50.48    &47.30 $ \pm $ 02.18    &51.74 $ \pm $ 01.33    &37.52 $ \pm $ 11.23    &38.45 $ \pm $ 03.12    &52.93 $ \pm $ 05.14    &56.76 $ \pm $ 06.31    &\textbf{64.70 $ \pm $ 00.92}    \\
11 &65.84    &65.02 $ \pm $ 00.47    &49.34    &44.40 $ \pm $ 02.06    &39.91 $ \pm $ 03.52    &40.80 $ \pm $ 08.50    &64.39 $ \pm $ 06.38    &80.99 $ \pm $ 03.06    &59.93 $ \pm $ 07.67    &\textbf{98.64 $ \pm $ 00.70}    \\
12 &81.94    &82.20 $ \pm $ 00.56    &72.05    &63.66 $ \pm $ 08.15    &82.20 $ \pm $ 03.85    &66.38 $ \pm $ 02.36    &77.07 $ \pm $ 05.79    &91.07 $ \pm $ 02.55    &70.00 $ \pm $ 02.98    &\textbf{94.24 $ \pm $ 02.19}    \\
13 &60.35    &67.72 $ \pm $ 00.50    &77.19    &50.41 $ \pm $ 01.86    &52.40 $ \pm $ 01.29    &68.77 $ \pm $ 12.16    &47.13 $ \pm $ 03.73    &87.84 $ \pm $ 01.91    &66.20 $ \pm $ 00.44    &\textbf{90.29 $ \pm $ 00.66}    \\
14 &83.81    &82.46 $ \pm $ 00.83    &75.30    &86.50 $ \pm $ 00.69    &92.44 $ \pm $ 02.89    &92.85 $ \pm $ 01.16    &97.98 $ \pm $ 00.87    &\textbf{100.0 $ \pm $ 00.00}    &92.04 $ \pm $ 04.50    &99.73 $ \pm $ 00.38    \\
15 &91.33    &95.56 $ \pm $ 00.30    &55.18    &40.10 $ \pm $ 03.20    &52.08 $ \pm $ 04.14    &96.90 $ \pm $ 01.41    &73.50 $ \pm $ 09.82    &\textbf{99.65 $ \pm $ 00.50}    &77.45 $ \pm $ 13.31    &99.58 $ \pm $ 00.17    \\ \hline \hline
OA   &73.50    &73.83 $ \pm $ 00.31    &71.94    &68.11 $ \pm $ 00.29    &71.88 $ \pm $ 00.36    &71.41 $ \pm $ 00.35    &72.31 $ \pm $ 02.44    &85.05 $ \pm $ 00.47    &76.87 $ \pm $ 02.21    &\textbf{89.80 $ \pm $ 00.53}    \\ %\hline
AA  &75.12    &76.72 $ \pm $ 00.28    &71.16    &66.15 $ \pm $ 00.09    &70.51 $ \pm $ 00.27    &74.28 $ \pm $ 00.30    &73.30 $ \pm $ 02.50    &86.83 $ \pm $ 00.53    &77.77 $ \pm $ 02.78    &\textbf{91.51 $ \pm $ 00.40}    \\ %\hline
$\kappa (\times 100)$   &71.41    &71.78 $ \pm $ 00.33    &69.67    &65.48 $ \pm $ 00.31    &69.56 $ \pm $ 00.39    &69.12 $ \pm $ 00.36    &70.14 $ \pm $ 02.62    &83.84 $ \pm $ 00.51    &75.03 $ \pm $ 02.37    &\textbf{88.93 $ \pm $ 00.59}    \\ \hline
\end{tabular}}
\label{tab:UH_HL}
\end{table*}

%%%%%%%%%%%%%%%%%%%%%%%%%%%%%%%%%%%%%%%%%%%%%%%%%

%%%%%%%%%%%%%%%%%%%%%%%%%%%%%%%%%%%%%%%%%%%%%%%%%
\begin{figure*}[!t]
\centering
	\begin{subfigure}{0.32\textwidth}
		\includegraphics[width=0.99\textwidth]{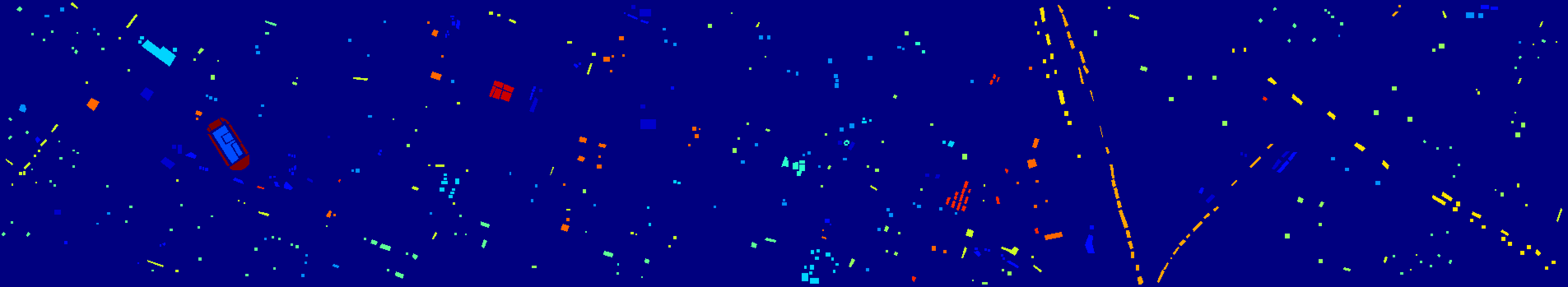}
		\caption{GT}
		% \label{Fig6A}
	\end{subfigure}
	\begin{subfigure}{0.32\textwidth}
		\includegraphics[width=0.99\textwidth]{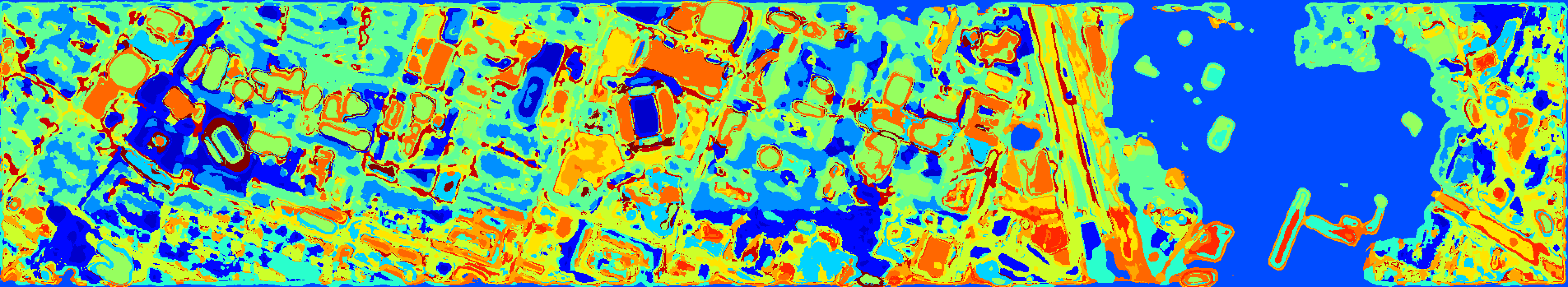}
		\caption{KNN}
		% \label{Fig6B}
	\end{subfigure}
	\begin{subfigure}{0.32\textwidth}
		\includegraphics[width=0.99\textwidth]{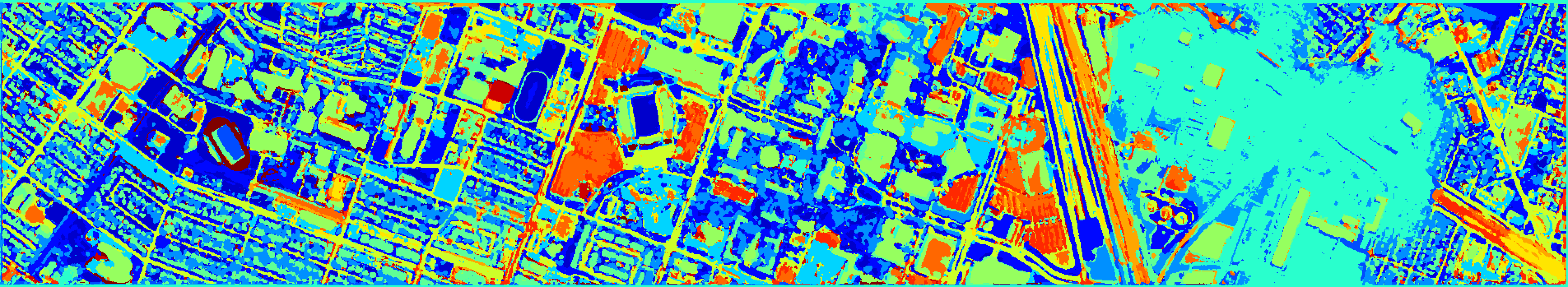}
		\caption{RF}
		% \label{Fig6C}
	\end{subfigure}
	\begin{subfigure}{0.32\textwidth}
		\includegraphics[width=0.99\textwidth]{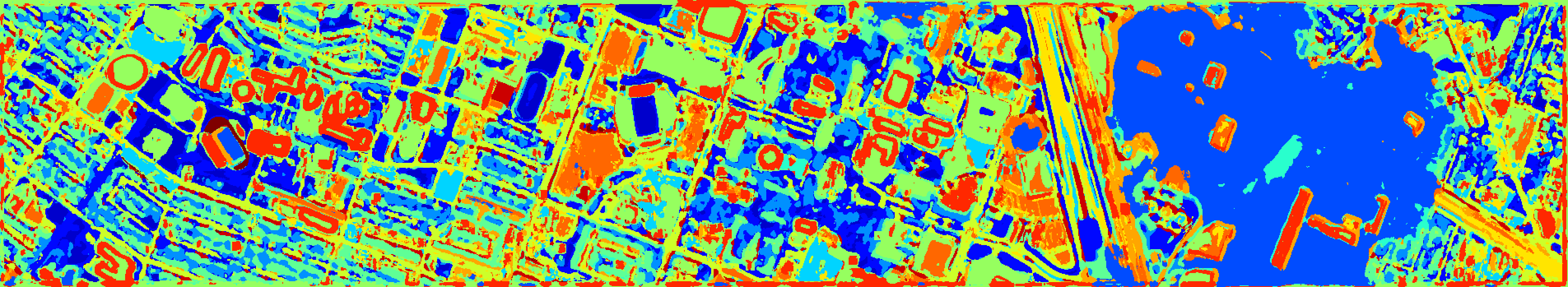}
		\caption{SVM} 
		% \label{Fig6D}
	\end{subfigure}
	\begin{subfigure}{0.32\textwidth}
		\includegraphics[width=0.99\textwidth]{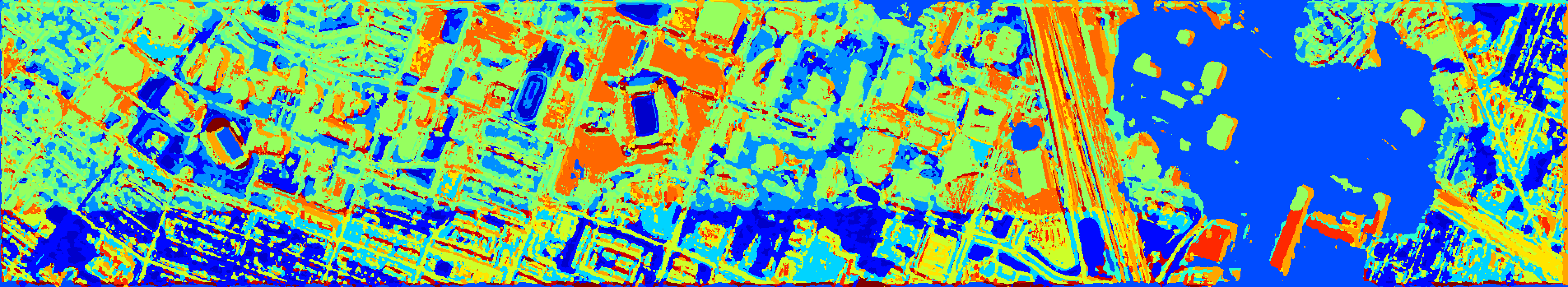}
		\caption{CNN1D} 
		% \label{Fig6E}
	\end{subfigure}
	\begin{subfigure}{0.32\textwidth}
		\includegraphics[width=0.99\textwidth]{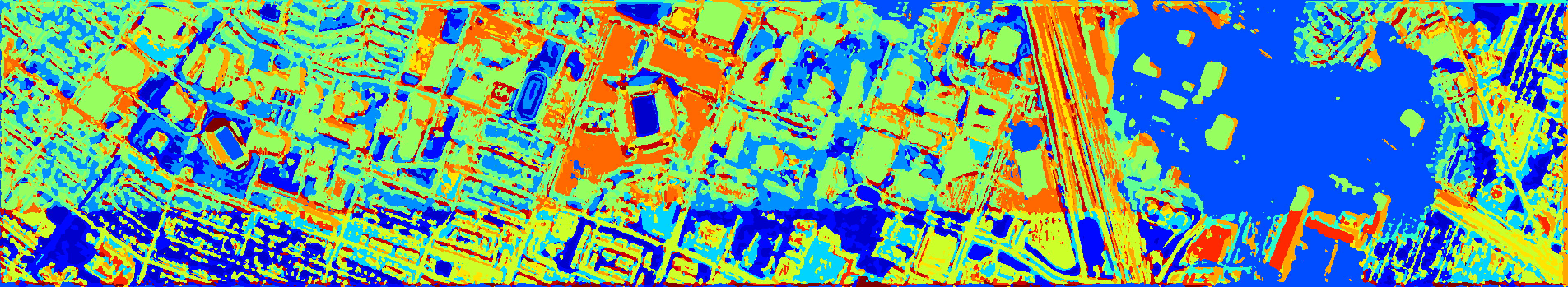}
		\caption{CNN2D}
		% \label{Fig6F}
	\end{subfigure}
    \begin{subfigure}{0.32\textwidth}
		\includegraphics[width=0.99\textwidth]{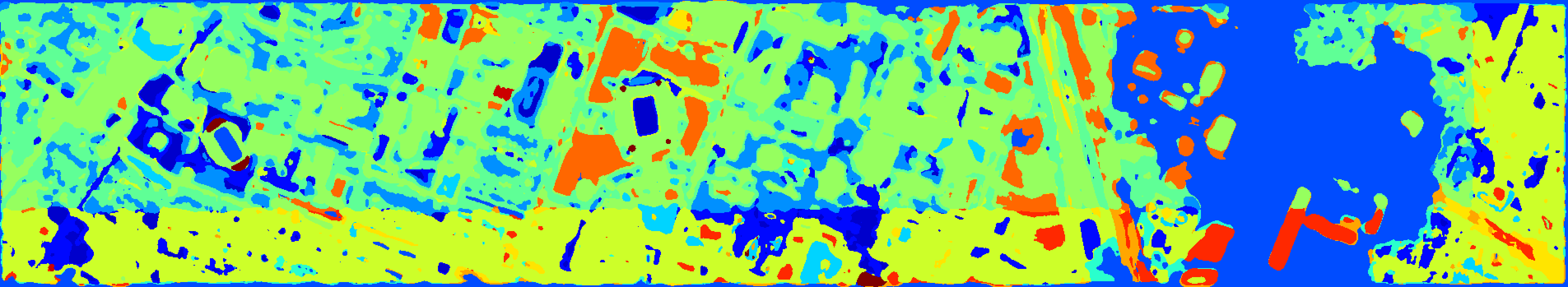}
		\caption{CNN3D}
		% \label{Fig6G}
	\end{subfigure}
	\begin{subfigure}{0.32\textwidth}
		\includegraphics[width=0.99\textwidth]{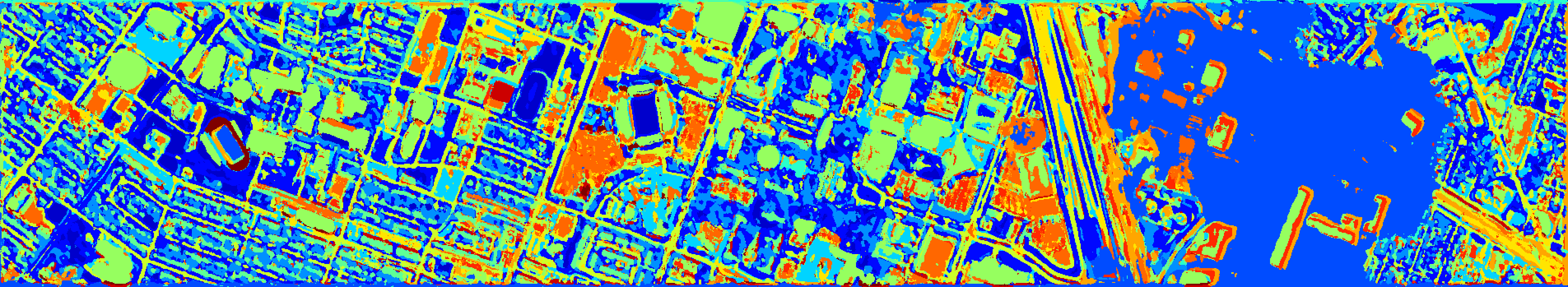}
		\caption{RNN}
		% \label{Fig6H}
	\end{subfigure}
	\begin{subfigure}{0.32\textwidth}
		\includegraphics[width=0.99\textwidth]{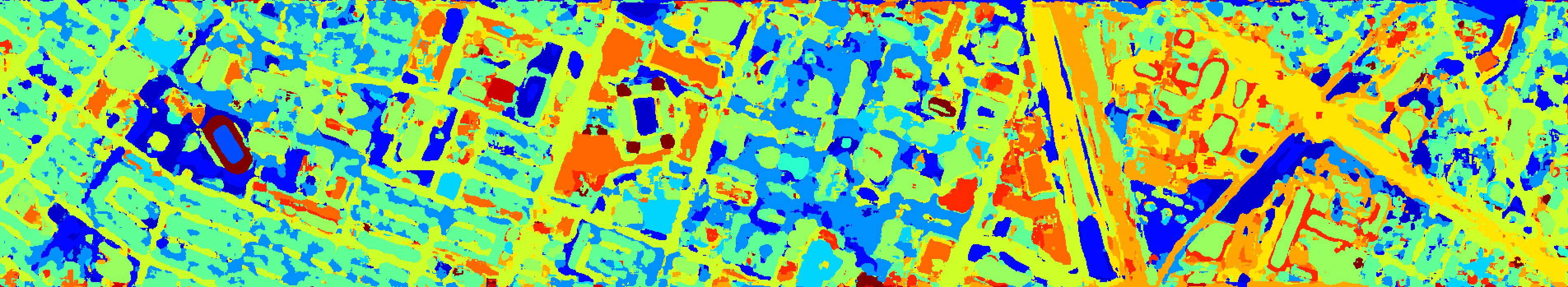}
		\caption{ViT} 
		% \label{Fig6I}
	\end{subfigure}
	\begin{subfigure}{0.32\textwidth}
		\includegraphics[width=0.99\textwidth]{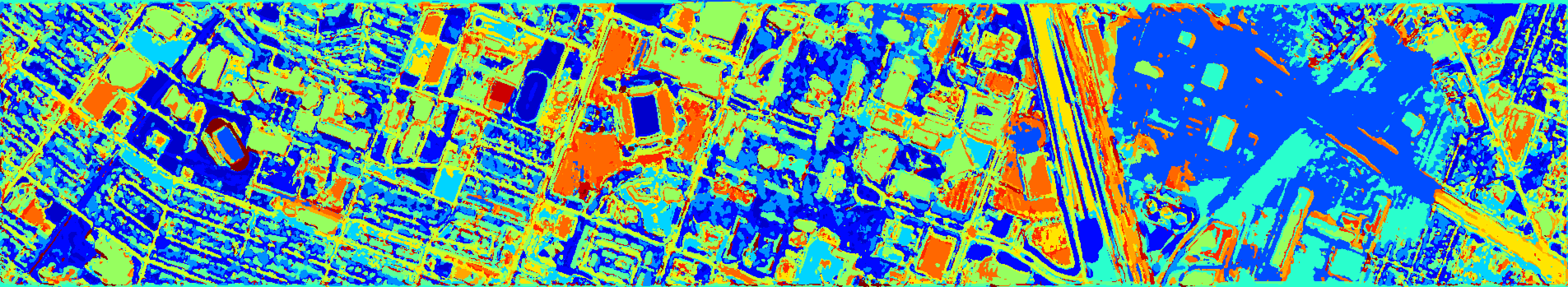}
		\caption{SpectralFormer}
		% \label{Fig6J}
	\end{subfigure}
	\begin{subfigure}{0.32\textwidth}
		\includegraphics[width=0.99\textwidth]{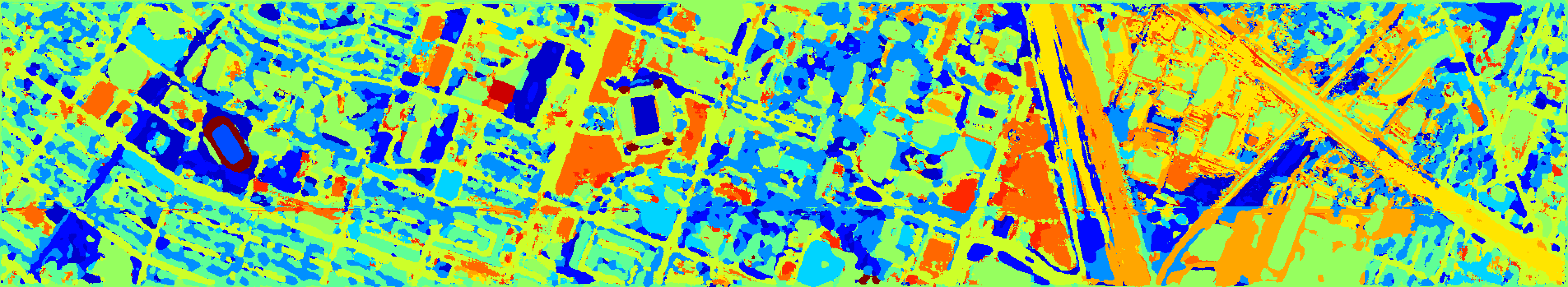}
		\caption{MFT}
		% \label{Fig6K}
	\end{subfigure}
\caption{(a) Ground truth and classification maps obtained for the UH data set by: (b) KNN, (c) RF, (d) CNN1D, (e) CNN2D, (f) CNN3D, (g) RNN, (h) ViT, (i) SpectralFormer, and (j) MFT using disjoint training samples.}
\label{fig:comparativeUH(H+L)}
\end{figure*}

%%%%%%%%%%%%%%%%%%%%%%%%%%%%%%%%%%%%%%%%%%%%%%%%%

%%%%%%%%%%%%%%%%%%%%%%%%%%%%%%%%%%%%%%%%%%%%%%%%%
\begin{table*}[!t]
\centering
\caption{OA, AA and Kappa values on the University of Houston dataset (in \%) by considering HSI and MS image data.}
\resizebox{\linewidth}{!}{\color{black}
% Please add the following required packages to your document preamble:
% \usepackage{multirow}
\begin{tabular}{c||ccc|cccc||ccc}
\hline 
\multirow{2}{*}{\begin{tabular}[c]{@{}c@{}}\textbf{Class}\\ \textbf{No.}\end{tabular}} 
& \multicolumn{3}{c|}{\textbf{Conventional Classifiers}}            
& \multicolumn{4}{c||}{\textbf{Classical Convolutional Networks}}
& \multicolumn{3}{c}{\textbf{Transformer Networks}}  \\ \cline{2-11}
& \multicolumn{1}{c|}{\textbf{KNN}} & \multicolumn{1}{c|}{\textbf{RF}} & \textbf{SVM} 
& \multicolumn{1}{c|}{\textbf{CNN1D}} 
& \multicolumn{1}{c|}{\textbf{CNN2D}} 
& \multicolumn{1}{l|}{\textbf{CNN3D}} 
& \textbf{RNN} 
& \multicolumn{1}{c|}{\textbf{ViT}} & \multicolumn{1}{c|}{\textbf{SpectralFormer}} 
& \textbf{MFT} \\ \hline \hline
1 &78.06    &78.82 $ \pm $ 00.40    &79.87    &84.52 $ \pm $ 02.26    &88.86 $ \pm $ 04.90    &82.15 $ \pm $ 00.68    &79.23 $ \pm $ 00.39    &\textbf{89.02 $ \pm $ 05.03}    &81.54 $ \pm $ 00.65    &82.72 $ \pm $ 00.41    \\ %\hline
2 &75.28    &72.65 $ \pm $ 02.53    &79.14    &81.05 $ \pm $ 02.70    &\textbf{86.18 $ \pm $ 05.21}    &84.62 $ \pm $ 00.04    &81.42 $ \pm $ 00.51    &82.11 $ \pm $ 00.67    &80.20 $ \pm $ 02.86    &85.09 $ \pm $ 00.09    \\ 
3 &96.63    &96.63 $ \pm $ 00.49    &61.25    &38.75 $ \pm $ 04.96    &45.35 $ \pm $ 03.93    &49.70 $ \pm $ 04.53    &38.75 $ \pm $ 01.22    &\textbf{99.21 $ \pm $ 00.16}    &59.27 $ \pm $ 11.24    &98.55 $ \pm $ 00.76    \\ 
4 &68.47    &77.15 $ \pm $ 00.27    &85.23    &81.09 $ \pm $ 03.63    &75.47 $ \pm $ 06.68    &86.68 $ \pm $ 00.50    &88.35 $ \pm $ 01.54    &81.72 $ \pm $ 01.08    &90.59 $ \pm $ 00.31    &\textbf{95.99 $ \pm $ 02.39}    \\ 
5 &89.77    &90.03 $ \pm $ 00.89    &87.22    &72.92 $ \pm $ 04.99    &87.12 $ \pm $ 04.46    &92.20 $ \pm $ 03.75    &90.91 $ \pm $ 01.01    &93.15 $ \pm $ 00.97    &95.01 $ \pm $ 00.96    &\textbf{99.78 $ \pm $ 00.09}    \\ 
6 &71.33    &86.48 $ \pm $ 06.49    &61.54    &63.64 $ \pm $ 02.97    &72.03 $ \pm $ 04.67    &79.95 $ \pm $ 02.87    &77.39 $ \pm $ 02.87    &\textbf{99.77 $ \pm $ 00.33}    &73.43 $ \pm $ 02.06    &97.20 $ \pm $ 01.98    \\ 
7 &72.57    &77.89 $ \pm $ 03.73    &81.53    &58.18 $ \pm $ 01.21    &68.13 $ \pm $ 05.26    &78.17 $ \pm $ 04.71    &65.76 $ \pm $ 03.53    &72.45 $ \pm $ 05.50    &78.42 $ \pm $ 00.62    &\textbf{86.32 $ \pm $ 00.69}    \\
8 &30.10    &41.63 $ \pm $ 00.25    &18.80    &26.34 $ \pm $ 16.50    &23.36 $ \pm $ 15.68    &27.95 $ \pm $ 06.36    &35.61 $ \pm $ 01.58    &77.87 $ \pm $ 02.72    &48.40 $ \pm $ 01.57    &\textbf{81.16 $ \pm $ 08.04}    \\ 
9 &68.18    &74.66 $ \pm $ 00.40    &68.27    &54.93 $ \pm $ 06.23    &59.05 $ \pm $ 04.45    &83.41 $ \pm $ 00.76    &72.05 $ \pm $ 03.16    &83.47 $ \pm $ 07.68    &73.25 $ \pm $ 01.69    &\textbf{87.76 $ \pm $ 00.74}    \\ 
10 &42.28    &38.42 $ \pm $ 00.21    &48.46    &28.73 $ \pm $ 04.69    &32.37 $ \pm $ 01.22    &69.76 $ \pm $ 04.74    &34.04 $ \pm $ 05.16    &51.03 $ \pm $ 01.04    &50.87 $ \pm $ 01.76    &\textbf{74.71 $ \pm $ 15.98}    \\ 
11 &53.80    &58.16 $ \pm $ 00.54    &38.71    &42.50 $ \pm $ 01.88    &43.33 $ \pm $ 01.90    &60.25 $ \pm $ 04.92    &36.78 $ \pm $ 02.50    &76.03 $ \pm $ 01.53    &55.03 $ \pm $ 00.43    &\textbf{93.71 $ \pm $ 06.01}    \\ 
12 &68.68    &80.24 $ \pm $ 00.09    &57.54    &43.84 $ \pm $ 13.82    &35.57 $ \pm $ 11.98    &65.10 $ \pm $ 15.37    &58.44 $ \pm $ 03.47    &75.47 $ \pm $ 02.24    &71.41 $ \pm $ 03.96    &\textbf{96.16 $ \pm $ 00.87}    \\ 
13 &57.89    &66.08 $ \pm $ 01.47    &65.96    &48.65 $ \pm $ 04.63    &52.28 $ \pm $ 04.22    &47.49 $ \pm $ 10.75    &52.63 $ \pm $ 06.82    &83.63 $ \pm $ 04.12    &57.08 $ \pm $ 00.17    &\textbf{92.51 $ \pm $ 00.17}    \\
14 &79.35    &80.30 $ \pm $ 01.34    &83.40    &63.16 $ \pm $ 02.17    &70.58 $ \pm $ 12.48    &96.22 $ \pm $ 03.34    &81.24 $ \pm $ 07.20    &99.60 $ \pm $ 00.33    &88.26 $ \pm $ 08.02    &\textbf{100.0 $ \pm $ 00.00}    \\ 
15 &86.68    &\textbf{94.15 $ \pm $ 00.20}    &48.41    &55.60 $ \pm $ 02.85    &45.24 $ \pm $ 09.15    &58.42 $ \pm $ 08.47    &40.94 $ \pm $ 15.84    &91.97 $ \pm $ 01.70    &46.51 $ \pm $ 05.61    &86.82 $ \pm $ 08.99    \\ \hline \hline
OA   &67.16    &71.51 $ \pm $ 00.46    &64.18    &56.63 $ \pm $ 00.61    &59.06 $ \pm $ 00.57    &71.49 $ \pm $ 01.07    &62.61 $ \pm $ 01.80    &80.49 $ \pm $ 01.19    &70.95 $ \pm $ 00.58    &\textbf{89.15 $ \pm $ 00.96}    \\ %\hline
AA  &69.27    &74.22 $ \pm $ 00.06    &64.36    &56.26 $ \pm $ 00.93    &58.99 $ \pm $ 00.25    &70.80 $ \pm $ 01.16    &62.24 $ \pm $ 02.51    &83.77 $ \pm $ 01.15    &69.95 $ \pm $ 00.63    &\textbf{90.56 $ \pm $ 00.93}    \\ %\hline
$\kappa (\times 100)$   &64.61    &69.30 $ \pm $ 00.42    &61.30    &53.07 $ \pm $ 00.68    &55.70 $ \pm $ 00.60    &69.02 $ \pm $ 01.16    &59.64 $ \pm $ 01.98    &78.88 $ \pm $ 01.29    &68.59 $ \pm $ 00.64    &\textbf{88.22 $ \pm $ 01.04}    \\ \hline
\end{tabular}}
\label{tab:UH_HSMS}
\end{table*}

%%%%%%%%%%%%%%%%%%%%%%%%%%%%%%%%%%%%%%%%%%%%%%%%%

%%%%%%%%%%%%%%%%%%%%%%%%%%%%%%%%%%%%%%%%%%%%%%%%%
\begin{figure*}[!t]
\centering
	\begin{subfigure}{0.32\textwidth}
		\includegraphics[width=0.99\textwidth]{clsfmaps/UH_H+L/Houston_gt_.png}
		\caption{GT}
		% \label{Fig6A}
	\end{subfigure}
	\begin{subfigure}{0.32\textwidth}
		\includegraphics[width=0.99\textwidth]{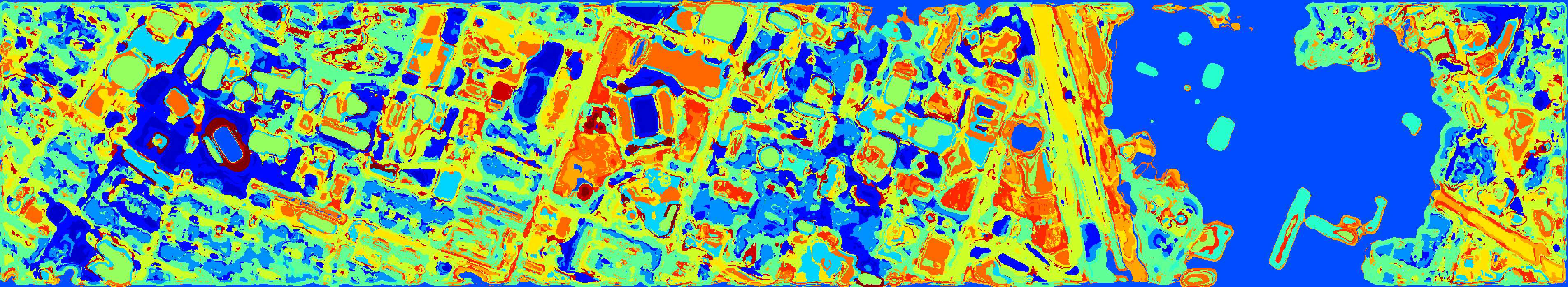}
		\caption{KNN}
		% \label{Fig6A}
	\end{subfigure}
	\begin{subfigure}{0.32\textwidth}
		\includegraphics[width=0.99\textwidth]{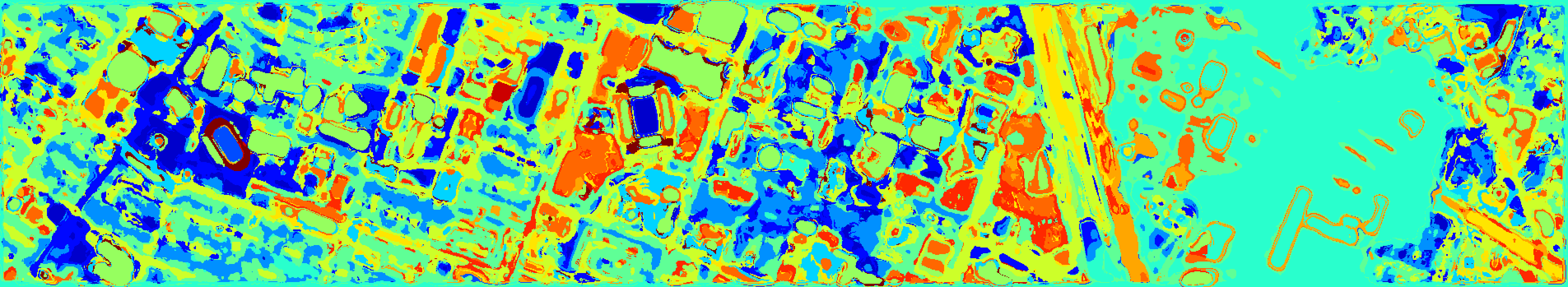}
		\caption{RF}
		% \label{Fig6B}
	\end{subfigure}
	\begin{subfigure}{0.32\textwidth}
		\includegraphics[width=0.99\textwidth]{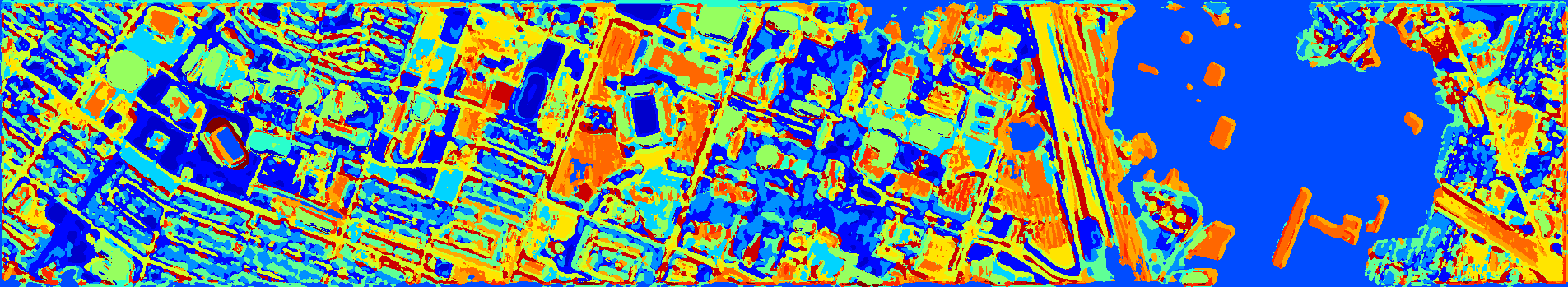}
		\caption{SVM} 
		% \label{Fig6C}
	\end{subfigure}
	\begin{subfigure}{0.32\textwidth}
		\includegraphics[width=0.99\textwidth]{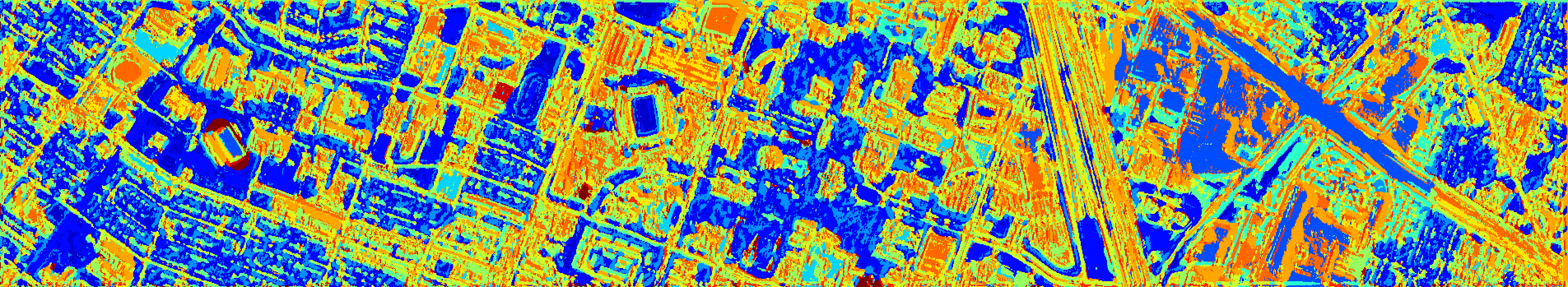}
		\caption{CNN1D} 
		% \label{Fig6D}
	\end{subfigure}
	\begin{subfigure}{0.32\textwidth}
		\includegraphics[width=0.99\textwidth]{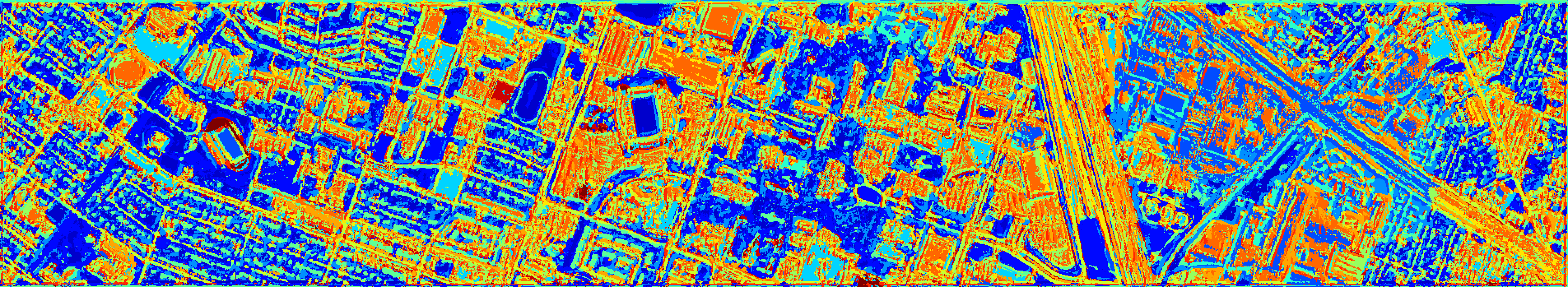}
		\caption{CNN2D}
		% \label{Fig6E}
	\end{subfigure}
    \begin{subfigure}{0.32\textwidth}
		\includegraphics[width=0.99\textwidth]{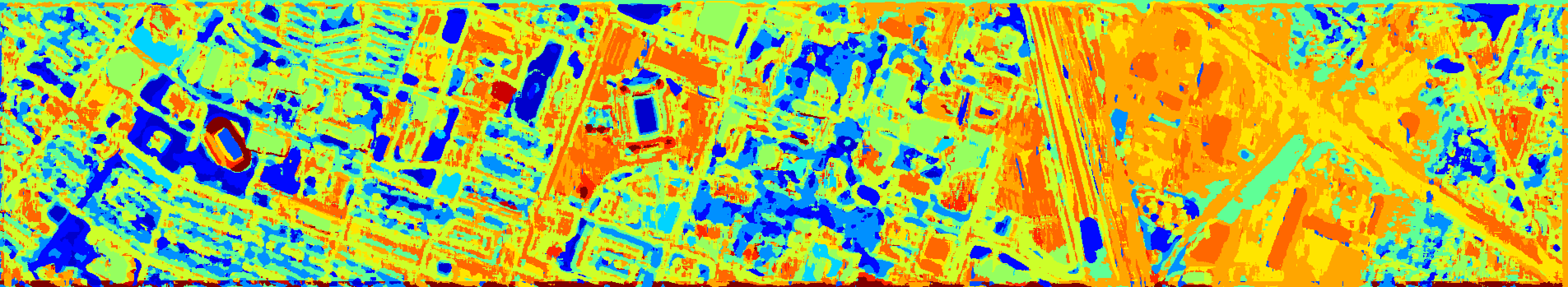}
		\caption{CNN3D}
		% \label{Fig6F}
	\end{subfigure}
	\begin{subfigure}{0.32\textwidth}
		\includegraphics[width=0.99\textwidth]{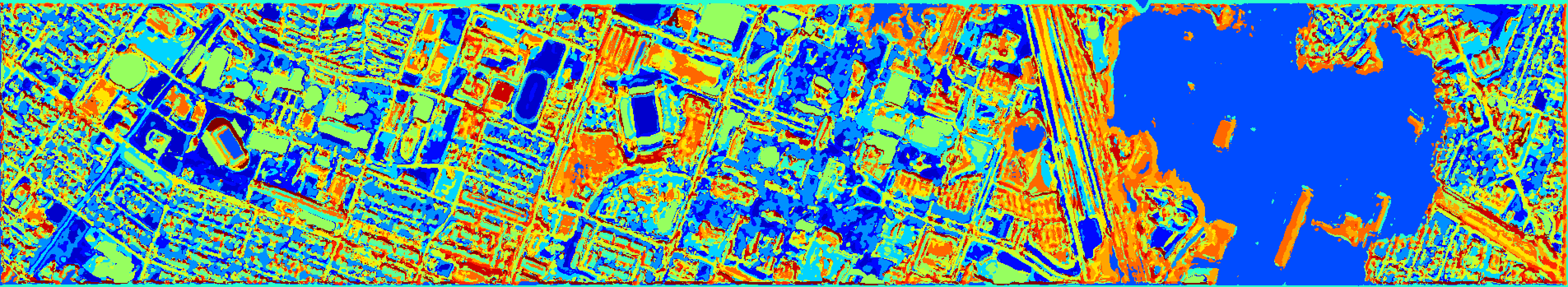}
		\caption{RNN}
		% \label{Fig6G}
	\end{subfigure}
	\begin{subfigure}{0.32\textwidth}
		\includegraphics[width=0.99\textwidth]{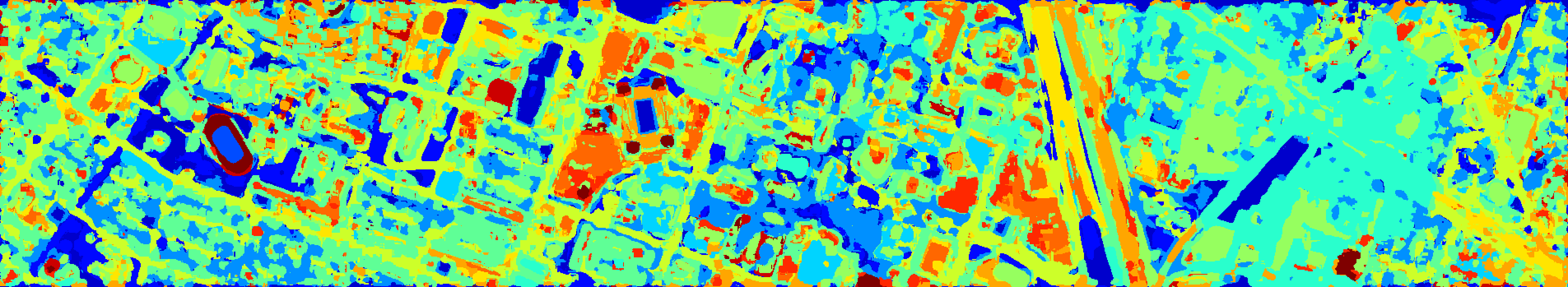}
		\caption{VIT} 
		% \label{Fig6H}
	\end{subfigure}
	\begin{subfigure}{0.32\textwidth}
		\includegraphics[width=0.99\textwidth]{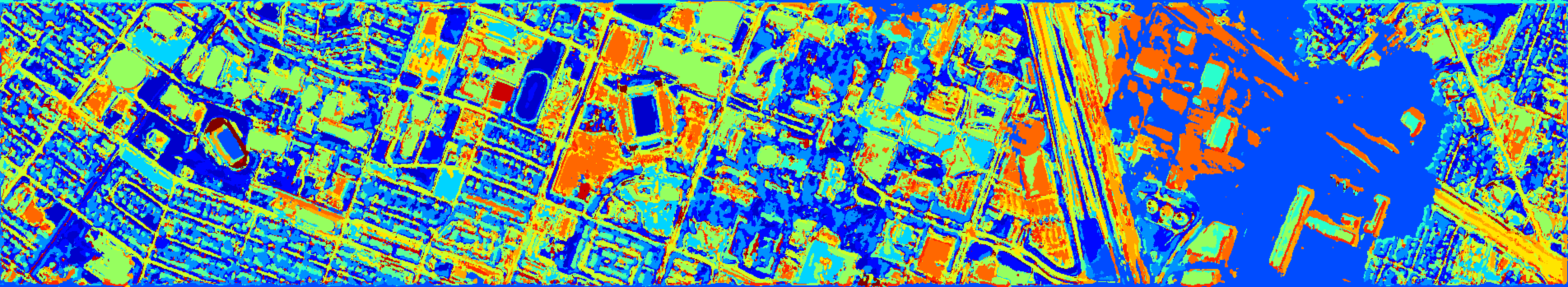}
		\caption{SpectralFormer}
		% \label{Fig6I}
	\end{subfigure}
	\begin{subfigure}{0.32\textwidth}
		\includegraphics[width=0.99\textwidth]{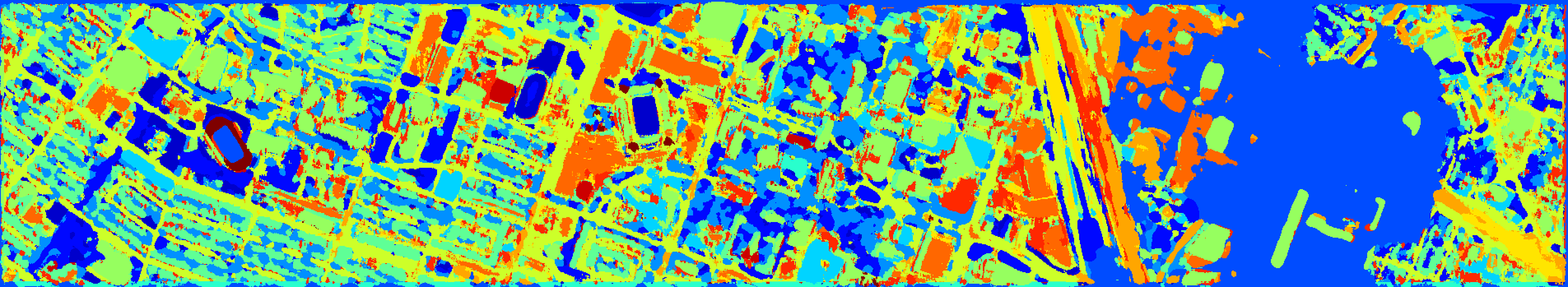}
		\caption{MFT}
		% \label{Fig6J}
	\end{subfigure}
\caption{(a) Ground truth and classification maps obtained by HSI and LiDAR data for the UH dataset by: (b) KNN, (c) RF, (d) CNN1D, (e) CNN2D, (f) CNN3D, (g) RNN, (h) ViT, (i) SpectralFormer, and (j) MFT using disjoint training samples.}
\label{fig:comparativeUH(H+M)}
\end{figure*}

%%%%%%%%%%%%%%%%%%%%%%%%%%%%%%%%%%%%%%%%%%%%%%%%%

\subsection{Performance Evaluation with Disjoint Train/Test Samples}
Table \ref{tab:UH_HS}, \ref{tab:UH_HL} and \ref{tab:UH_HSMS} reports the quantitative OAs, AAs, $\kappa$s, and each class accuracy using the proposed model as well as the other methods for the HSI, the HSI with LiDAR data, and HSI with MS data, respectively, for the disjoint Houston dataset. The best results are shown in bold. The evaluation data indicate that the proposed method outperforms the rest by obtaining the highest OAs, AAs, and $\kappa$s and outperforms others in most class-wise classification accuracies.  

%%%%%%%%%%%%%%%%%%%%%%%%%%%%%%%%%%%%%%%%%%%%%%%%%
\begin{table*}[!t]
\centering
\caption{OA, AA, and Kappa values on the University of Trento dataset (in \%) by considering HS image only.}
\resizebox{\linewidth}{!}{\color{black}
% Please add the following required packages to your document preamble:
% \usepackage{multirow}
\begin{tabular}{c||ccc|cccc||ccc}
\hline 
\multirow{2}{*}{\begin{tabular}[c]{@{}c@{}}\textbf{Class}\\ \textbf{No.}\end{tabular}} 
& \multicolumn{3}{c|}{\textbf{Conventional Classifiers}}            
& \multicolumn{4}{c||}{\textbf{Classical Convolutional Networks}}
& \multicolumn{3}{c}{\textbf{Transformer Networks}}  \\ \cline{2-11}
& \multicolumn{1}{c|}{\textbf{KNN}} & \multicolumn{1}{c|}{\textbf{RF}} & \textbf{SVM} 
& \multicolumn{1}{c|}{\textbf{CNN1D}} 
& \multicolumn{1}{c|}{\textbf{CNN2D}} 
& \multicolumn{1}{l|}{\textbf{CNN3D}} 
& \textbf{RNN} 
& \multicolumn{1}{c|}{\textbf{ViT}} & \multicolumn{1}{c|}{\textbf{SpectralFormer}} 
& \textbf{MFT} \\ \hline \hline
1 &89.07    &96.65 $ \pm $ 00.36    &97.21    &97.29 $ \pm $ 00.20    &96.76 $ \pm $ 00.24    &92.22 $ \pm $ 00.66    &88.06 $ \pm $ 04.36    &89.07 $ \pm $ 00.86    &94.01 $ \pm $ 03.77    &\textbf{98.20 $ \pm $ 00.64}    \\ %\hlaine
2 &88.59    &89.38 $ \pm $ 00.74    &94.17    &86.73 $ \pm $ 01.60    &84.17 $ \pm $ 01.05    &91.42 $ \pm $ 00.69    &79.11 $ \pm $ 06.15    &84.74 $ \pm $ 06.22    &91.26 $ \pm $ 01.23    &\textbf{91.65 $ \pm $ 01.43}    \\
3 &87.17    &73.71 $ \pm $ 01.61    &56.15    &50.62 $ \pm $ 00.13    &50.53 $ \pm $ 02.84    &\textbf{96.88 $ \pm $ 00.91}    &39.22 $ \pm $ 19.36    &92.25 $ \pm $ 01.65    &46.52 $ \pm $ 05.46    &94.92 $ \pm $ 00.95    \\
4 &80.24    &99.92 $ \pm $ 00.02    &84.78    &99.17 $ \pm $ 00.14    &96.12 $ \pm $ 00.24    &99.62 $ \pm $ 00.10    &83.31 $ \pm $ 04.10    &\textbf{99.63 $ \pm $ 00.21}    &85.48 $ \pm $ 11.89    &99.45 $ \pm $ 00.25    \\
5 &95.15    &99.96 $ \pm $ 00.01    &98.13    &99.18 $ \pm $ 00.02    &98.76 $ \pm $ 00.16    &98.99 $ \pm $ 00.12    &97.96 $ \pm $ 00.64    &98.23 $ \pm $ 00.10    &97.60 $ \pm $ 01.45    &\textbf{99.72 $ \pm $ 00.10}    \\
6 &69.59    &66.73 $ \pm $ 01.52    &55.05    &59.62 $ \pm $ 01.47    &66.19 $ \pm $ 00.90    &85.47 $ \pm $ 00.62    &70.87 $ \pm $ 15.09    &84.04 $ \pm $ 06.23    &61.42 $ \pm $ 01.79    &\textbf{91.52 $ \pm $ 01.72}    \\ \hline \hline
OA   &86.42    &94.73 $ \pm $ 00.14    &88.55    &93.02 $ \pm $ 00.06    &92.31 $ \pm $ 00.05    &96.14 $ \pm $ 00.02    &86.83 $ \pm $ 01.91    &94.62 $ \pm $ 00.21    &88.42 $ \pm $ 03.45    &\textbf{97.76 $ \pm $ 00.40}    \\ %\hline
AA  &84.97    &87.73 $ \pm $ 00.43    &80.91    &82.10 $ \pm $ 00.08    &82.09 $ \pm $ 00.46    &94.10 $ \pm $ 00.14    &76.42 $ \pm $ 04.46    &91.33 $ \pm $ 00.22    &79.38 $ \pm $ 00.96    &\textbf{95.91 $ \pm $ 00.41}    \\ %\hline
$\kappa (\times 100)$   &82.21    &92.92 $ \pm $ 00.18    &84.83    &90.65 $ \pm $ 00.09    &89.69 $ \pm $ 00.07    &94.83 $ \pm $ 00.02    &82.49 $ \pm $ 02.53    &92.81 $ \pm $ 00.28    &84.68 $ \pm $ 04.41    &\textbf{97.00 $ \pm $ 00.53}    \\ \hline
\end{tabular}}
\label{tab:UT_HSI}
\end{table*}

%%%%%%%%%%%%%%%%%%%%%%%%%%%%%%%%%%%%%%%%%%%%%%%%%

%%%%%%%%%%%%%%%%%%%%%%%%%%%%%%%%%%%%%%%%%%%%%%%%%
\begin{table*}[!t]
\centering
\caption{OA, AA, and Kappa values on the University of Trento dataset (in \%) by considering HS image and LiDAR data.}
\resizebox{\linewidth}{!}{\color{black}
% Please add the following required packages to your document preamble:
% \usepackage{multirow}
\begin{tabular}{c||ccc|cccc||ccc}
\hline 
\multirow{2}{*}{\begin{tabular}[c]{@{}c@{}}\textbf{Class}\\ \textbf{No.}\end{tabular}} 
& \multicolumn{3}{c|}{\textbf{Conventional Classifiers}}            
& \multicolumn{4}{c||}{\textbf{Classical Convolutional Networks}}
& \multicolumn{3}{c}{\textbf{Transformer Networks}}  \\ \cline{2-11}
& \multicolumn{1}{c|}{\textbf{KNN}} & \multicolumn{1}{c|}{\textbf{RF}} & \textbf{SVM} 
& \multicolumn{1}{c|}{\textbf{CNN1D}} 
& \multicolumn{1}{c|}{\textbf{CNN2D}} 
& \multicolumn{1}{l|}{\textbf{CNN3D}} 
& \textbf{RNN} 
& \multicolumn{1}{c|}{\textbf{ViT}} & \multicolumn{1}{c|}{\textbf{SpectralFormer}} 
& \textbf{MFT} \\ \hline \hline
1 &87.94    &83.73 $ \pm $ 00.06    &97.44    &97.00 $ \pm $ 00.50    &96.98 $ \pm $ 00.21    &92.95 $ \pm $ 00.10    &91.75 $ \pm $ 04.30    &90.87 $ \pm $ 00.77    &96.76 $ \pm $ 01.71    &\textbf{98.23 $ \pm $ 00.38}    \\ %\hline
2 &95.79    &96.30 $ \pm $ 00.06    &98.12    &96.51 $ \pm $ 01.70    &97.56 $ \pm $ 00.14    &98.09 $ \pm $ 00.23    &\textbf{99.47 $ \pm $ 00.37}    &99.32 $ \pm $ 00.77    &97.25 $ \pm $ 00.66    &99.34 $ \pm $ 00.02    \\
3 &81.28    &70.94 $ \pm $ 01.55    &56.15    &42.34 $ \pm $ 06.33    &55.35 $ \pm $ 00.00    &\textbf{93.85 $ \pm $ 01.09}    &79.23 $ \pm $ 16.47    &92.69 $ \pm $ 01.53    &58.47 $ \pm $ 11.54    &89.84 $ \pm $ 09.00    \\
4 &96.25    &99.73 $ \pm $ 00.07    &97.53    &99.77 $ \pm $ 00.05    &99.66 $ \pm $ 00.03    &99.32 $ \pm $ 00.05    &99.58 $ \pm $ 00.42    &\textbf{100.0 $ \pm $ 00.00}    &99.24 $ \pm $ 00.21    &99.82 $ \pm $ 00.26    \\
5 &95.29    &95.35 $ \pm $ 00.25    &98.13    &99.27 $ \pm $ 00.09    &99.56 $ \pm $ 00.07    &98.74 $ \pm $ 00.04    &98.39 $ \pm $ 00.65    &97.77 $ \pm $ 00.86    &93.52 $ \pm $ 01.75    &\textbf{99.93 $ \pm $ 00.05}    \\
6 &83.85    &72.63 $ \pm $ 00.90    &78.96    &76.91 $ \pm $ 03.62    &76.91 $ \pm $ 00.15    &88.15 $ \pm $ 00.20    &85.86 $ \pm $ 02.89    &86.72 $ \pm $ 02.02    &73.39 $ \pm $ 06.78    &\textbf{88.72 $ \pm $ 00.94}    \\ \hline \hline
OA   &93.29    &92.57 $ \pm $ 00.07    &95.33    &95.81 $ \pm $ 00.13    &96.14 $ \pm $ 00.03    &96.93 $ \pm $ 00.03    &96.43 $ \pm $ 00.79    &96.47 $ \pm $ 00.49    &93.51 $ \pm $ 01.27    &\textbf{98.32 $ \pm $ 00.25}    \\ %\hline
AA  &90.07    &86.45 $ \pm $ 00.32    &87.72    &85.30 $ \pm $ 00.72    &87.67 $ \pm $ 00.04    &95.18 $ \pm $ 00.18    &92.38 $ \pm $ 03.50    &94.56 $ \pm $ 00.57    &86.44 $ \pm $ 02.96    &\textbf{95.98 $ \pm $ 01.64}    \\ %\hline
$\kappa (\times 100)$   &91.11    &90.11 $ \pm $ 00.09    &93.76    &94.39 $ \pm $ 00.17    &94.83 $ \pm $ 00.04    &95.89 $ \pm $ 00.04    &95.21 $ \pm $ 01.06    &95.28 $ \pm $ 00.65    &91.36 $ \pm $ 01.67    &\textbf{97.75 $ \pm $ 00.33}    \\ \hline
\end{tabular}}
\label{tab:UT_HL}
\end{table*}

%%%%%%%%%%%%%%%%%%%%%%%%%%%%%%%%%%%%%%%%%%%%%%%%%

Overall, the transformer networks outperform the conventional networks and classifiers. The proposed model dominates the class-wise accuracies and surpasses the other methods in terms of OA, AA, and Kappa in all three cases for the Houston dataset, i.e., HSI, HSI with LiDAR, and HSI with MS image. Combining HSI data with LiDAR data, even using the traditional fusion method, leads to performance gains in almost all models. Though the proposed model stands out on top in the case of Houston using HSI with LiDAR data, the increase in performance ($1.35\%$ OA, $1.46\%$ AA, and $1.47\%$ $\kappa$) is not significantly better than that achieved by the other models. The new transformer based fusion technique used in the proposed model provides the best result for Houston using HSI data with MS image, where almost all the other models that use the traditional fusion method have lower OAs, AAs, and $\kappa$s when compared with those in the case Houston using HSI data only, but the gains obtained by the proposed model remain consistent.

It can be seen from Table \ref{tab:UH_HS} that, among the conventional classifiers, RF performs the best, having mean OA, AA and $\kappa$ $74.87\%$, $77.94\%$ and $72.93\%$ with standard deviations of $0.09\%$, $0.14\%$ and $0.09\%$, respectively. RF can also beat all the classical networks shown in the table but is not as good when compared with transformer networks. The proposed transformer network provides the best performance among the three transformer-based networks giving mean OA, AA, and $\kappa$ $88.45\%$, $90.05\%$ and $87.46\%$ with standard deviations of $0.28\%$, $0.14\%$ and $0.09\%$ respectively, followed by ViT and SpectralFormer. The standard deviations of the proposed model are also lower than those of the other transformer networks, since our model can represent long-range dependencies between the feature maps better than ViT and SpectralFormer. Conventional classifiers, which rely completely on the spectral representation of HSI data, cannot capture the spatial information and hence perform worse than transformer networks.
%%%%%%%%%%%%%%%%%%%%%%%%%%%%%%%%%%%%%%%%%%%%%%%%%
\begin{figure*}[!t]
\centering
	\begin{subfigure}{0.32\textwidth}
		\includegraphics[width=0.99\textwidth]{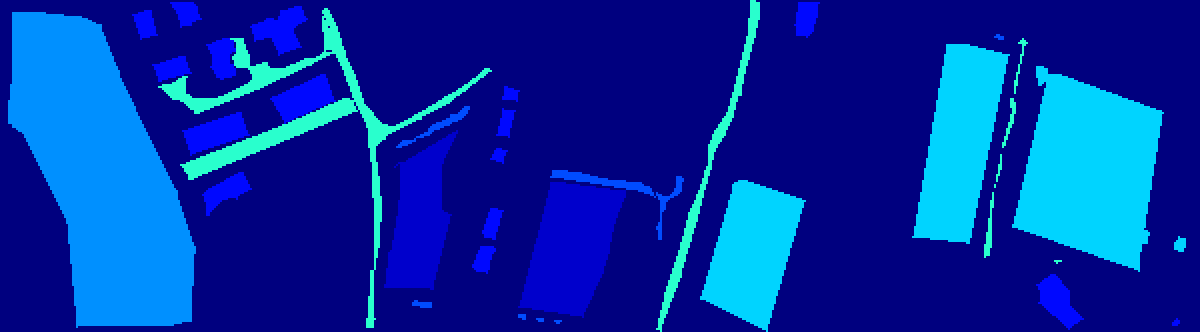}
		\caption{GT}
		% \label{Fig6A}
	\end{subfigure}
	\begin{subfigure}{0.32\textwidth}
		\includegraphics[width=0.99\textwidth]{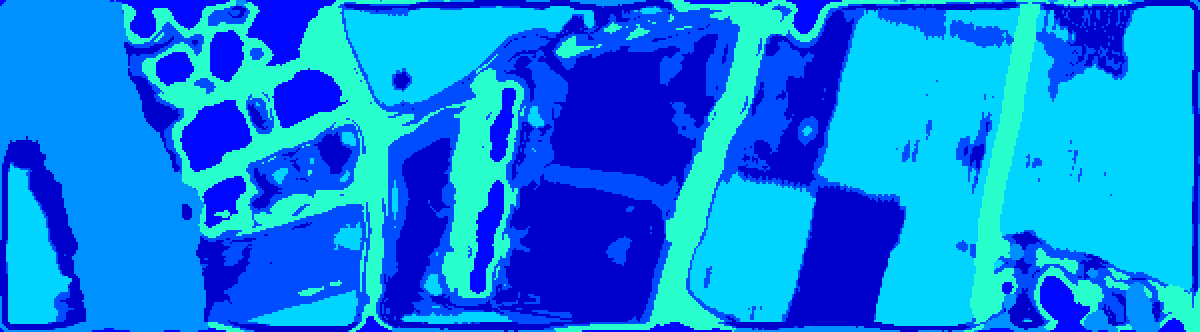}
		\caption{KNN}
		% \label{Fig6A}
	\end{subfigure}
	\begin{subfigure}{0.32\textwidth}
		\includegraphics[width=0.99\textwidth]{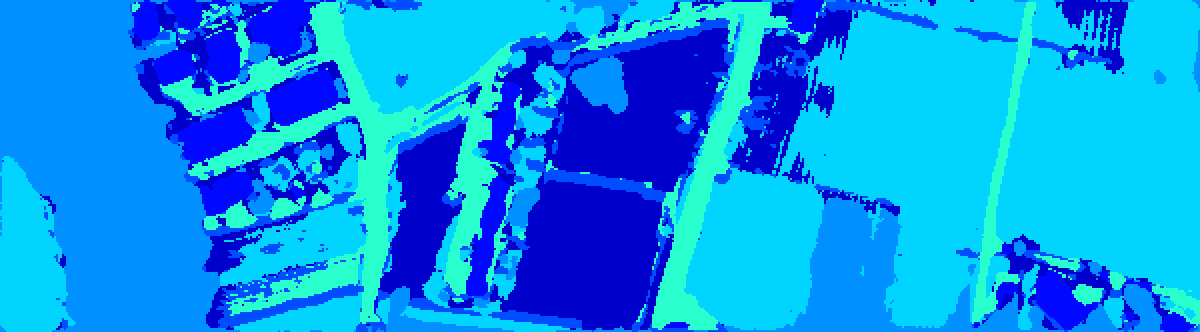}
		\caption{RF}
		% \label{Fig6B}
	\end{subfigure}
	\begin{subfigure}{0.32\textwidth}
		\includegraphics[width=0.99\textwidth]{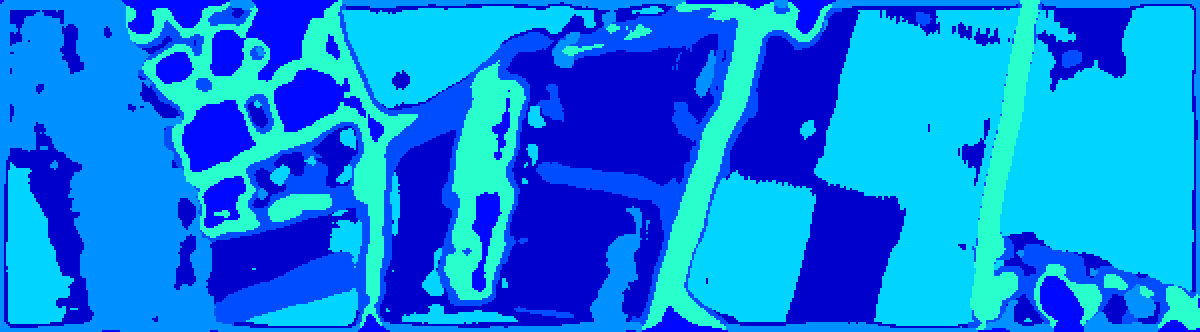}
		\caption{SVM} 
		% \label{Fig6C}
	\end{subfigure}
	\begin{subfigure}{0.32\textwidth}
		\includegraphics[width=0.99\textwidth]{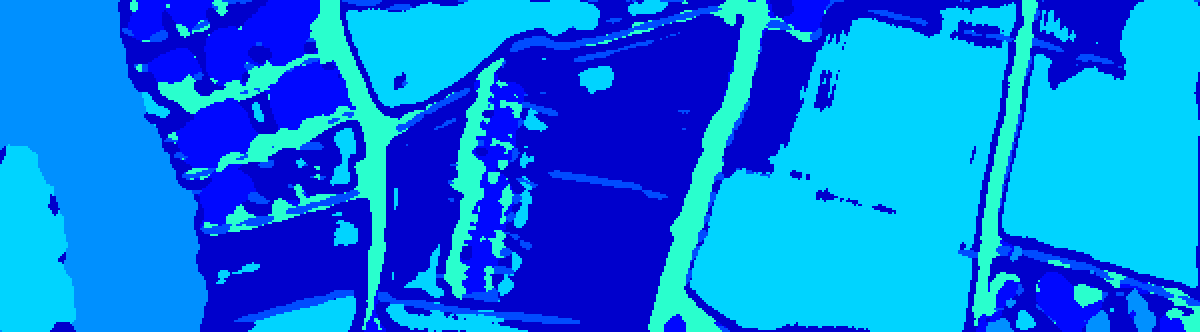}
		\caption{CNN1D} 
		% \label{Fig6D}
	\end{subfigure}
	\begin{subfigure}{0.32\textwidth}
		\includegraphics[width=0.99\textwidth]{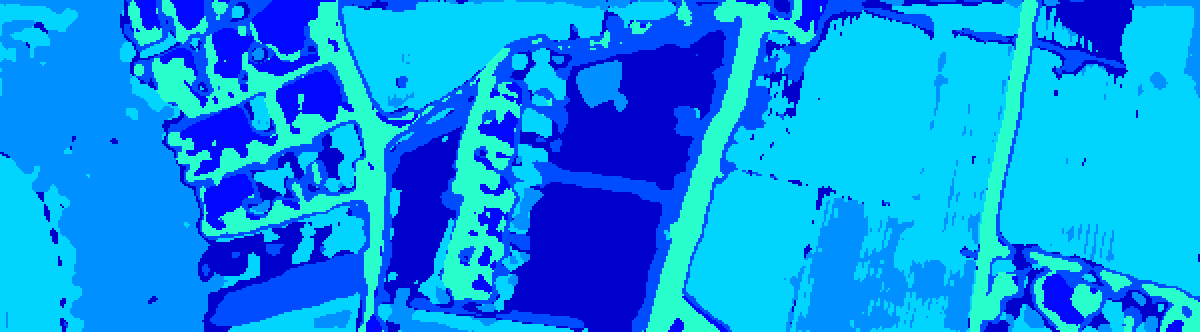}
		\caption{CNN2D}
		% \label{Fig6E}
	\end{subfigure}
    \begin{subfigure}{0.32\textwidth}
		\includegraphics[width=0.99\textwidth]{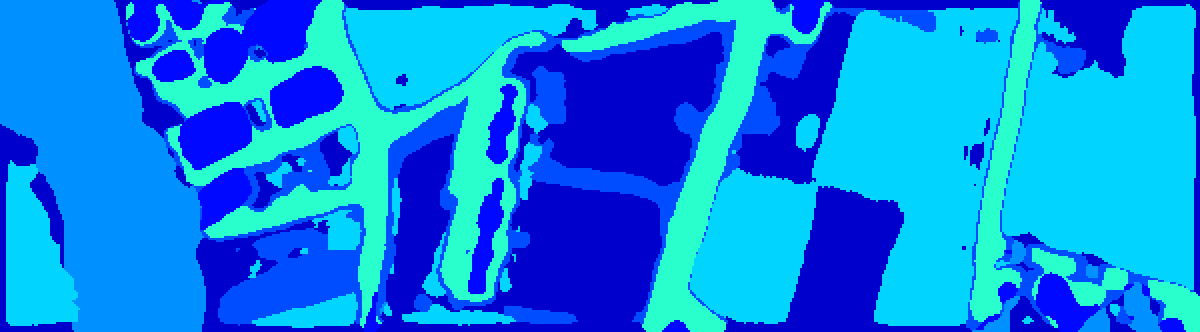}
		\caption{CNN3D}
		% \label{Fig6F}
	\end{subfigure}
	\begin{subfigure}{0.32\textwidth}
		\includegraphics[width=0.99\textwidth]{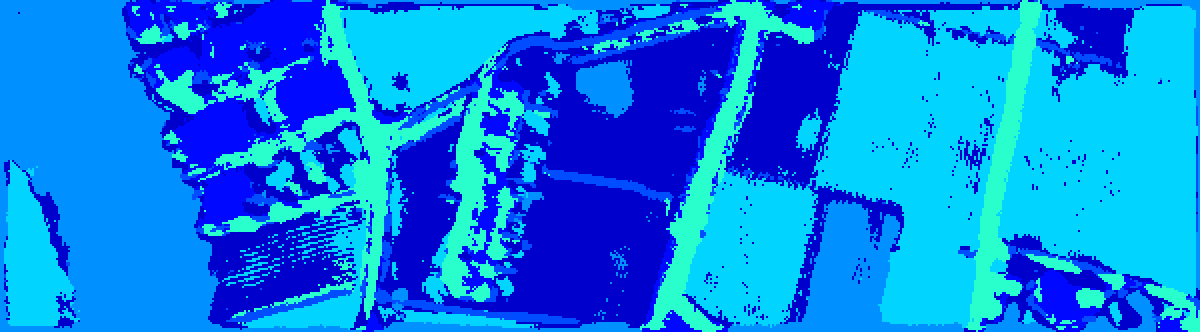}
		\caption{RNN}
		% \label{Fig6G}
	\end{subfigure}
	\begin{subfigure}{0.32\textwidth}
		\includegraphics[width=0.99\textwidth]{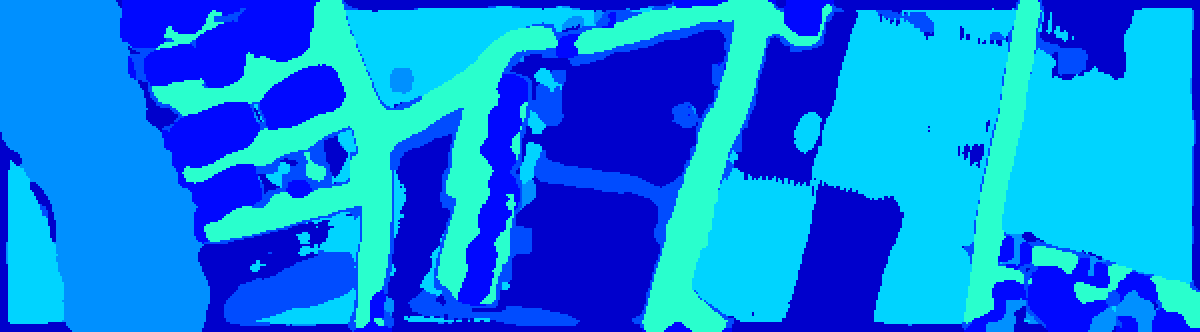}
		\caption{ViT} 
		% \label{Fig6H}
	\end{subfigure}
	\begin{subfigure}{0.32\textwidth}
		\includegraphics[width=0.99\textwidth]{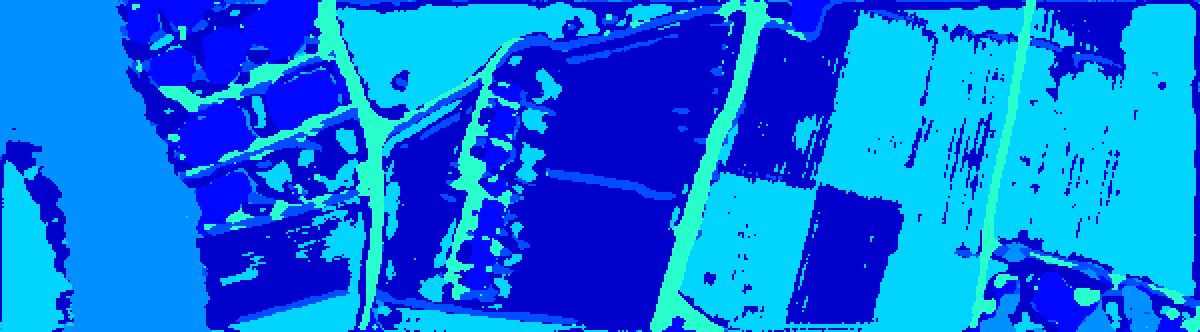}
		\caption{SpectralFormer}
		% \label{Fig6I}
	\end{subfigure}
	\begin{subfigure}{0.32\textwidth}
		\includegraphics[width=0.99\textwidth]{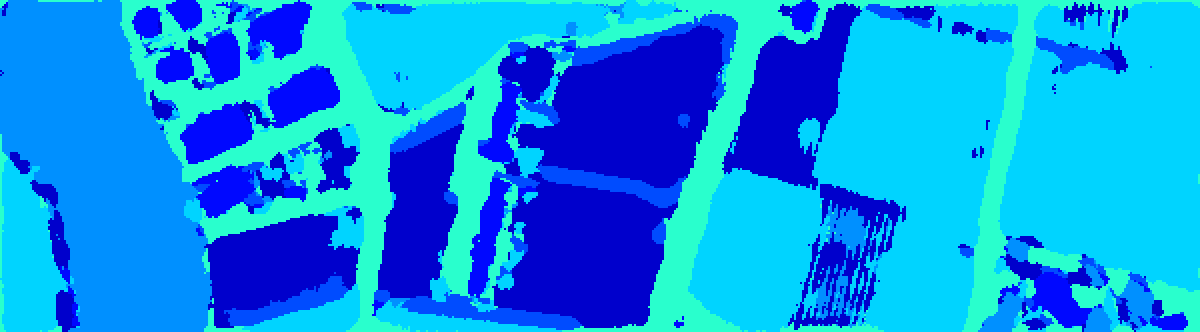}
		\caption{MFT}
		% \label{Fig6J}
	\end{subfigure}
\caption{(a) Ground truth and classification maps obtained by HSI and LiDAR data for the Trento data set by: (b) KNN, (c) RF, (d) CNN1D, (e) CNN2D, (f) CNN3D, (g) RNN, (h) ViT, (i) SpectralFormer, and (j) MFT using disjoint training samples.}
\label{fig:comparativeUT}
\end{figure*}

%%%%%%%%%%%%%%%%%%%%%%%%%%%%%%%%%%%%%%%%%%%%%%%%%
Table \ref{tab:UH_HL} indicates a substantial rise in the accuracies of all the given classical networks, where the accuracies of the best-performing classical networks are close to those of the best-performing conventional classifiers. Table \ref{tab:UH_HS} confirms that RF outperforms the conventional classifiers as well as the classical networks due to its better discriminative feature learning, but still cannot compete with the transformer networks. The proposed model again beats the rest by quite a margin due to its superior learning of spectral and spatial information, with mean OA, AA, and $\kappa$ $89.80\%$, $91.51\%$ and $88.93\%$ with minimal standard deviations of $0.53\%$, $0.40\%$ and $0.59\%$ respectively.
%%%%%%%%%%%%%%%%%%%%%%%%%%%%%%%%%%%%%%%%%%%%%%%%%
\begin{table*}[!t]
\centering
\caption{OA, AA, and Kappa values on the MUUFL dataset (in \%) by considering HS image data only.}
\resizebox{\linewidth}{!}{\color{black}
% Please add the following required packages to your document preamble:
% \usepackage{multirow}
\begin{tabular}{c||ccc|cccc||ccc}
\hline 
\multirow{2}{*}{\begin{tabular}[c]{@{}c@{}}\textbf{Class}\\ \textbf{No.}\end{tabular}} 
& \multicolumn{3}{c|}{\textbf{Conventional Classifiers}}            
& \multicolumn{4}{c||}{\textbf{Classical Convolutional Networks}}
& \multicolumn{3}{c}{\textbf{Transformer Networks}}  \\ \cline{2-11}
& \multicolumn{1}{c|}{\textbf{KNN}} & \multicolumn{1}{c|}{\textbf{RF}} & \textbf{SVM} 
& \multicolumn{1}{c|}{\textbf{CNN1D}} 
& \multicolumn{1}{c|}{\textbf{CNN2D}} 
& \multicolumn{1}{l|}{\textbf{CNN3D}} 
& \textbf{RNN} 
& \multicolumn{1}{c|}{\textbf{ViT}} & \multicolumn{1}{c|}{\textbf{SpectralFormer}} 
& \textbf{MFT} \\ \hline \hline
1 &92.74    &97.97 $ \pm $ 00.07    &96.64    &95.21 $ \pm $ 00.05    &95.94 $ \pm $ 00.20    &94.72 $ \pm $ 00.65    &96.06 $ \pm $ 00.19    &97.40 $ \pm $ 00.25    &94.63 $ \pm $ 00.89    &\textbf{97.61 $ \pm $ 00.15}    \\ %\hline
2 &47.46    &77.86 $ \pm $ 00.19    &59.37    &69.31 $ \pm $ 01.06    &70.87 $ \pm $ 01.50    &62.98 $ \pm $ 03.76    &80.34 $ \pm $ 02.85    &77.34 $ \pm $ 01.62    &77.12 $ \pm $ 03.77    &\textbf{92.51 $ \pm $ 00.87}    \\
3 &69.10    &84.33 $ \pm $ 00.23    &81.49    &74.31 $ \pm $ 00.60    &78.65 $ \pm $ 00.78    &71.15 $ \pm $ 03.46    &80.65 $ \pm $ 00.87    &86.10 $ \pm $ 01.49    &74.26 $ \pm $ 02.81    &\textbf{92.12 $ \pm $ 01.66}    \\
4 &53.43    &86.09 $ \pm $ 00.37    &74.35    &78.71 $ \pm $ 01.27    &82.96 $ \pm $ 01.17    &64.86 $ \pm $ 02.15    &86.09 $ \pm $ 01.95    &\textbf{93.35 $ \pm $ 00.84}    &84.59 $ \pm $ 01.18    &92.83 $ \pm $ 01.30    \\
5 &83.68    &92.10 $ \pm $ 00.08    &83.80    &75.84 $ \pm $ 01.93    &78.76 $ \pm $ 01.09    &77.95 $ \pm $ 02.15    &90.15 $ \pm $ 00.39    &\textbf{94.82 $ \pm $ 00.77}    &89.38 $ \pm $ 02.06    &94.31 $ \pm $ 01.26    \\
6 &01.13    &68.77 $ \pm $ 00.65    &15.35    &46.35 $ \pm $ 00.74    &48.53 $ \pm $ 02.60    &75.24 $ \pm $ 90.92    &53.42 $ \pm $ 04.33    &82.69 $ \pm $ 01.31    &60.42 $ \pm $ 09.08    &\textbf{88.56 $ \pm $ 02.86}    \\
7 &43.33    &80.47 $ \pm $ 00.42    &77.09    &78.45 $ \pm $ 00.58    &78.96 $ \pm $ 00.30    &48.44 $ \pm $ 01.87    &80.39 $ \pm $ 02.75    &87.44 $ \pm $ 00.97    &83.77 $ \pm $ 03.58    &\textbf{92.68 $ \pm $ 01.42}    \\
8 &73.48    &91.57 $ \pm $ 00.17    &87.11    &68.28 $ \pm $ 02.36    &68.86 $ \pm $ 00.37    &69.48 $ \pm $ 01.91    &88.67 $ \pm $ 02.04    &\textbf{97.41 $ \pm $ 00.75}    &88.91 $ \pm $ 02.78    &97.08 $ \pm $ 00.87    \\
9 &04.94    &45.87 $ \pm $ 00.31    &21.50    &39.97 $ \pm $ 00.27    &45.64 $ \pm $ 00.48    &04.79 $ \pm $ 03.39    &58.94 $ \pm $ 00.73    &58.92 $ \pm $ 07.76    &55.02 $ \pm $ 01.43    &\textbf{59.80 $ \pm $ 01.56}    \\
10 &00.00    &04.98 $ \pm $ 03.05    &00.00    &08.81 $ \pm $ 00.72    &13.03 $ \pm $ 00.27    &00.00 $ \pm $ 00.00    &22.80 $ \pm $ 04.51    &\textbf{32.76 $ \pm $ 07.01}    &17.62 $ \pm $ 02.12    &12.45 $ \pm $ 04.79    \\
11 &58.59    &48.31 $ \pm $ 02.12    &61.33    &23.83 $ \pm $ 03.24    &24.74 $ \pm $ 00.97    &61.20 $ \pm $ 02.17    &\textbf{80.99 $ \pm $ 04.01}    &66.67 $ \pm $ 04.94    &45.44 $ \pm $ 07.29    &71.09 $ \pm $ 01.66    \\ \hline \hline
OA   &75.80    &89.85 $ \pm $ 00.03    &84.30    &81.17 $ \pm $ 00.09    &82.95 $ \pm $ 00.12    &77.59 $ \pm $ 00.05    &88.60 $ \pm $ 00.69    &91.99 $ \pm $ 00.35    &86.68 $ \pm $ 00.93    &\textbf{94.18 $ \pm $ 00.09}    \\ %\hline
AA  &47.99    &70.76 $ \pm $ 00.29    &59.82    &59.92 $ \pm $ 00.52    &62.45 $ \pm $ 00.22    &50.58 $ \pm $ 00.34    &74.41 $ \pm $ 01.40    &79.54 $ \pm $ 01.93    &70.11 $ \pm $ 01.52    &\textbf{81.00 $ \pm $ 00.55}    \\ %\hline
$\kappa (\times 100)$   &67.34    &86.44 $ \pm $ 00.04    &78.89    &74.95 $ \pm $ 00.12    &77.30 $ \pm $ 00.16    &69.81 $ \pm $ 00.10    &84.90 $ \pm $ 00.92    &89.37 $ \pm $ 00.46    &82.43 $ \pm $ 01.19    &\textbf{92.30 $ \pm $ 00.12}    \\ \hline
\end{tabular}}
\label{tab:MUUFL_HSI}
\end{table*}

%%%%%%%%%%%%%%%%%%%%%%%%%%%%%%%%%%%%%%%%%%%%%%%%%
%%%%%%%%%%%%%%%%%%%%%%%%%%%%%%%%%%%%%%%%%%%%%%%%%
\begin{table*}[!t]
\centering
\caption{OA, AA and Kappa values on the MUUFL dataset (in \%) by considering HS image and LiDAR data.}
\resizebox{\linewidth}{!}{\color{black}
% Please add the following required packages to your document preamble:
% \usepackage{multirow}
\begin{tabular}{c||ccc|cccc||ccc}
\hline 
\multirow{2}{*}{\begin{tabular}[c]{@{}c@{}}\textbf{Class}\\ \textbf{No.}\end{tabular}} 
& \multicolumn{3}{c|}{\textbf{Conventional Classifiers}}            
& \multicolumn{4}{c||}{\textbf{Classical Convolutional Networks}}
& \multicolumn{3}{c}{\textbf{Transformer Networks}}  \\ \cline{2-11}
& \multicolumn{1}{c|}{\textbf{KNN}} & \multicolumn{1}{c|}{\textbf{RF}} & \textbf{SVM} 
& \multicolumn{1}{c|}{\textbf{CNN1D}} 
& \multicolumn{1}{c|}{\textbf{CNN2D}} 
& \multicolumn{1}{l|}{\textbf{CNN3D}} 
& \textbf{RNN} 
& \multicolumn{1}{c|}{\textbf{ViT}} & \multicolumn{1}{c|}{\textbf{SpectralFormer}} 
& \textbf{MFT} \\ \hline \hline
1 &92.12    &95.42 $ \pm $ 00.09    &96.63    &95.05 $ \pm $ 00.22    &95.79 $ \pm $ 00.11    &95.10 $ \pm $ 00.19    &95.84 $ \pm $ 00.14    &97.85 $ \pm $ 00.29    &97.30 $ \pm $ 00.83    &\textbf{97.90 $ \pm $ 00.39}    \\ %\hline
2 &51.85    &74.03 $ \pm $ 00.11    &59.25    &70.35 $ \pm $ 03.30    &72.76 $ \pm $ 00.58    &63.72 $ \pm $ 03.18    &81.93 $ \pm $ 02.11    &76.06 $ \pm $ 02.40    &69.35 $ \pm $ 05.16    &\textbf{92.11 $ \pm $ 01.58}    \\
3 &69.35    &75.81 $ \pm $ 00.38    &81.46    &75.80 $ \pm $ 02.09    &78.92 $ \pm $ 00.52    &69.94 $ \pm $ 03.42    &80.47 $ \pm $ 02.13    &87.58 $ \pm $ 03.46    &78.48 $ \pm $ 03.41    &\textbf{91.80 $ \pm $ 00.82}    \\
4 &57.00    &68.59 $ \pm $ 00.77    &73.54    &78.60 $ \pm $ 02.77    &83.59 $ \pm $ 00.99    &63.90 $ \pm $ 01.84    &87.01 $ \pm $ 01.46    &\textbf{92.05 $ \pm $ 02.31}    &82.63 $ \pm $ 03.68    &91.59 $ \pm $ 02.25    \\
5 &83.87    &88.17 $ \pm $ 00.18    &83.79    &78.31 $ \pm $ 01.48    &78.29 $ \pm $ 01.12    &79.48 $ \pm $ 01.43    &90.65 $ \pm $ 00.65    &94.73 $ \pm $ 00.60    &87.91 $ \pm $ 02.97    &\textbf{95.60 $ \pm $ 01.21}    \\
6 &19.19    &77.28 $ \pm $ 00.93    &15.35    &46.35 $ \pm $ 02.49    &50.34 $ \pm $ 02.13    &02.86 $ \pm $ 03.00    &54.25 $ \pm $ 03.14    &82.02 $ \pm $ 01.13    &58.77 $ \pm $ 02.76    &\textbf{88.19 $ \pm $ 03.49}    \\
7 &44.60    &64.83 $ \pm $ 00.97    &77.04    &78.31 $ \pm $ 00.20    &79.70 $ \pm $ 00.26    &47.96 $ \pm $ 01.00    &81.24 $ \pm $ 01.32    &87.11 $ \pm $ 01.54    &85.87 $ \pm $ 00.62    &\textbf{90.27 $ \pm $ 02.13}    \\
8 &76.97    &93.29 $ \pm $ 00.27    &86.94    &66.72 $ \pm $ 01.17    &71.95 $ \pm $ 01.10    &70.47 $ \pm $ 01.25    &88.39 $ \pm $ 01.50    &\textbf{97.60 $ \pm $ 00.16}    &95.60 $ \pm $ 01.26    &97.26 $ \pm $ 00.53    \\
9 &09.95    &19.15 $ \pm $ 01.37    &21.28    &40.15 $ \pm $ 02.96    &43.92 $ \pm $ 01.24    &06.28 $ \pm $ 04.91    &60.54 $ \pm $ 04.40    &57.83 $ \pm $ 04.45    &53.52 $ \pm $ 04.32    &\textbf{61.35 $ \pm $ 03.80}    \\
10 &00.00    &04.41 $ \pm $ 00.72    &00.00    &09.20 $ \pm $ 01.24    &12.45 $ \pm $ 00.27    &00.00 $ \pm $ 00.00    &26.44 $ \pm $ 02.82    &\textbf{31.99 $ \pm $ 08.86}    &08.43 $ \pm $ 02.22    &17.43 $ \pm $ 04.63    \\
11 &64.45    &71.88 $ \pm $ 00.84    &62.89    &25.65 $ \pm $ 02.89    &26.82 $ \pm $ 02.60    &66.93 $ \pm $ 01.76    &\textbf{87.50 $ \pm $ 02.92}    &58.72 $ \pm $ 03.85    &35.29 $ \pm $ 06.00    &72.79 $ \pm $ 09.25    \\ \hline \hline
OA   &76.83    &85.32 $ \pm $ 00.09    &84.24    &81.50 $ \pm $ 00.03    &83.40 $ \pm $ 00.04    &77.99 $ \pm $ 00.06    &88.79 $ \pm $ 00.45    &92.15 $ \pm $ 00.19    &88.25 $ \pm $ 00.56    &\textbf{94.34 $ \pm $ 00.07}    \\ %\hline
AA  &51.76    &66.62 $ \pm $ 00.16    &59.83    &60.41 $ \pm $ 00.48    &63.14 $ \pm $ 00.21    &51.51 $ \pm $ 00.40    &75.84 $ \pm $ 00.62    &78.50 $ \pm $ 01.28    &68.47 $ \pm $ 01.44    &\textbf{81.48 $ \pm $ 00.70}    \\ %\hline
$\kappa (\times 100)$   &68.92    &80.39 $ \pm $ 00.12    &78.80    &75.43 $ \pm $ 00.07    &77.94 $ \pm $ 00.06    &70.31 $ \pm $ 00.03    &85.18 $ \pm $ 00.60    &89.56 $ \pm $ 00.27    &84.40 $ \pm $ 00.77    &\textbf{92.51 $ \pm $ 00.10}    \\ \hline
\end{tabular}}
\label{tab:MUUFL_HL}
\end{table*}

%%%%%%%%%%%%%%%%%%%%%%%%%%%%%%%%%%%%%%%%%%%%%%%%%

Table \ref{tab:UH_HSMS} reveals an opposite trend than Table \ref{tab:UH_HL} in all but two networks. The networks other than CNN3D and the proposed network perform worse with the fusion of both HSI and MS image using the traditional fusion method when compared to that of HSI only (Table \ref{tab:UH_HS}). Unsurprisingly, the novel transformer-based fusion technique allows the proposed model to increase its performance resulting in mean OA, AA, and $\kappa$ of $89.15\%$, $90.56\%$, and $88.22\%$ with acceptable standard deviations of $0.96\%$, $0.93\%$ and $1.04\%$ respectively.

%%%%%%%%%%%%%%%%%%%%%%%%%%%%%%%%%%%%%%%%%%%%%%%%%

To demonstrate the generalization ability of the proposed model for remote sensing image classification methods, we have considered Trento and MUUFL datasets (HSI data only and HSI with LiDAR data) with disjoint training and test samples. Fig.~\ref{fig:Trento} shows the pseudo color map, corresponding LiDAR data, and the disjoint train and test ground truth for the Trento dataset. Fig.~\ref{fig:Trento} also provides $6$ different land-cover classes and the number of samples belonging to the train and test set for each class in a table. Fig.~\ref{fig:MUUFL} shows the pseudo color map corresponding to the LiDAR image, and the disjoint train and test ground truth for the MUUFL dataset. It also gives the $11$ different land-cover classes and the number of training and testing samples per class.

Tables \ref{tab:UT_HSI} and \ref{tab:MUUFL_HSI} show the experimental results for Trento and MUUFL datasets using HS images only. Among conventional classifiers and classical networks, CNN3D performs the best for Trento. In the case of MUUFL, both RF and RNN perform similarly and better than the other conventional classifiers and classical networks. In the case of transformer-based models, the proposed model takes the lead with OA =$97.76 \pm 00.40\%$, AA =$95.91 \pm 00.41\%$ and $\kappa$ =$97.00 \pm 00.53\%$ for Trento and OA =$94.18 \pm 00.09\%$, AA =$81.00 \pm 00.55\%$ and $\kappa$ =$92.30 \pm 00.12\%$ for MUUFL.

%%%%%%%%%%%%%%%%%%%%%%%%%%%%%%%%%%%%%%%%%%%%%%%%%
\begin{figure*}[!t]
\centering
	\begin{subfigure}{0.15\textwidth}
		\includegraphics[width=0.99\textwidth]{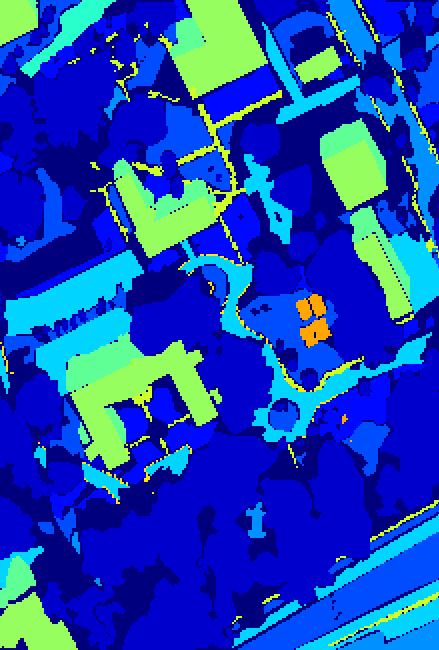}
		\caption{GT}
		% \label{Fig6A}
	\end{subfigure}
	\begin{subfigure}{0.15\textwidth}
		\includegraphics[width=0.99\textwidth]{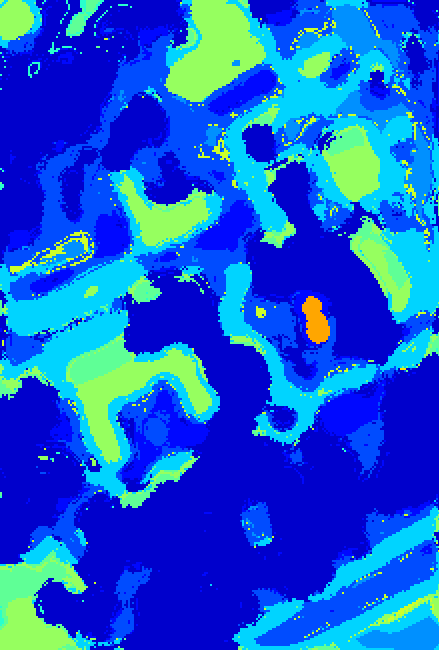}
		\caption{KNN}
		% \label{Fig6A}
	\end{subfigure}
	\begin{subfigure}{0.15\textwidth}
		\includegraphics[width=0.99\textwidth]{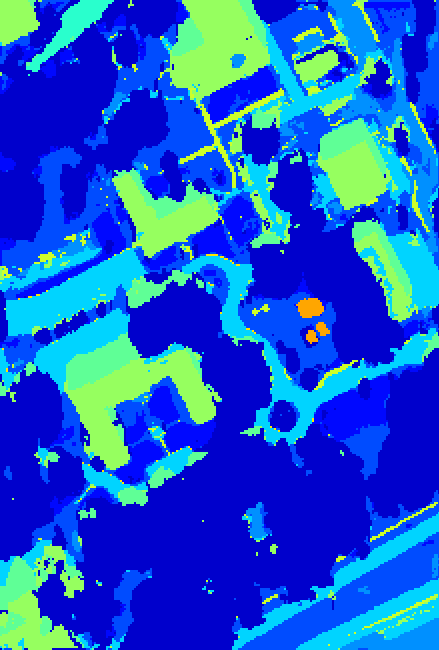}
		\caption{RF}
		% \label{Fig6B}
	\end{subfigure}
	\begin{subfigure}{0.15\textwidth}
		\includegraphics[width=0.99\textwidth]{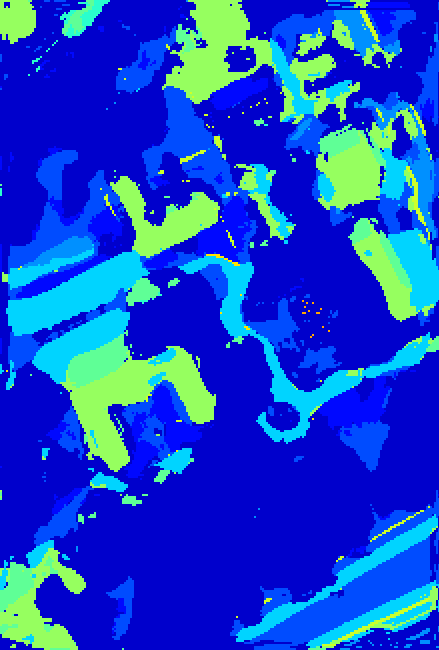}
		\caption{SVM} 
		% \label{Fig6C}
	\end{subfigure}
	\begin{subfigure}{0.15\textwidth}
		\includegraphics[width=0.99\textwidth]{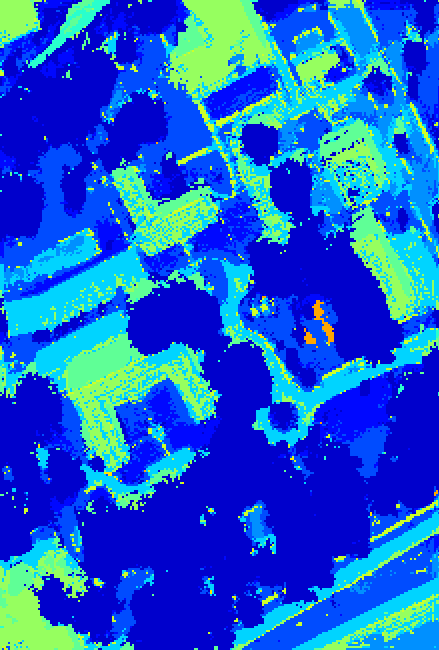}
		\caption{CNN1D} 
		% \label{Fig6D}
	\end{subfigure}
	\begin{subfigure}{0.15\textwidth}
		\includegraphics[width=0.99\textwidth]{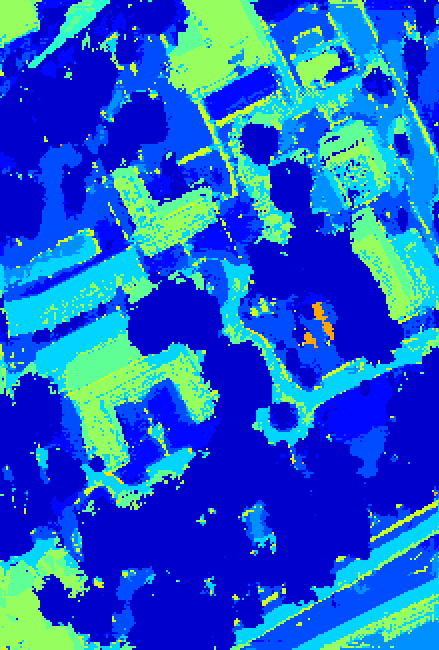}
		\caption{CNN2D}
		% \label{Fig6E}
	\end{subfigure}
    \begin{subfigure}{0.15\textwidth}
		\includegraphics[width=0.99\textwidth]{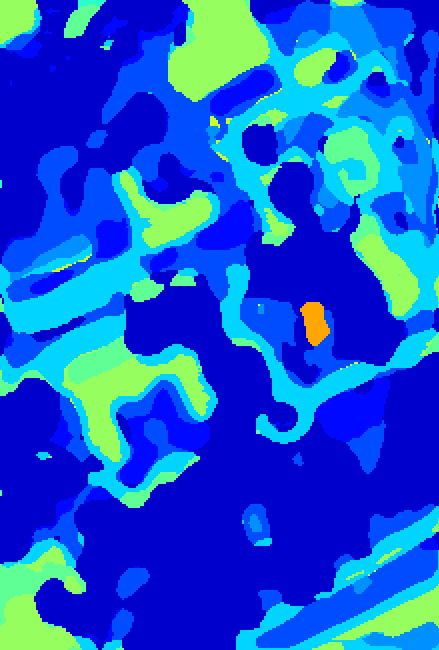}
		\caption{CNN3D}
		% \label{Fig6F}
	\end{subfigure}
	\begin{subfigure}{0.15\textwidth}
		\includegraphics[width=0.99\textwidth]{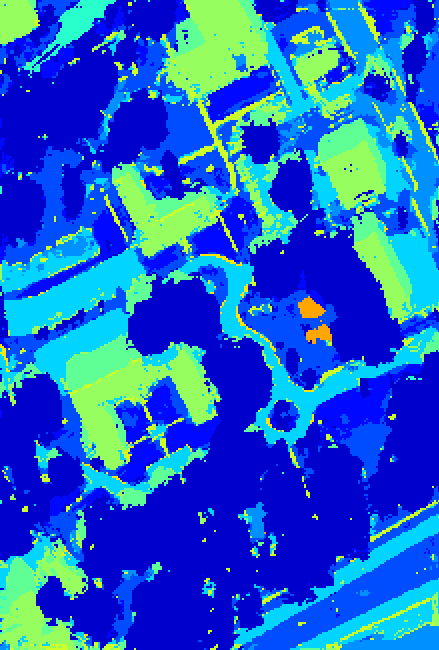}
		\caption{RNN}
		% \label{Fig6G}
	\end{subfigure}
	\begin{subfigure}{0.15\textwidth}
		\includegraphics[width=0.99\textwidth]{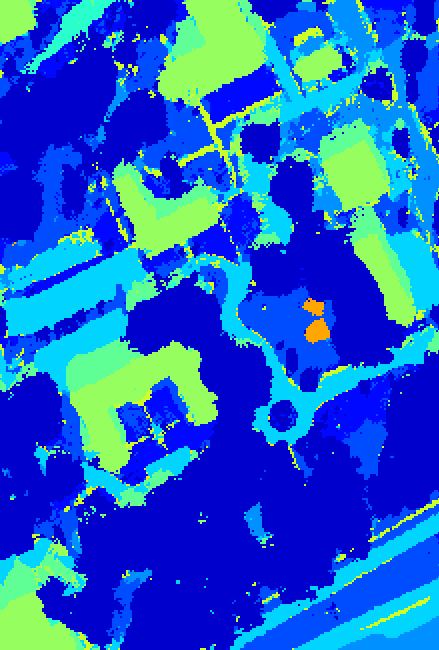}
		\caption{VIT} 
		% \label{Fig6H}
	\end{subfigure}
	\begin{subfigure}{0.15\textwidth}
		\includegraphics[width=0.99\textwidth]{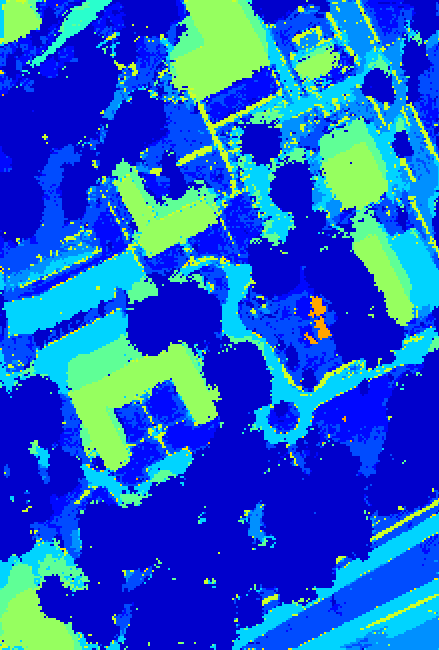}
		\caption{SpectralFormer}
		% \label{Fig6I}
	\end{subfigure}
	\begin{subfigure}{0.15\textwidth}
		\includegraphics[width=0.99\textwidth]{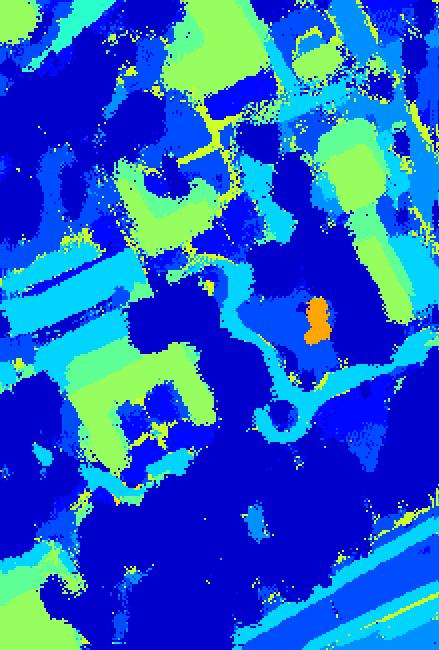}
		\caption{MFT}
		% \label{Fig6J}
	\end{subfigure}
\caption{(a) Ground truth and classification maps obtained by HS image and LiDAR data for the MUUFL data set by: (b) KNN, (c) RF, (d) CNN1D, (e) CNN2D, (f) CNN3D, (g) RNN, (h) ViT, (i) SpectralFormer, and (j) MFT using disjoint training samples.}
\label{fig:comparativeMUUFL}
\end{figure*}

%%%%%%%%%%%%%%%%%%%%%%%%%%%%%%%%%%%%%%%%%%%%%%%%%

Table \ref{tab:UT_HSI} and \ref{tab:UT_HSI} show the experimental results for Trento and MUUFL datasets using combined HS image and LiDAR data. In Trento, except for RF, it is quite evident from the better accuracies of all the models that using LiDAR and HS images improves all models' performance. But, even with this significant increase in performance, the proposed model (in Tables \ref{tab:UT_HSI} and \ref{tab:MUUFL_HSI}) performs better than the rest of the models while the proposed model (in Tables \ref{tab:UT_HSI} and \ref{tab:UT_HSI}) performs even better. For the Trento data using HS image and LiDAR, the gain in the proposed model's OA is $0.56\%$, AA is $0.07\%$, and $\kappa$ is $0.75\%$. For MUUFL using HS image and LiDAR data, the gain in the proposed model's OA, AA, and $\kappa$ are $0.26\%$,  $0.48\%$, and $0.21\%$, respectively. Among conventional classifiers and classical networks, SVM and CNN3D perform the best for Trento, while MUUFL, RF and CNN2D perform the best.

%%%%%%%%%%%%%%%%%%%%%%%%%%%%%%%%%%%%%%%%%%%%%%%%%
\begin{table*}[!t]
\centering
\caption{OA, AA and Kappa values on the Augsburg dataset (in \%) by considering HS image data only.}
\resizebox{\linewidth}{!}{\color{black}
% Please add the following required packages to your document preamble:
% \usepackage{multirow}
\begin{tabular}{c||ccc|cccc||ccc}
\hline 
\multirow{2}{*}{\begin{tabular}[c]{@{}c@{}}\textbf{Class}\\ \textbf{No.}\end{tabular}} 
& \multicolumn{3}{c|}{\textbf{Conventional Classifiers}}            
& \multicolumn{4}{c||}{\textbf{Classical Convolutional Networks}}
& \multicolumn{3}{c}{\textbf{Transformer Networks}}  \\ \cline{2-11}
& \multicolumn{1}{c|}{\textbf{KNN}} & \multicolumn{1}{c|}{\textbf{RF}} & \textbf{SVM} 
& \multicolumn{1}{c|}{\textbf{CNN1D}} 
& \multicolumn{1}{c|}{\textbf{CNN2D}} 
& \multicolumn{1}{l|}{\textbf{CNN3D}} 
& \textbf{RNN} 
& \multicolumn{1}{c|}{\textbf{ViT}} & \multicolumn{1}{c|}{\textbf{SpectralFormer}} 
& \textbf{MFT} \\ \hline \hline
1 &83.27    &90.18 $ \pm $ 00.34    &91.51    &77.79 $ \pm $ 04.77    &82.77 $ \pm $ 02.64    &78.66 $ \pm $ 01.26    &21.21 $ \pm $ 29.99    &88.62 $ \pm $ 03.36    &83.23 $ \pm $ 04.20    &\textbf{93.93 $ \pm $ 00.62}    \\ %\hline
2 &86.38    &97.41 $ \pm $ 00.14    &86.18    &87.40 $ \pm $ 01.55    &86.19 $ \pm $ 01.41    &96.10 $ \pm $ 00.98    &\textbf{98.99 $ \pm $ 01.39}    &94.90 $ \pm $ 00.84    &86.63 $ \pm $ 02.81    &96.97 $ \pm $ 00.49    \\
3 &36.79    &04.59 $ \pm $ 00.33    &10.34    &30.30 $ \pm $ 08.46    &56.08 $ \pm $ 00.99    &49.25 $ \pm $ 23.59    &00.00 $ \pm $ 00.00    &68.15 $ \pm $ 03.93    &27.75 $ \pm $ 05.25    &\textbf{68.34 $ \pm $ 01.97}    \\
4 &49.72    &76.16 $ \pm $ 00.99    &62.75    &65.25 $ \pm $ 01.80    &64.95 $ \pm $ 02.05    &84.96 $ \pm $ 01.88    &05.35 $ \pm $ 07.56    &84.40 $ \pm $ 02.26    &60.96 $ \pm $ 04.83    &\textbf{93.14 $ \pm $ 01.80}    \\
5 &38.62    &18.16 $ \pm $ 00.62    &23.14    &15.74 $ \pm $ 04.72    &10.26 $ \pm $ 02.98    &10.13 $ \pm $ 09.16    &00.00 $ \pm $ 00.00    &34.29 $ \pm $ 02.60    &37.54 $ \pm $ 07.18    &\textbf{46.02 $ \pm $ 05.76}    \\
6 &08.30    &00.12 $ \pm $ 00.09    &00.00    &\textbf{20.35 $ \pm $ 13.47}    &14.12 $ \pm $ 08.17    &08.45 $ \pm $ 03.66    &00.00 $ \pm $ 00.00    &17.83 $ \pm $ 05.12    &03.24 $ \pm $ 00.90    &14.59 $ \pm $ 04.29    \\
7 &05.37    &08.03 $ \pm $ 00.99    &10.68    &14.36 $ \pm $ 02.30    &24.62 $ \pm $ 02.92    &11.70 $ \pm $ 04.50    &00.00 $ \pm $ 00.00    &\textbf{45.90 $ \pm $ 00.92}    &13.49 $ \pm $ 08.87    &26.45 $ \pm $ 15.31    \\ \hline \hline
OA   &67.27    &79.96 $ \pm $ 00.23    &71.60    &72.00 $ \pm $ 01.70    &73.59 $ \pm $ 00.59    &82.89 $ \pm $ 00.78    &40.26 $ \pm $ 02.03    &85.90 $ \pm $ 00.26    &70.81 $ \pm $ 01.47    &\textbf{90.26 $ \pm $ 00.91}    \\ %\hline
AA  &44.07    &42.09 $ \pm $ 00.21    &40.66    &44.45 $ \pm $ 01.41    &48.43 $ \pm $ 01.99    &48.46 $ \pm $ 04.17    &14.94 $ \pm $ 00.86    &62.01 $ \pm $ 01.18    &44.69 $ \pm $ 02.31    &\textbf{62.78 $ \pm $ 02.31}    \\ %\hline
$\kappa (\times 100)$   &53.86    &70.04 $ \pm $ 00.33    &58.57    &59.96 $ \pm $ 02.40    &62.16 $ \pm $ 01.10    &75.08 $ \pm $ 01.35    &02.65 $ \pm $ 03.60    &79.88 $ \pm $ 00.27    &58.05 $ \pm $ 02.01    &\textbf{85.98 $ \pm $ 01.27}    \\ \hline
\end{tabular}}
\label{tab:AUG_HSI}
\end{table*}

%%%%%%%%%%%%%%%%%%%%%%%%%%%%%%%%%%%%%%%%%%%%%%%%%

%%%%%%%%%%%%%%%%%%%%%%%%%%%%%%%%%%%%%%%%%%%%%%%%%
\begin{table*}[!t]
\centering
\caption{OA, AA and Kappa values on the Augsburg dataset (in \%) by considering HS image and SAR data.}
\resizebox{\linewidth}{!}{\color{black}
% Please add the following required packages to your document preamble:
% \usepackage{multirow}
\begin{tabular}{c||ccc|cccc||ccc}
\hline 
\multirow{2}{*}{\begin{tabular}[c]{@{}c@{}}\textbf{Class}\\ \textbf{No.}\end{tabular}} 
& \multicolumn{3}{c|}{\textbf{Conventional Classifiers}}            
& \multicolumn{4}{c||}{\textbf{Classical Convolutional Networks}}
& \multicolumn{3}{c}{\textbf{Transformer Networks}}  \\ \cline{2-11}
& \multicolumn{1}{c|}{\textbf{KNN}} & \multicolumn{1}{c|}{\textbf{RF}} & \textbf{SVM} 
& \multicolumn{1}{c|}{\textbf{CNN1D}} 
& \multicolumn{1}{c|}{\textbf{CNN2D}} 
& \multicolumn{1}{l|}{\textbf{CNN3D}} 
& \textbf{RNN} 
& \multicolumn{1}{c|}{\textbf{ViT}} & \multicolumn{1}{c|}{\textbf{SpectralFormer}} 
& \textbf{MFT} \\ \hline \hline
1 &81.98    &87.15 $ \pm $ 00.77    &91.51    &81.26 $ \pm $ 02.95    &90.96 $ \pm $ 01.32    &78.21 $ \pm $ 00.56    &87.60 $ \pm $ 06.19    &90.01 $ \pm $ 00.48    &92.04 $ \pm $ 01.87    &\textbf{94.65 $ \pm $ 00.62}    \\ %\hline
2 &86.17    &95.82 $ \pm $ 00.10    &86.32    &88.32 $ \pm $ 01.78    &82.07 $ \pm $ 00.74    &93.82 $ \pm $ 00.67    &95.26 $ \pm $ 02.62    &94.27 $ \pm $ 00.78    &93.60 $ \pm $ 02.29    &\textbf{96.90 $ \pm $ 00.62}    \\
3 &29.66    &64.62 $ \pm $ 02.34    &10.03    &42.43 $ \pm $ 14.40    &48.31 $ \pm $ 10.85    &63.11 $ \pm $ 11.45    &51.35 $ \pm $ 72.62    &64.62 $ \pm $ 05.57    &43.28 $ \pm $ 07.08    &\textbf{69.80 $ \pm $ 06.43}    \\
4 &51.28    &84.33 $ \pm $ 00.80    &62.78    &67.42 $ \pm $ 02.97    &61.05 $ \pm $ 04.86    &86.24 $ \pm $ 01.07    &91.14 $ \pm $ 02.98    &85.53 $ \pm $ 00.91    &77.08 $ \pm $ 04.11    &\textbf{93.98 $ \pm $ 01.65}    \\
5 &36.90    &27.72 $ \pm $ 01.02    &22.75    &09.88 $ \pm $ 06.07    &18.48 $ \pm $ 07.96    &11.15 $ \pm $ 08.03    &10.52 $ \pm $ 11.42    &29.00 $ \pm $ 00.63    &\textbf{45.32 $ \pm $ 05.54}    &32.70 $ \pm $ 11.81    \\
6 &13.80    &13.06 $ \pm $ 02.27    &00.00    &10.26 $ \pm $ 05.48    &12.11 $ \pm $ 02.80    &09.67 $ \pm $ 04.92    &00.00 $ \pm $ 00.00    &\textbf{20.41 $ \pm $ 01.26}    &06.96 $ \pm $ 01.01    &10.52 $ \pm $ 04.31    \\
7 &07.37    &11.44 $ \pm $ 00.39    &11.28    &09.93 $ \pm $ 03.22    &29.13 $ \pm $ 01.00    &09.42 $ \pm $ 05.13    &24.33 $ \pm $ 34.41    &\textbf{44.17 $ \pm $ 02.06}    &29.44 $ \pm $ 04.77    &23.98 $ \pm $ 07.85    \\ \hline \hline
OA   &67.29    &85.00 $ \pm $ 00.28    &71.65    &73.96 $ \pm $ 01.17    &71.79 $ \pm $ 00.65    &83.04 $ \pm $ 00.47    &83.42 $ \pm $ 01.14    &86.10 $ \pm $ 00.26    &81.77 $ \pm $ 02.30    &\textbf{90.49 $ \pm $ 00.20}    \\ %\hline
AA  &43.88    &54.88 $ \pm $ 00.28    &40.67    &44.21 $ \pm $ 02.81    &48.88 $ \pm $ 02.61    &50.23 $ \pm $ 01.09    &40.75 $ \pm $ 02.04    &\textbf{61.14 $ \pm $ 00.85}    &55.39 $ \pm $ 01.98    &60.36 $ \pm $ 02.36    \\ %\hline
$\kappa (\times 100)$   &54.13    &78.62 $ \pm $ 00.38    &58.62    &62.20 $ \pm $ 01.14    &59.84 $ \pm $ 00.63    &75.49 $ \pm $ 00.62    &75.26 $ \pm $ 01.90    &80.06 $ \pm $ 00.33    &74.06 $ \pm $ 03.04    &\textbf{86.26 $ \pm $ 00.23}    \\ \hline
\end{tabular}}
\label{tab:AUG_HS}
\end{table*}

%%%%%%%%%%%%%%%%%%%%%%%%%%%%%%%%%%%%%%%%%%%%%%%%%
%%%%%%%%%%%%%%%%%%%%%%%%%%%%%%%%%%%%%%%%%%%%%%%%%
\begin{figure*}[!t]
\centering
	\begin{subfigure}{0.35\columnwidth}
		\includegraphics[width=0.99\textwidth]{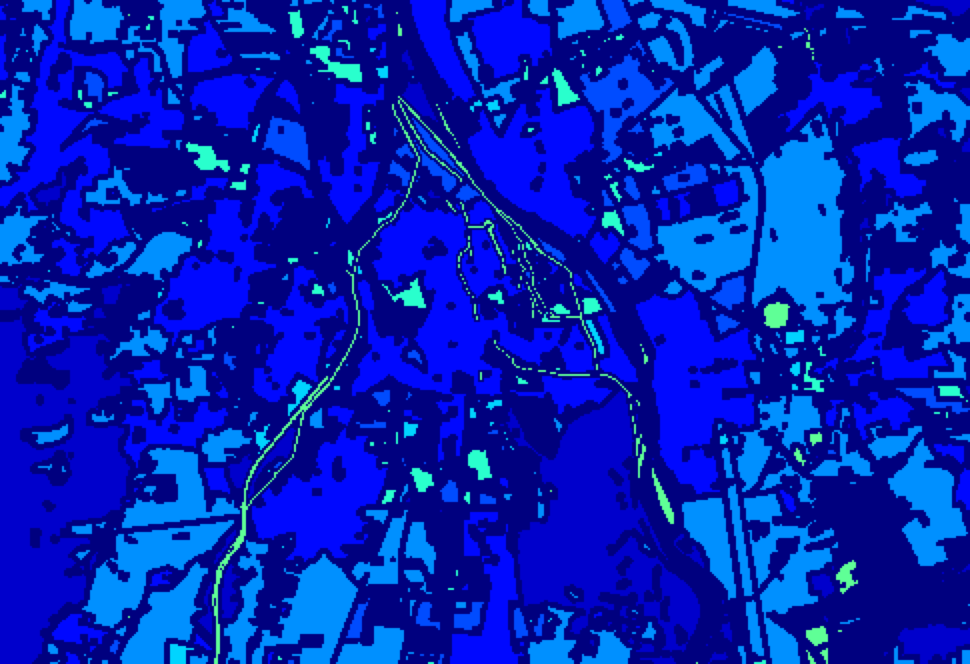}
		\caption{GT}
% 		\label{Fig6A}
	\end{subfigure}
	\begin{subfigure}{0.35\columnwidth}
		\includegraphics[width=0.99\textwidth]{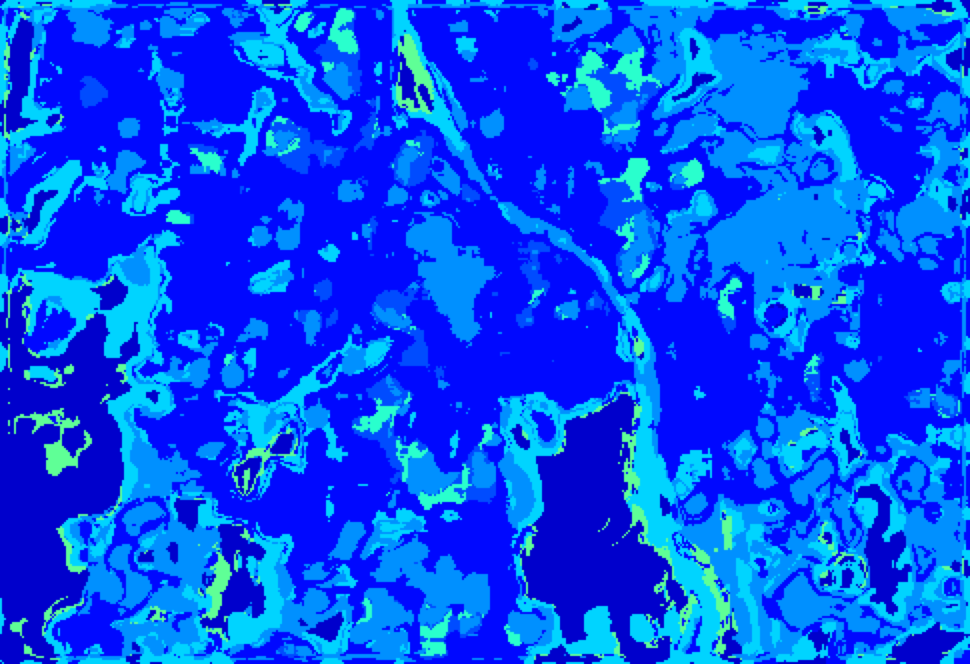}
		\caption{KNN}
% 		\label{Fig6A}
	\end{subfigure}
	\begin{subfigure}{0.35\columnwidth}
		\includegraphics[width=0.99\textwidth]{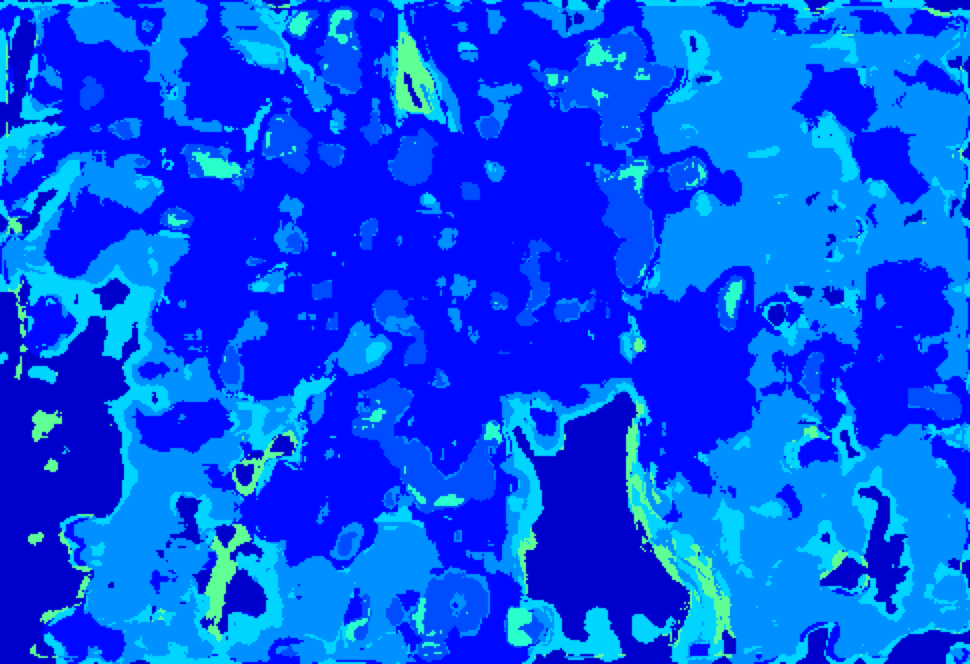}
		\caption{RF}
% 		\label{Fig6B}
	\end{subfigure}
	\begin{subfigure}{0.35\columnwidth}
		\includegraphics[width=0.99\textwidth]{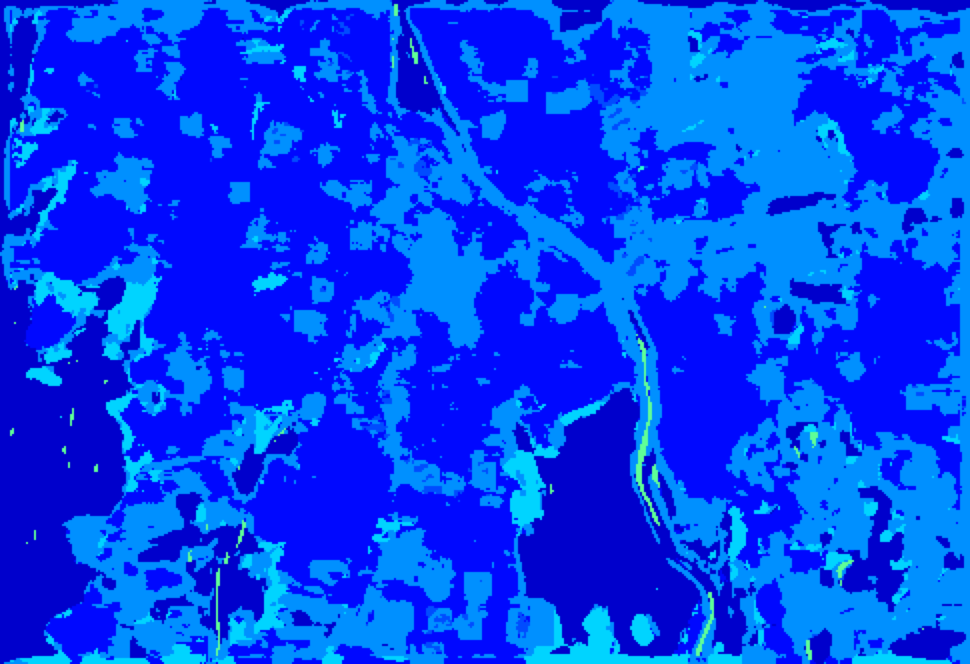}
		\caption{SVM} 
		\label{Fig6C}
	\end{subfigure}
	\begin{subfigure}{0.35\columnwidth}
		\includegraphics[width=0.99\textwidth]{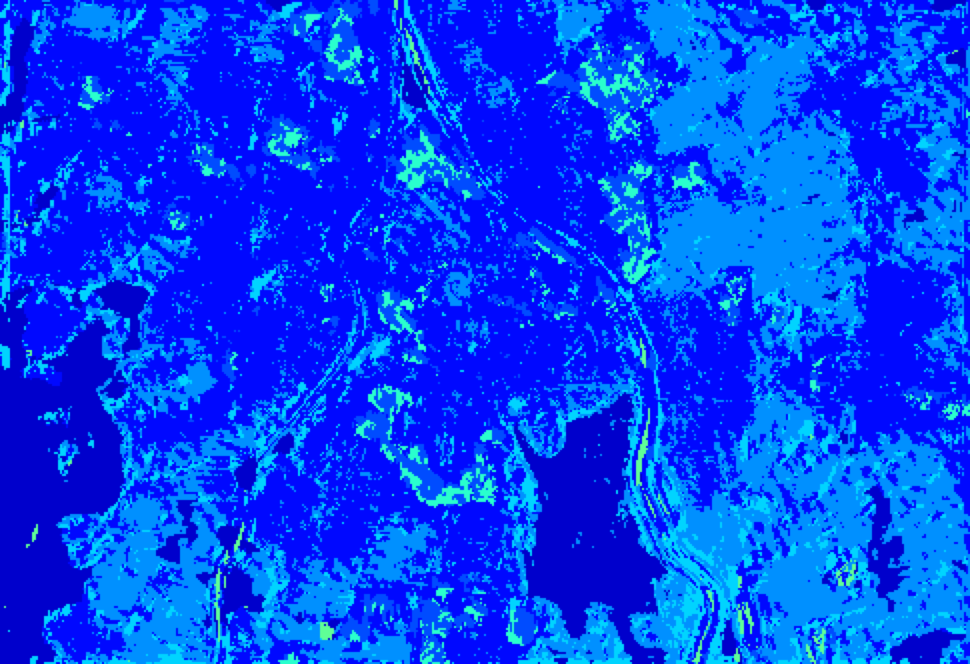}
		\caption{CNN1D} 
% 		\label{Fig6D}
	\end{subfigure}
	\begin{subfigure}{0.35\columnwidth}
		\includegraphics[width=0.99\textwidth]{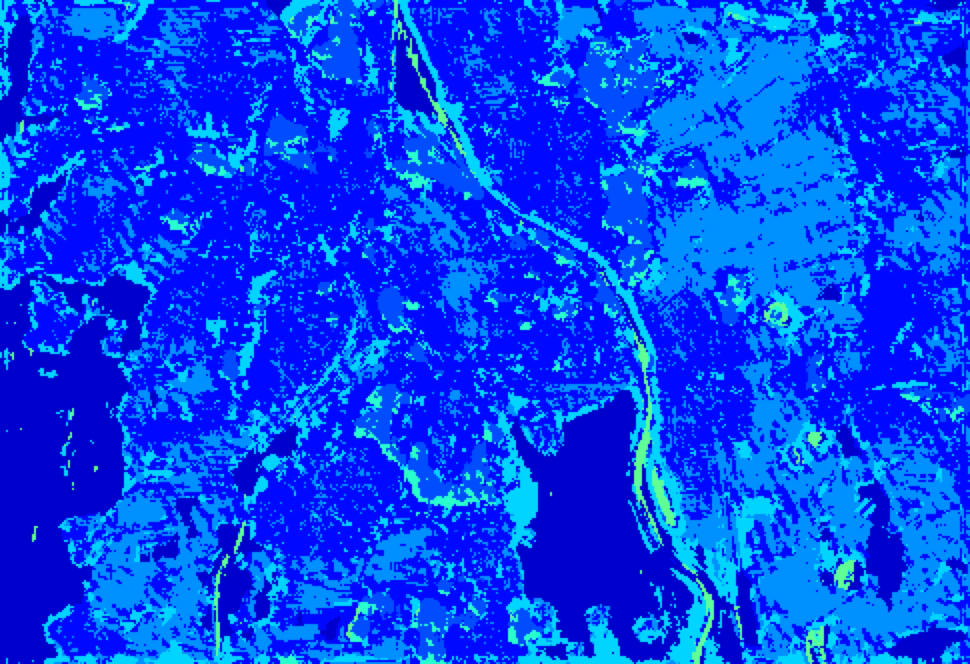}
		\caption{CNN2D}
% 		\label{Fig6E}
	\end{subfigure}
    \begin{subfigure}{0.35\columnwidth}
		\includegraphics[width=0.99\textwidth]{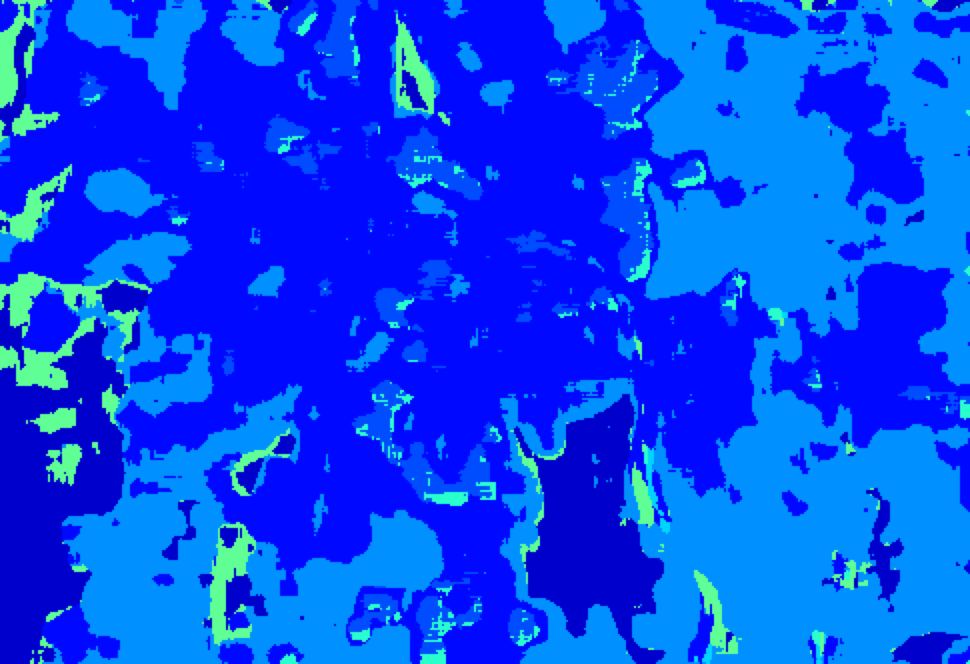}
		\caption{CNN3D}
% 		\label{Fig6F}
	\end{subfigure}
% 	\begin{subfigure}{0.35\columnwidth}
% 		\includegraphics[width=0.99\textwidth]{clsfmaps/Augsburg_H+SAR/AugsburgSAR_RNNPytorch_H+L_Deviation.png}
% 		\caption{RNN}
% 		\label{Fig6G}
% 	\end{subfigure}
	\begin{subfigure}{0.35\columnwidth}
		\includegraphics[width=0.99\textwidth]{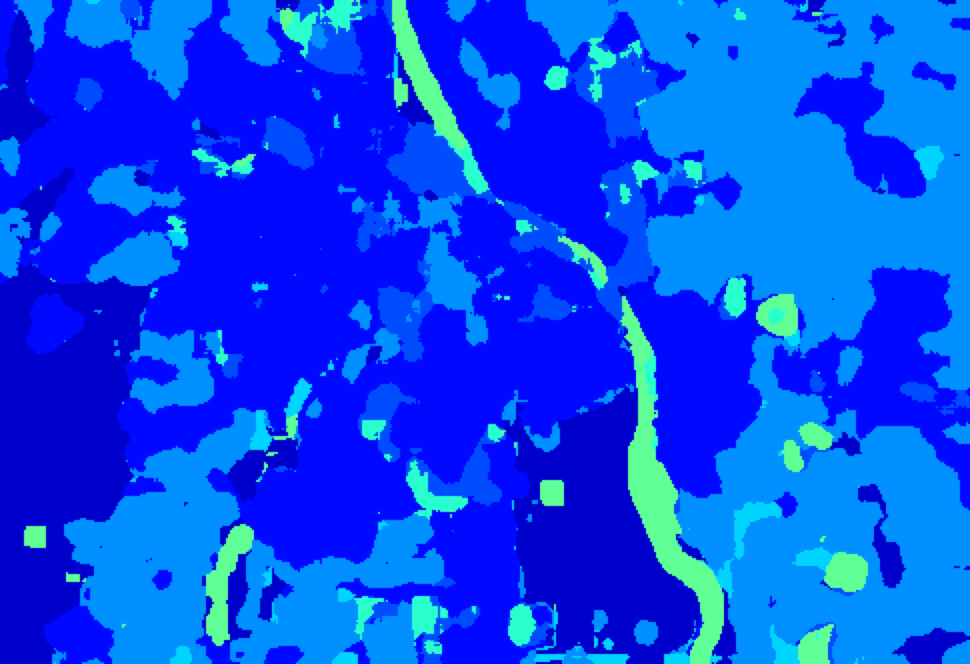}
		\caption{ViT} 
		\label{Fig6H}
	\end{subfigure}
	\begin{subfigure}{0.35\columnwidth}
		\includegraphics[width=0.99\textwidth]{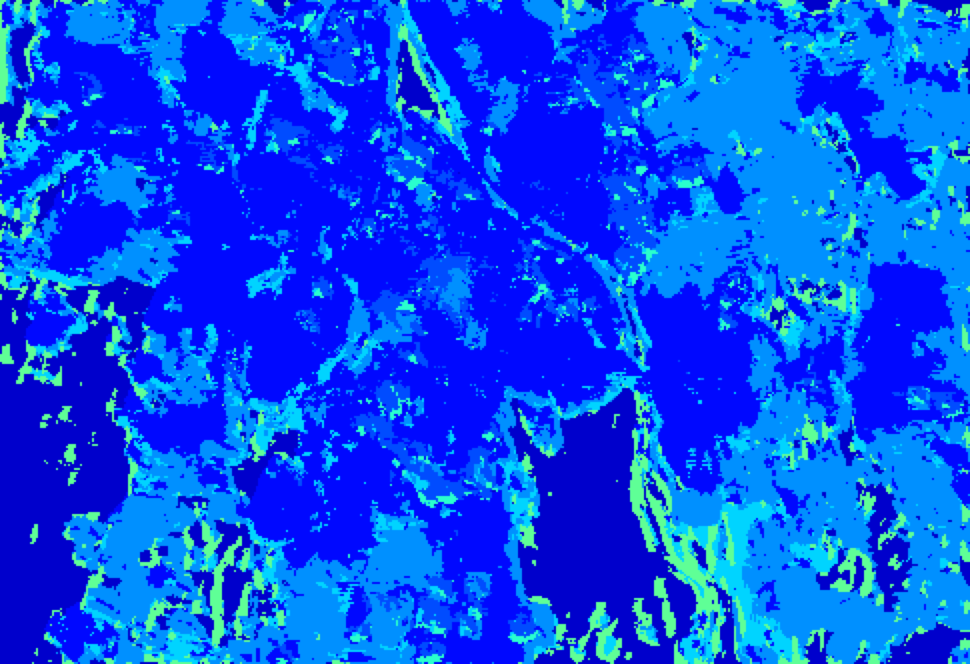}
		\caption{SpectralFormer}
% 		\label{Fig6I}
	\end{subfigure}
	\begin{subfigure}{0.35\columnwidth}
		\includegraphics[width=0.99\textwidth]{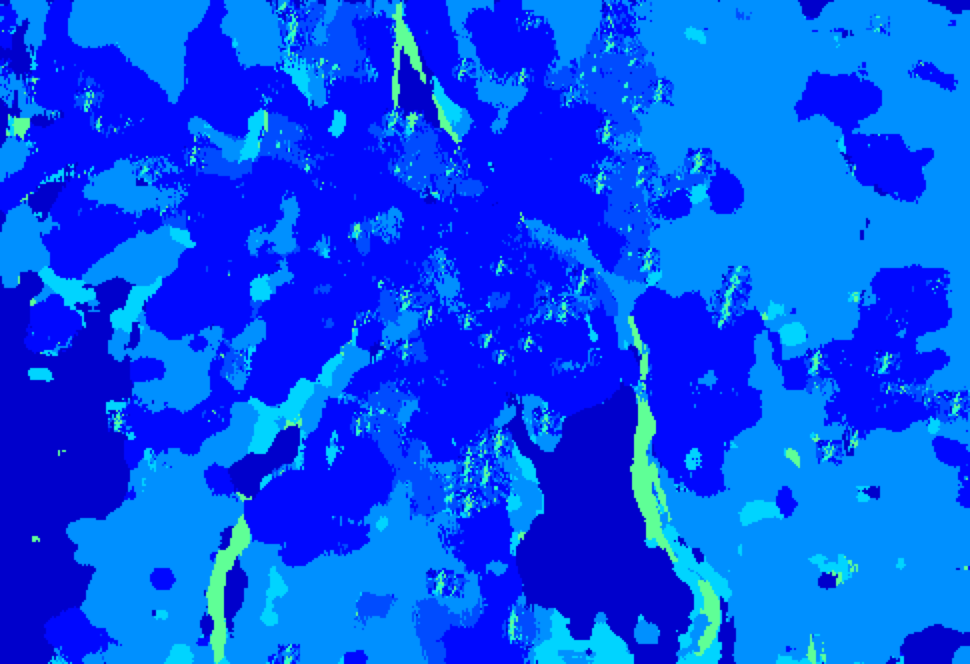}
		\caption{MFT}
% 		\label{Fig6J}
	\end{subfigure}
\caption{(a) Ground truth and classification maps obtained by HSI and SAR data for the Augsburg data set by: (b) KNN, (c) RF, (d) CNN1D, (e) CNN2D, (f) CNN3D, (g) ViT, (h) SpectralFormer, and (i) MFT using disjoint training samples.}
\label{fig:comparativeAugSAR}
\end{figure*}

%%%%%%%%%%%%%%%%%%%%%%%%%%%%%%%%%%%%%%%%%%%%%%%%%

The proposed MFT is not only limited to HS image, MS image, and LiDAR data but can also work with SAR and DSM data. To illustrate this capability of the proposed MFT,
we have considered the Augsburg dataset (using HS image data only, HS image and SAR data, and HS image and DSM data) with disjoint training and test samples. Fig.~\ref{fig:Augsburg} shows the pseudo color map, corresponding SAR data, corresponding DSM data, and the disjoint train and test ground truth for the Augsburg dataset. It also provides the $7$ different land-cover classes and the number of samples belonging to the train and test set per class.

After observing the AA in Table \ref{tab:AUG_HSI}, it is evident that Augsburg is a considerably harder dataset than the previous ones. Consequently, the gain in performance in most models with fusion is not significant. However, the proposed model gives the highest class-wise classification accuracies for most classes in all three cases. It also provides the highest OAs, AAs, and $\kappa$s with the exception in the case of the Augsburg dataset using HS and SAR data. Table \ref{tab:AUG_HS} indicates that ViT has $0.76\%$ more AA than the proposed model. However, the proposed MFT has $4.39\%$ and $6.20\%$ more OA and $\kappa$ than ViT, respectively. In Table \ref{tab:AUG_HD}, the AA of the proposed MFT reaches $64.70 \pm 00.44\%$ which is the highest recorded for all three cases in all models.

RNN performs the worst in Tables \ref{tab:AUG_HSI} and \ref{tab:AUG_HD} but gets a significant boost in performance in Table \ref{tab:AUG_HS}. Comparing Tables \ref{tab:AUG_HS} and \ref{tab:AUG_HD}, conventional classifiers like KNN and SVM show barely noticeable changes in performance. On the other hand, RF gives higher accuracy using SAR data than DSM data. In the case of classical networks, CNN1D performs better-using DSM, CNN2D and RNN perform better using SAR, and CNN3D performs similarly in both cases.

%%%%%%%%%%%%%%%%%%%%%%%%%%%%%%%%%%%%%%%%%%%%%%%%%
\begin{table*}[!ht]
\centering
\caption{OA, AA and Kappa values on the Augsburg dataset (in \%) by considering HS image and DSM data.}
\resizebox{\linewidth}{!}{\color{black}
% Please add the following required packages to your document preamble:
% \usepackage{multirow}
\begin{tabular}{c||ccc|cccc||ccc}
\hline 
\multirow{2}{*}{\begin{tabular}[c]{@{}c@{}}\textbf{Class}\\ \textbf{No.}\end{tabular}} 
& \multicolumn{3}{c|}{\textbf{Conventional Classifiers}}            
& \multicolumn{4}{c||}{\textbf{Classical Convolutional Networks}}
& \multicolumn{3}{c}{\textbf{Transformer Networks}}  \\ \cline{2-11}
& \multicolumn{1}{c|}{\textbf{KNN}} & \multicolumn{1}{c|}{\textbf{RF}} & \textbf{SVM} 
& \multicolumn{1}{c|}{\textbf{CNN1D}} 
& \multicolumn{1}{c|}{\textbf{CNN2D}} 
& \multicolumn{1}{l|}{\textbf{CNN3D}} 
& \textbf{RNN} 
& \multicolumn{1}{c|}{\textbf{ViT}} & \multicolumn{1}{c|}{\textbf{SpectralFormer}} 
& \textbf{MFT} \\ \hline \hline
1 &81.99    &84.94 $ \pm $ 00.97    &91.51    &87.75 $ \pm $ 00.49    &78.92 $ \pm $ 01.14    &77.41 $ \pm $ 04.03    &00.00 $ \pm $ 00.00    &90.39 $ \pm $ 00.85    &90.12 $ \pm $ 02.42    &\textbf{94.77 $ \pm $ 02.29}    \\ %\hline
2 &86.16    &88.48 $ \pm $ 00.06    &86.21    &93.69 $ \pm $ 00.22    &85.50 $ \pm $ 01.49    &95.47 $ \pm $ 00.93    &\textbf{99.92 $ \pm $ 00.12}    &95.04 $ \pm $ 01.07    &86.98 $ \pm $ 01.18    &96.03 $ \pm $ 01.84    \\
3 &29.66    &64.50 $ \pm $ 02.06    &10.23    &28.96 $ \pm $ 04.51    &56.41 $ \pm $ 19.63    &61.22 $ \pm $ 06.55    &00.00 $ \pm $ 00.00    &59.49 $ \pm $ 03.02    &37.31 $ \pm $ 03.93    &\textbf{72.52 $ \pm $ 10.54}    \\
4 &51.29    &73.27 $ \pm $ 00.55    &62.80    &89.25 $ \pm $ 00.62    &57.72 $ \pm $ 02.86    &86.15 $ \pm $ 00.07    &16.91 $ \pm $ 23.92    &86.34 $ \pm $ 01.78    &58.17 $ \pm $ 11.82    &\textbf{93.52 $ \pm $ 01.96}    \\
5 &36.90    &26.83 $ \pm $ 00.59    &22.94    &18.61 $ \pm $ 01.86    &10.45 $ \pm $ 04.10    &09.31 $ \pm $ 02.24    &00.00 $ \pm $ 00.00    &21.16 $ \pm $ 14.48    &47.74 $ \pm $ 08.32    &\textbf{51.31 $ \pm $ 05.13}    \\
6 &13.80    &\textbf{17.97 $ \pm $ 01.99}    &00.00    &15.85 $ \pm $ 05.68    &08.85 $ \pm $ 08.48    &09.22 $ \pm $ 03.42    &00.00 $ \pm $ 00.00    &15.53 $ \pm $ 11.50    &02.32 $ \pm $ 00.18    &13.49 $ \pm $ 09.44    \\
7 &07.37    &10.20 $ \pm $ 00.17    &10.75    &23.60 $ \pm $ 01.14    &23.65 $ \pm $ 04.78    &09.67 $ \pm $ 01.79    &00.00 $ \pm $ 00.00    &\textbf{44.72 $ \pm $ 01.77}    &20.86 $ \pm $ 03.16    &31.28 $ \pm $ 10.80    \\ \hline \hline
OA   &67.29    &78.04 $ \pm $ 00.07    &71.62    &84.43 $ \pm $ 00.18    &70.07 $ \pm $ 01.11    &83.40 $ \pm $ 00.76    &38.80 $ \pm $ 00.04    &86.34 $ \pm $ 00.40    &71.84 $ \pm $ 03.87    &\textbf{90.49 $ \pm $ 00.54}    \\ %\hline
AA  &43.88    &52.31 $ \pm $ 00.17    &40.64    &51.10 $ \pm $ 00.61    &45.93 $ \pm $ 02.28    &49.78 $ \pm $ 00.59    &14.30 $ \pm $ 00.02    &58.95 $ \pm $ 03.61    &49.07 $ \pm $ 01.38    &\textbf{64.70 $ \pm $ 00.44}    \\ %\hline
$\kappa (\times 100)$   &54.14    &68.91 $ \pm $ 00.09    &58.60    &77.21 $ \pm $ 00.26    &57.06 $ \pm $ 01.91    &75.90 $ \pm $ 01.14    &06.22 $ \pm $ 08.80    &80.31 $ \pm $ 00.68    &60.15 $ \pm $ 04.79    &\textbf{86.35 $ \pm $ 00.77}    \\ \hline
\end{tabular}}
\label{tab:AUG_HD}
\end{table*}

%%%%%%%%%%%%%%%%%%%%%%%%%%%%%%%%%%%%%%%%%%%%%%%%%

%%%%%%%%%%%%%%%%%%%%%%%%%%%%%%%%%%%%%%%%%%%%%%%%%
\begin{figure*}[!t]
\centering
	\begin{subfigure}{0.36\columnwidth}
		\includegraphics[width=0.99\textwidth]{clsfmaps/Augsburg_H+SAR/Augsburg_gt.png}
		\caption{GT}
% 		\label{Fig6A}
	\end{subfigure}
	\begin{subfigure}{0.36\columnwidth}
		\includegraphics[width=0.99\textwidth]{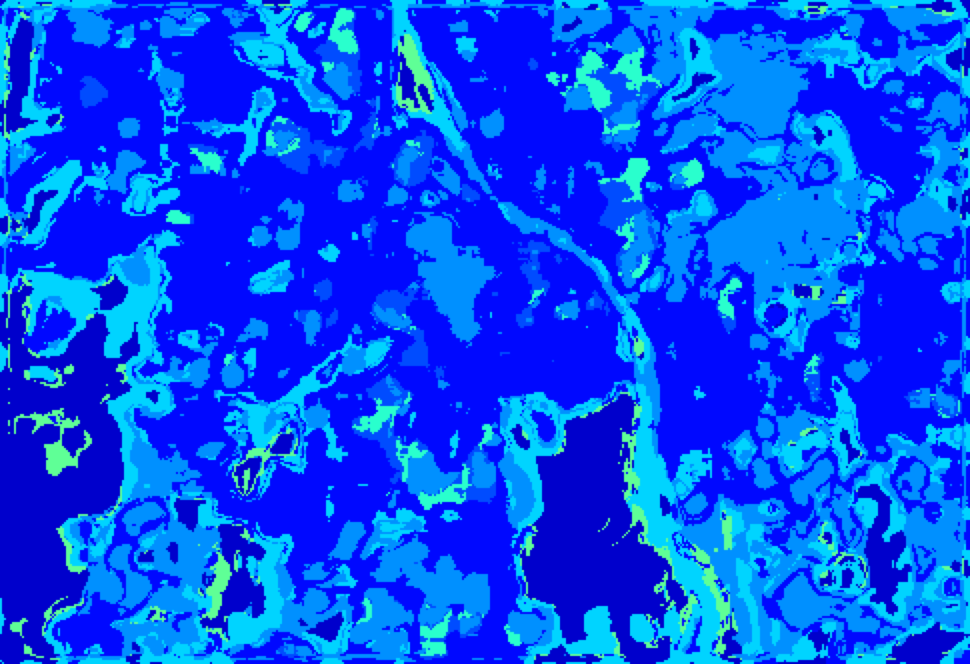}
		\caption{KNN}
% 		\label{Fig6A}
	\end{subfigure}
	\begin{subfigure}{0.36\columnwidth}
		\includegraphics[width=0.99\textwidth]{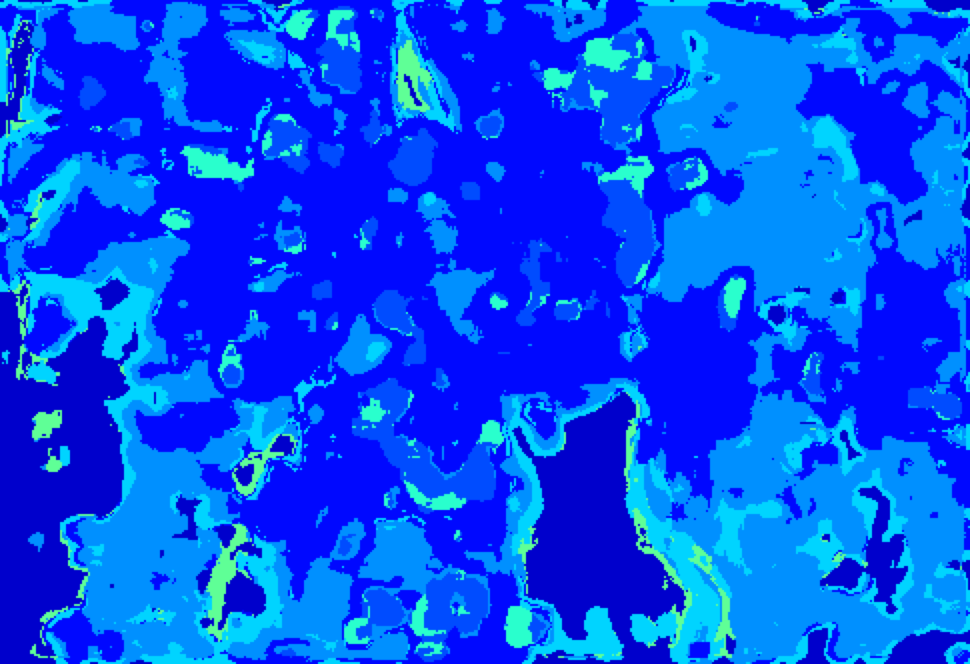}
		\caption{RF}
% 		\label{Fig6B}
	\end{subfigure}
	\begin{subfigure}{0.36\columnwidth}
		\includegraphics[width=0.99\textwidth]{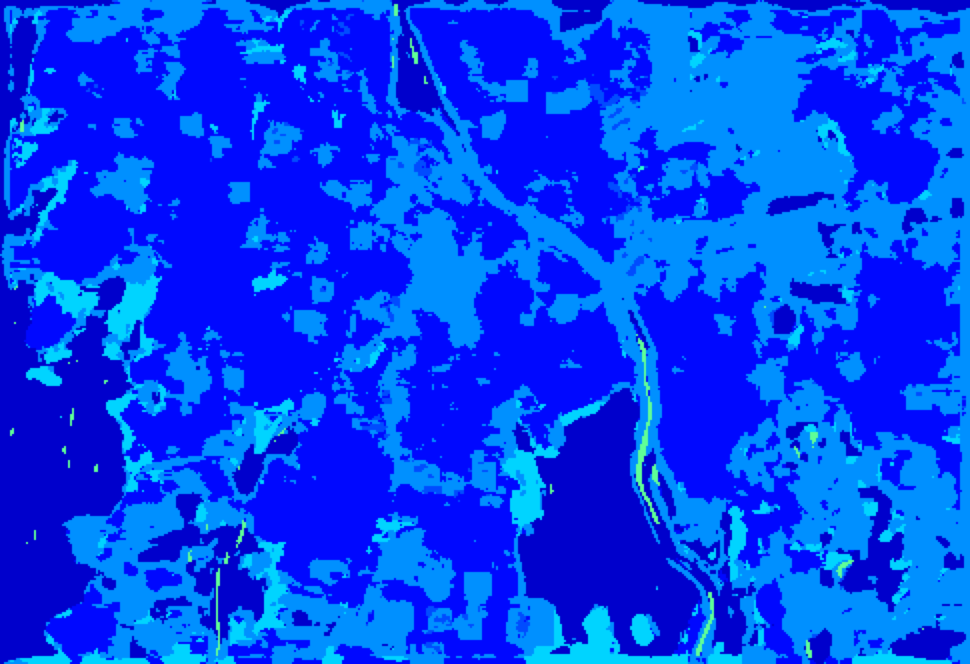}
		\caption{SVM} 
% 		\label{Fig6C}
	\end{subfigure}
	\begin{subfigure}{0.36\columnwidth}
		\includegraphics[width=0.99\textwidth]{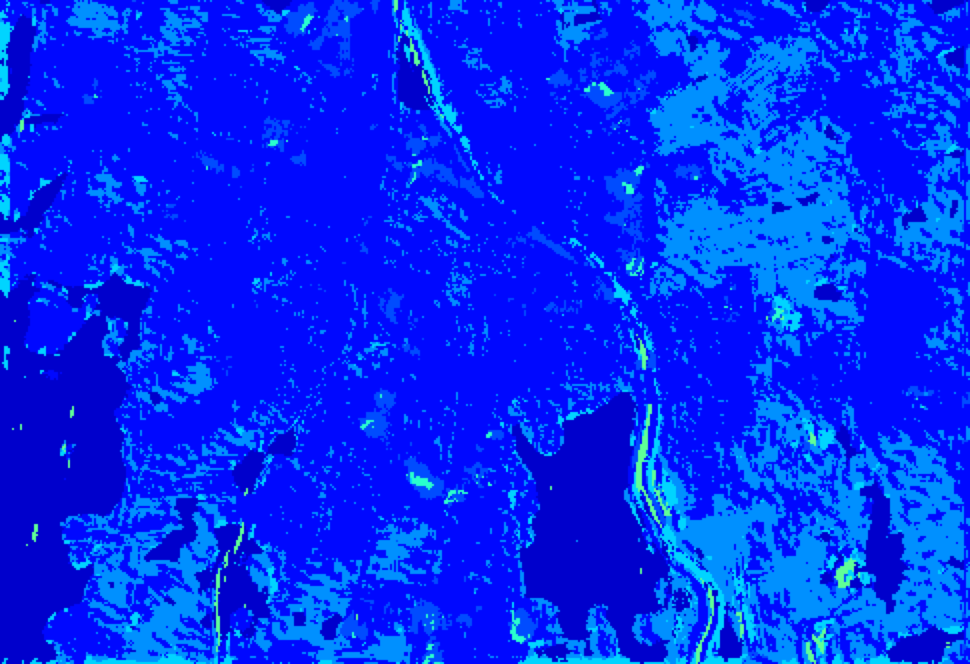}
		\caption{CNN1D} 
% 		\label{Fig6D}
	\end{subfigure}
	\begin{subfigure}{0.36\columnwidth}
		\includegraphics[width=0.99\textwidth]{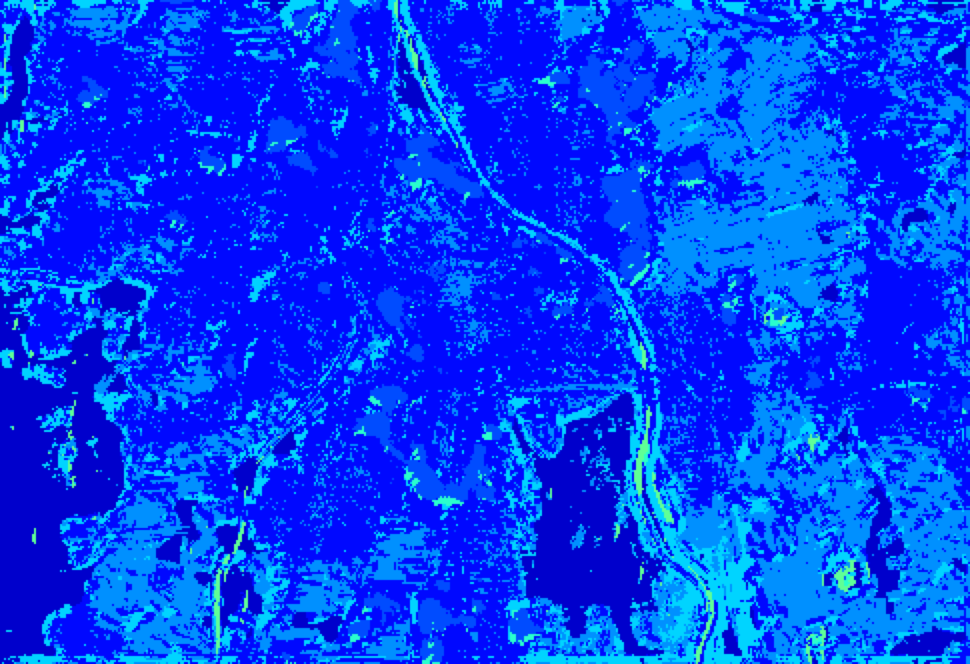}
		\caption{CNN2D}
% 		\label{Fig6E}
	\end{subfigure}
    \begin{subfigure}{0.36\columnwidth}
		\includegraphics[width=0.99\textwidth]{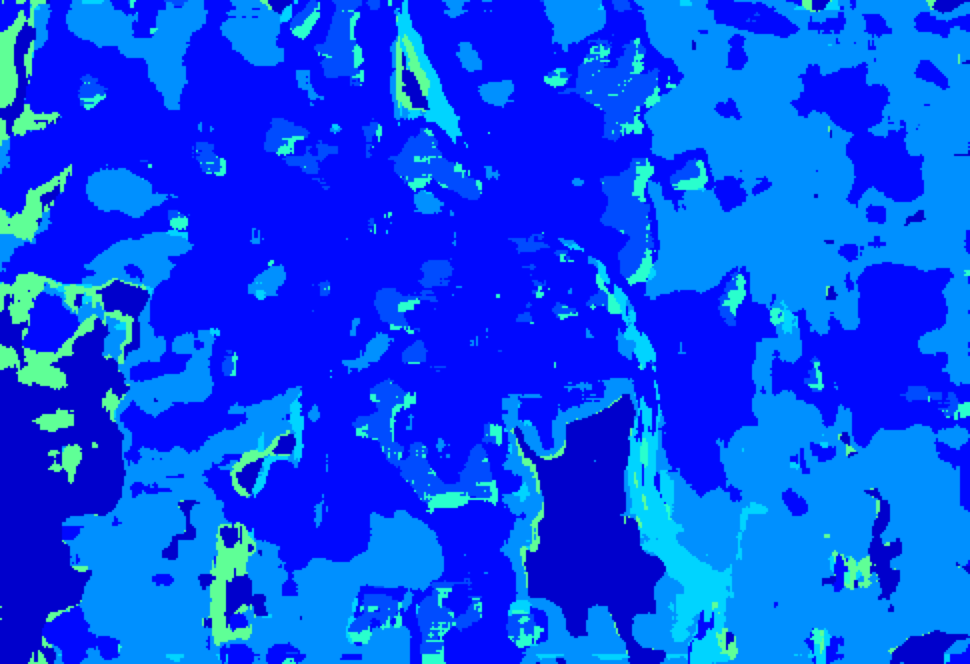}
		\caption{CNN3D}
% 		\label{Fig6F}
	\end{subfigure}
% 	\begin{subfigure}{0.35\columnwidth}
% 		\includegraphics[width=0.99\textwidth]{clsfmaps/Augsburg_H+DSM/AugsburgDSM_RNNPytorch_H+L_Deviation.png}
% 		\caption{RNN}
% 		\label{Fig6G}
% 	\end{subfigure}
	\begin{subfigure}{0.36\columnwidth}
		\includegraphics[width=0.99\textwidth]{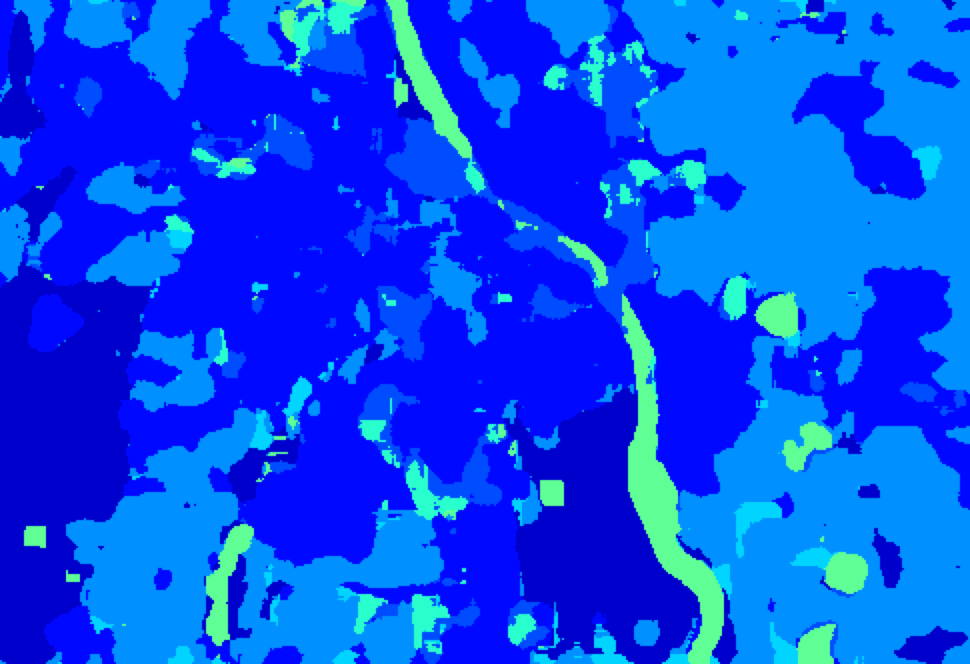}
		\caption{ViT} 
% 		\label{Fig6H}
	\end{subfigure}
	\begin{subfigure}{0.36\columnwidth}
		\includegraphics[width=0.99\textwidth]{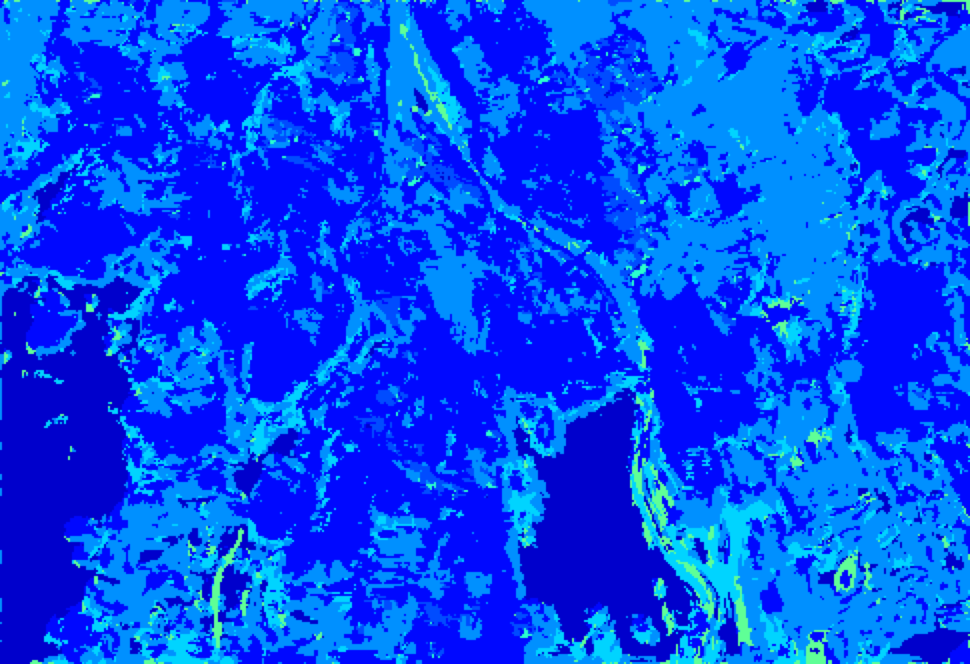}
		\caption{SpectralFormer}
% 		\label{Fig6I}
	\end{subfigure}
	\begin{subfigure}{0.36\columnwidth}
		\includegraphics[width=0.99\textwidth]{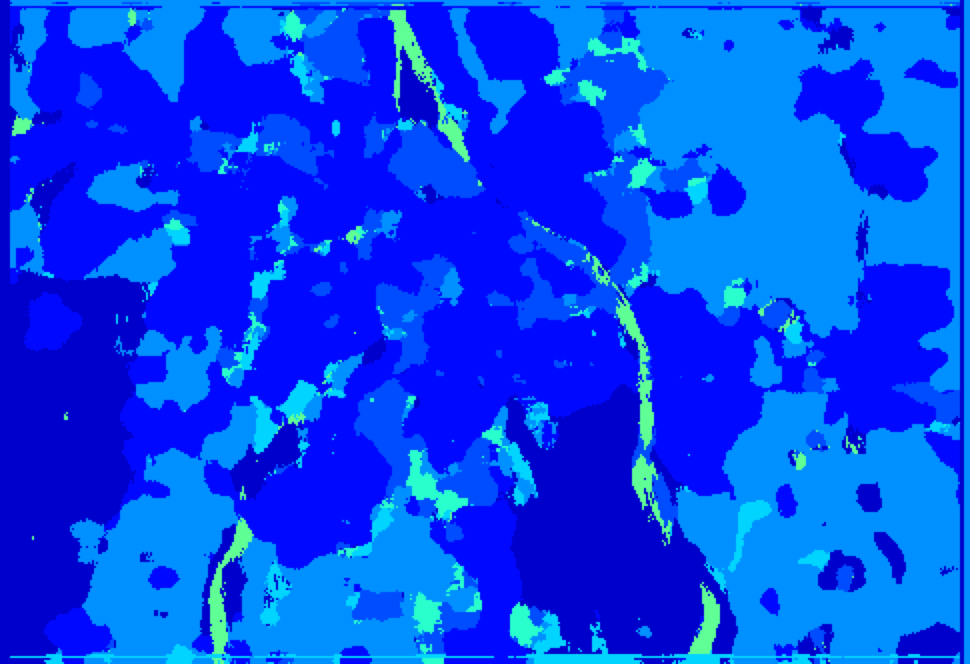}
		\caption{MFT}
% 		\label{Fig6J}
	\end{subfigure}
\caption{(a) Ground truth and classification maps obtained by HSI and DSM data for the Augsburg dataset by: (b) KNN, (c) RF, (d) CNN1D, (e) CNN2D, (f) CNN3D, (g) ViT, (h) SpectralFormer, and (i) MFT using disjoint training samples.}
\label{fig:comparativeAugDSM}
\end{figure*}
%%%%%%%%%%%%%%%%%%%%%%%%%%%%

%%%%%%%%%%%%%%%%%%%%%%%%%%%%%%%%%%%%%%%%%%%%%%%%%
\begin{table*}[!ht]
\centering
\caption{OA, AA and Kappa values achieved by MFT with different tokenization i.e. pixel tokenizer (PT) and channel tokenizer (CT) for UH (HSI+LiDAR), UT (HSU+LiDAR), MUUFL (HSI+LiDAR), UH (HSI+MS), Augsburg (HSI+SAR), and Augsburg (HSI+DSM) datasets.}
\resizebox{\linewidth}{!}{\color{black}

\begin{tabular}{c|cc|cc|cc|cc|cc|cc}
\hline
\multirow{2}{*}{\begin{tabular}[c]{@{}c@{}}Class\\ No.\end{tabular}} & \multicolumn{2}{c|}{Houston (HSI+LiDAR)}    & \multicolumn{2}{c|}{Trento (HSI+LiDAR)}    & \multicolumn{2}{c|}{MUUFL (HSI+LiDAR)}    & \multicolumn{2}{c|}{Houston (HSI+MS)}    & \multicolumn{2}{c|}{Augsburg (HSI+SAR)}    & \multicolumn{2}{c}{Augsburg (HSI+DSM)}    \\ \cline{2-13} 
 & \multicolumn{1}{c|}{PT} & CT & \multicolumn{1}{c|}{PT} & CT & \multicolumn{1}{c|}{PT} & CT & \multicolumn{1}{c|}{PT} & CT & \multicolumn{1}{c|}{PT} & CT & \multicolumn{1}{c|}{PT} & CT \\ \hline \hline
1 &82.34 $ \pm $ 00.47    &\textbf{84.90 $ \pm $ 03.02}    &\textbf{98.71 $ \pm $ 00.80}    &98.23 $ \pm $ 00.38    &97.52 $ \pm $ 00.38    &\textbf{97.90 $ \pm $ 00.39}    &82.72 $ \pm $ 00.41    &\textbf{87.75 $ \pm $ 07.12}    &94.65 $ \pm $ 00.62    &\textbf{95.53 $ \pm $ 00.69}    &\textbf{95.38 $ \pm $ 00.38}    &94.77 $ \pm $ 02.29    \\ %\hline
2 &\textbf{88.78 $ \pm $ 05.27}    &84.71 $ \pm $ 00.56    &\textbf{99.50 $ \pm $ 00.12}    &99.34 $ \pm $ 00.02    &\textbf{92.35 $ \pm $ 02.13}    &92.11 $ \pm $ 01.58    &85.09 $ \pm $ 00.09    &\textbf{85.15 $ \pm $ 00.00}    &\textbf{96.90 $ \pm $ 00.62}    &96.81 $ \pm $ 00.91    &\textbf{97.21 $ \pm $ 00.09}    &96.03 $ \pm $ 01.84    \\
3 &98.15 $ \pm $ 01.08    &\textbf{98.61 $ \pm $ 01.01}    &80.12 $ \pm $ 07.78    &\textbf{89.84 $ \pm $ 09.00}    &\textbf{93.21 $ \pm $ 01.30}    &91.80 $ \pm $ 00.82    &\textbf{98.55 $ \pm $ 00.76}    &98.42 $ \pm $ 00.32    &\textbf{69.80 $ \pm $ 06.43}    &66.98 $ \pm $ 02.92    &55.10 $ \pm $ 07.22    &\textbf{72.52 $ \pm $ 10.54}    \\
4 &94.35 $ \pm $ 02.40    &\textbf{96.88 $ \pm $ 01.45}    &\textbf{99.96 $ \pm $ 00.03}    &99.82 $ \pm $ 00.26    &\textbf{93.37 $ \pm $ 01.85}    &91.59 $ \pm $ 02.25    &\textbf{95.99 $ \pm $ 02.39}    &92.90 $ \pm $ 02.89    &93.98 $ \pm $ 01.65    &\textbf{94.31 $ \pm $ 01.81}    &92.42 $ \pm $ 00.34    &\textbf{93.52 $ \pm $ 01.96}    \\
5 &99.12 $ \pm $ 01.25    &\textbf{99.97 $ \pm $ 00.04}    &99.82 $ \pm $ 00.23    &\textbf{99.93 $ \pm $ 00.05}    &95.18 $ \pm $ 00.13    &\textbf{95.60 $ \pm $ 01.21}    &\textbf{99.78 $ \pm $ 00.09}    &99.21 $ \pm $ 00.39    &\textbf{32.70 $ \pm $ 11.81}    &24.54 $ \pm $ 06.34    &43.28 $ \pm $ 11.66    &\textbf{51.31 $ \pm $ 05.13}    \\
6 &\textbf{99.30 $ \pm $ 00.00}    &90.44 $ \pm $ 07.58    &84.29 $ \pm $ 01.40    &\textbf{88.72 $ \pm $ 00.94}    &84.65 $ \pm $ 02.48    &\textbf{88.19 $ \pm $ 03.49}    &97.20 $ \pm $ 01.98    &\textbf{97.67 $ \pm $ 01.74}    &10.52 $ \pm $ 04.31    &\textbf{12.31 $ \pm $ 02.19}    &\textbf{14.06 $ \pm $ 04.71}    &13.49 $ \pm $ 09.44    \\
7 &\textbf{88.56 $ \pm $ 01.16}    &81.84 $ \pm $ 05.59    &    &    &89.93 $ \pm $ 02.23    &\textbf{90.27 $ \pm $ 02.13}    &86.32 $ \pm $ 00.69    &\textbf{87.28 $ \pm $ 01.38}    &\textbf{23.98 $ \pm $ 07.85}    &15.28 $ \pm $ 07.14    &16.46 $ \pm $ 04.41    &\textbf{31.28 $ \pm $ 10.80}    \\
8 &86.89 $ \pm $ 05.33    &\textbf{89.78 $ \pm $ 01.83}    &    &    &97.24 $ \pm $ 00.48    &\textbf{97.26 $ \pm $ 00.53}    &\textbf{81.16 $ \pm $ 08.04}    &77.71 $ \pm $ 00.98    &    &    &    &    \\
9 &87.91 $ \pm $ 03.90    &\textbf{91.94 $ \pm $ 01.20}    &    &    &58.89 $ \pm $ 05.71    &\textbf{61.35 $ \pm $ 03.80}    &87.76 $ \pm $ 00.74    &\textbf{88.29 $ \pm $ 01.72}    &    &    &    &    \\
10 &64.70 $ \pm $ 00.92    &\textbf{65.03 $ \pm $ 12.68}    &    &    &11.69 $ \pm $ 04.51    &\textbf{17.43 $ \pm $ 04.63}    &\textbf{74.71 $ \pm $ 15.98}    &65.67 $ \pm $ 02.15    &    &    &    &    \\
11 &98.64 $ \pm $ 00.70    &\textbf{99.81 $ \pm $ 00.20}    &    &    &\textbf{74.87 $ \pm $ 04.06}    &72.79 $ \pm $ 09.25    &93.71 $ \pm $ 06.01    &\textbf{96.90 $ \pm $ 00.47}    &    &    &    &    \\
12 &\textbf{94.24 $ \pm $ 02.19}    &89.95 $ \pm $ 01.93    &    &    &    &    &96.16 $ \pm $ 00.87    &\textbf{97.05 $ \pm $ 00.99}    &    &    &    &    \\
13 &\textbf{90.29 $ \pm $ 00.66}    &88.42 $ \pm $ 03.49    &    &    &    &    &\textbf{92.51 $ \pm $ 00.17}    &91.23 $ \pm $ 00.76    &    &    &    &    \\
14 &\textbf{99.73 $ \pm $ 00.38}    &99.73 $ \pm $ 00.38    &    &    &    &    &\textbf{100.0 $ \pm $ 00.00}    &99.46 $ \pm $ 00.76    &    &    &    &    \\
15 &\textbf{99.58 $ \pm $ 00.17}    &94.08 $ \pm $ 07.07    &    &    &    &    &86.82 $ \pm $ 08.99    &\textbf{95.07 $ \pm $ 02.46}    &    &    &    &    \\ \hline \hline
OA   &\textbf{89.80 $ \pm $ 00.53}    &89.39 $ \pm $ 00.78    &97.82 $ \pm $ 00.24    &\textbf{98.32 $ \pm $ 00.25}    &94.26 $ \pm $ 00.20    &\textbf{94.34 $ \pm $ 00.07}    &\textbf{89.15 $ \pm $ 00.96}    &88.97 $ \pm $ 00.36    &\textbf{90.49 $ \pm $ 00.20}    &90.40 $ \pm $ 00.17    &89.48 $ \pm $ 00.27    &\textbf{90.49 $ \pm $ 00.54}    \\ %\hline
AA  &\textbf{91.51 $ \pm $ 00.40}    &90.41 $ \pm $ 01.10    &93.73 $ \pm $ 01.52    &\textbf{95.98 $ \pm $ 01.64}    &80.81 $ \pm $ 01.02    &\textbf{81.48 $ \pm $ 00.70}    &90.56 $ \pm $ 00.93    &\textbf{90.65 $ \pm $ 00.29}    &\textbf{60.36 $ \pm $ 02.36}    &57.97 $ \pm $ 02.33    &59.13 $ \pm $ 02.80    &\textbf{64.70 $ \pm $ 00.44}    \\ %\hline
$\kappa (\times 100)$   &\textbf{88.93 $ \pm $ 00.59}    &88.48 $ \pm $ 00.85    &97.08 $ \pm $ 00.33    &\textbf{97.75 $ \pm $ 00.33}    &92.41 $ \pm $ 00.27    &\textbf{92.51 $ \pm $ 00.10}    &\textbf{88.22 $ \pm $ 01.04}    &88.02 $ \pm $ 00.39    &\textbf{86.26 $ \pm $ 00.23}    &86.07 $ \pm $ 00.28    &84.80 $ \pm $ 00.43    &\textbf{86.35 $ \pm $ 00.77}    \\ \hline \hline
\end{tabular}}
\label{tab:Ablation}
\end{table*}
%%%%%%%%%%%%%%%%%%%%%%%%%%%%%%%%%%%%%%%%%

%%%%%%%%%%%%%%%%%%%%%%%%%%%%%%%%%%%%
\begin{figure*}[!ht]
\centering
	\begin{subfigure}{0.24\textwidth}
		\includegraphics[width=0.99\textwidth]{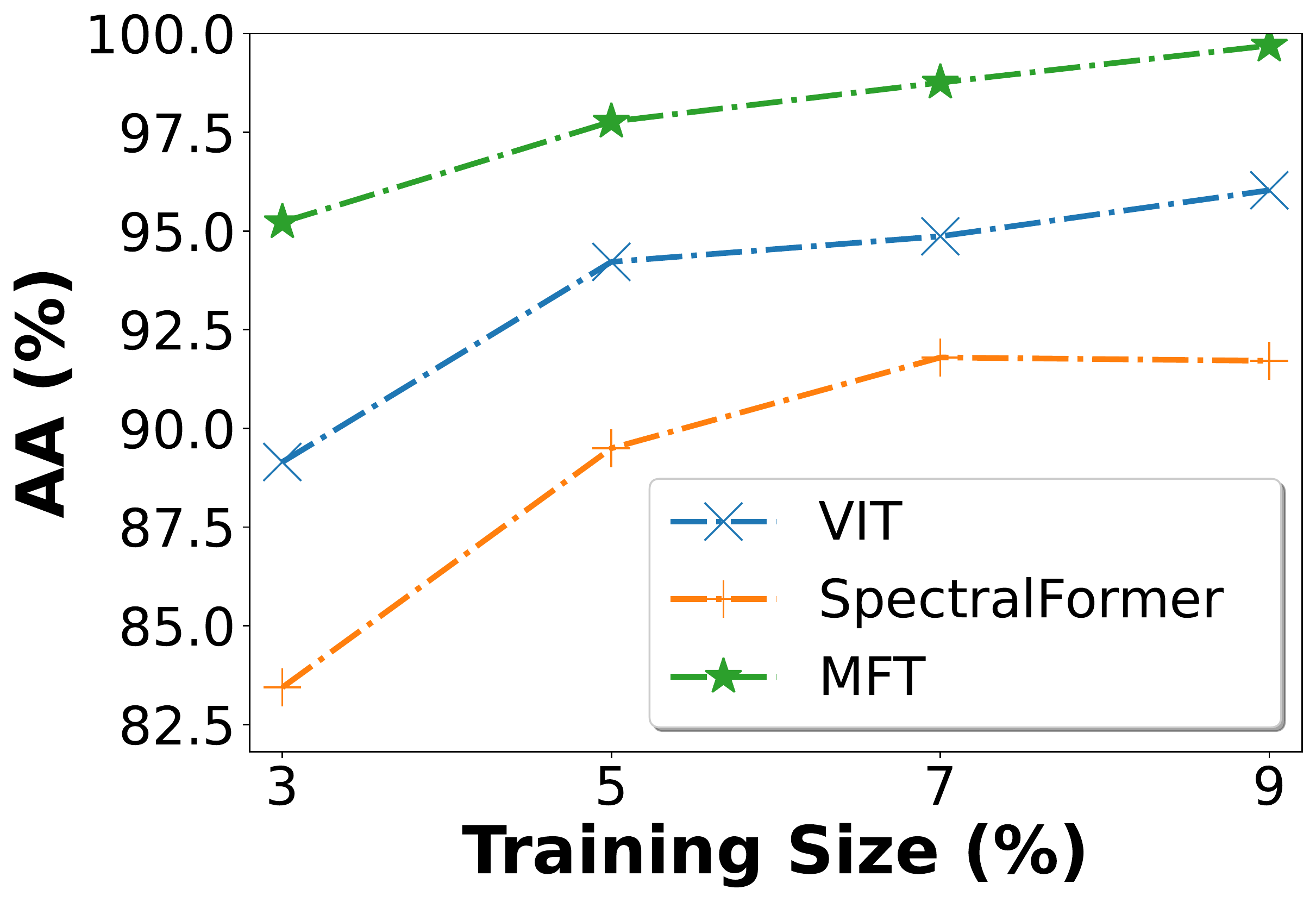}
		\centering
		\caption{UH (H+L)} 
		% \label{Fig1C}
	\end{subfigure}
	\begin{subfigure}{0.24\textwidth}
		\includegraphics[width=0.99\textwidth]{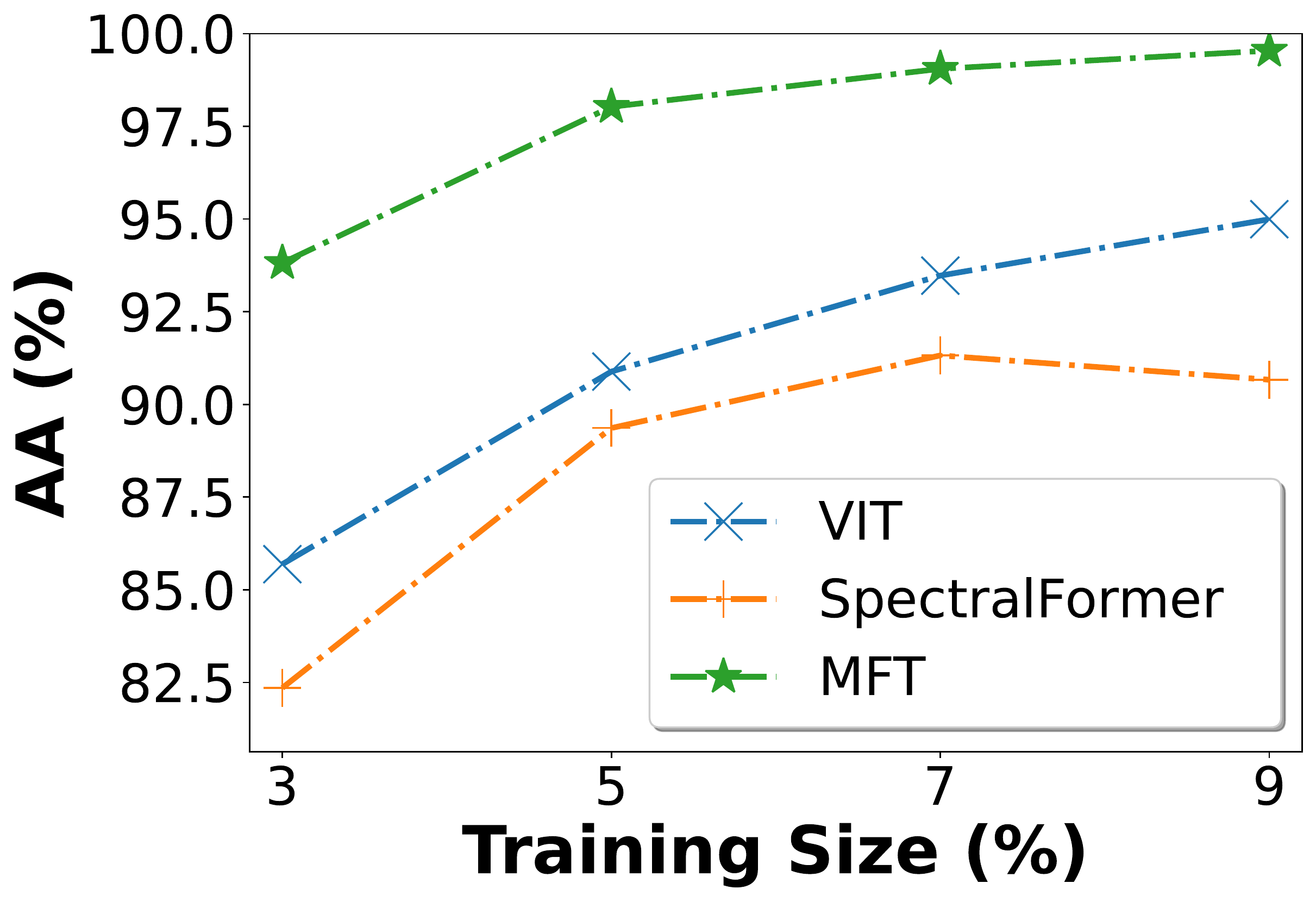}
		\centering
		\caption{UH (H+M)} 
		% \label{Fig1D}
	\end{subfigure}
	\begin{subfigure}{0.24\textwidth}
		\includegraphics[width=0.99\textwidth]{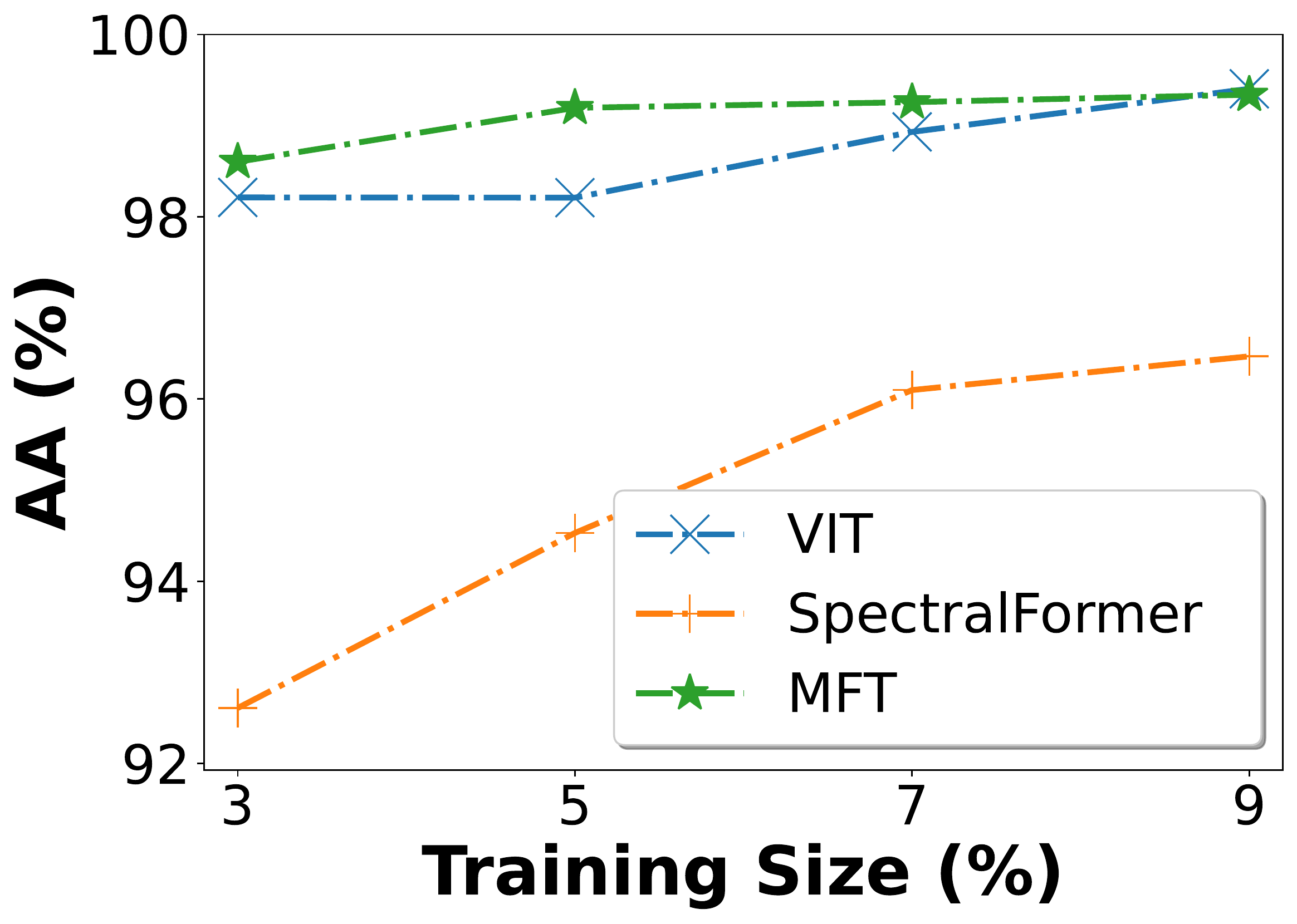}
		\centering
		\caption{UT}
		% \label{Fig1E}
	\end{subfigure}
	\begin{subfigure}{0.24\textwidth}
		\includegraphics[width=0.99\textwidth]{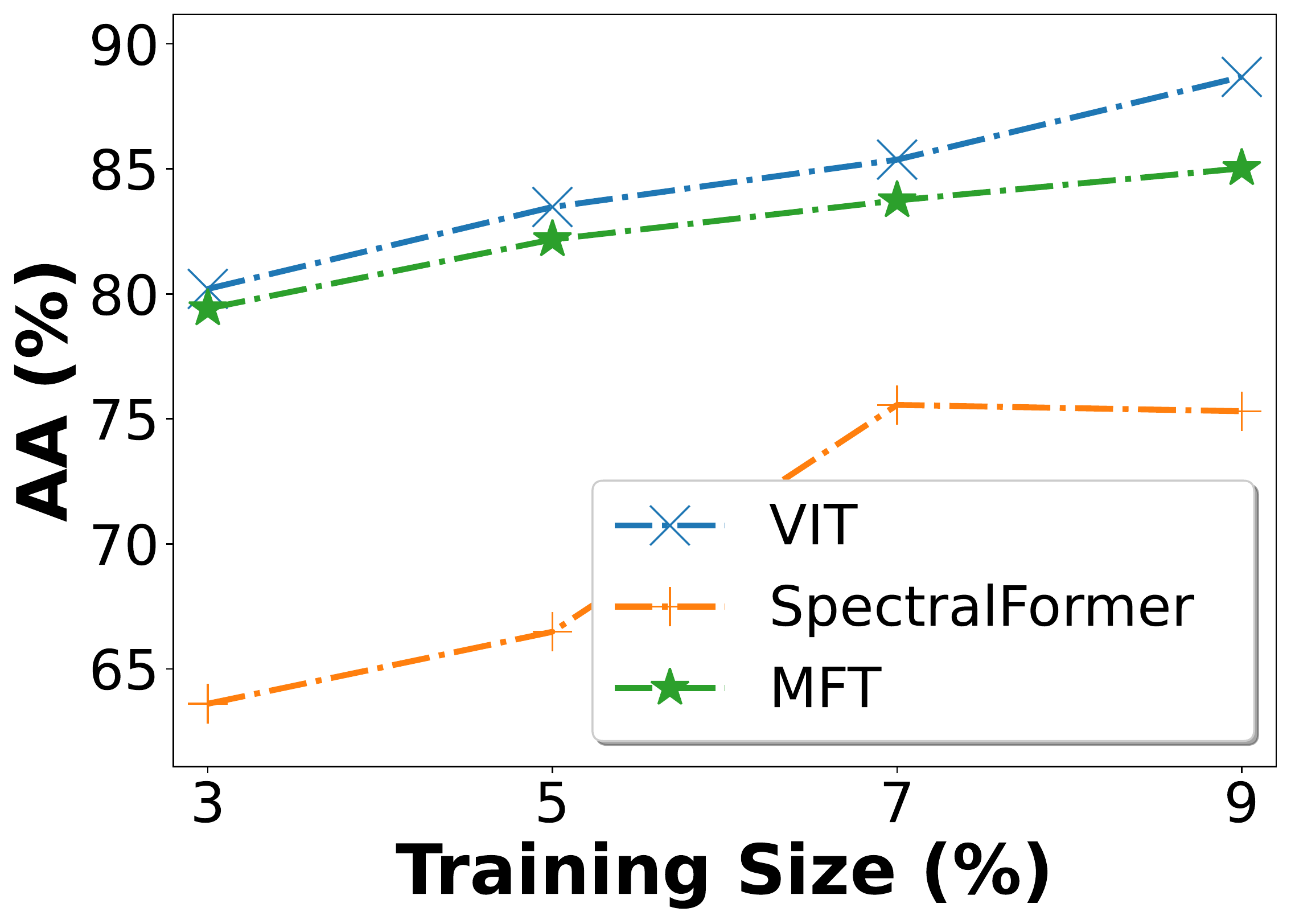}
		\centering
		\caption{MUUFL}
		% \label{Fig1E}
	\end{subfigure}	
	\begin{subfigure}{0.24\textwidth}
		\includegraphics[width=0.99\textwidth]{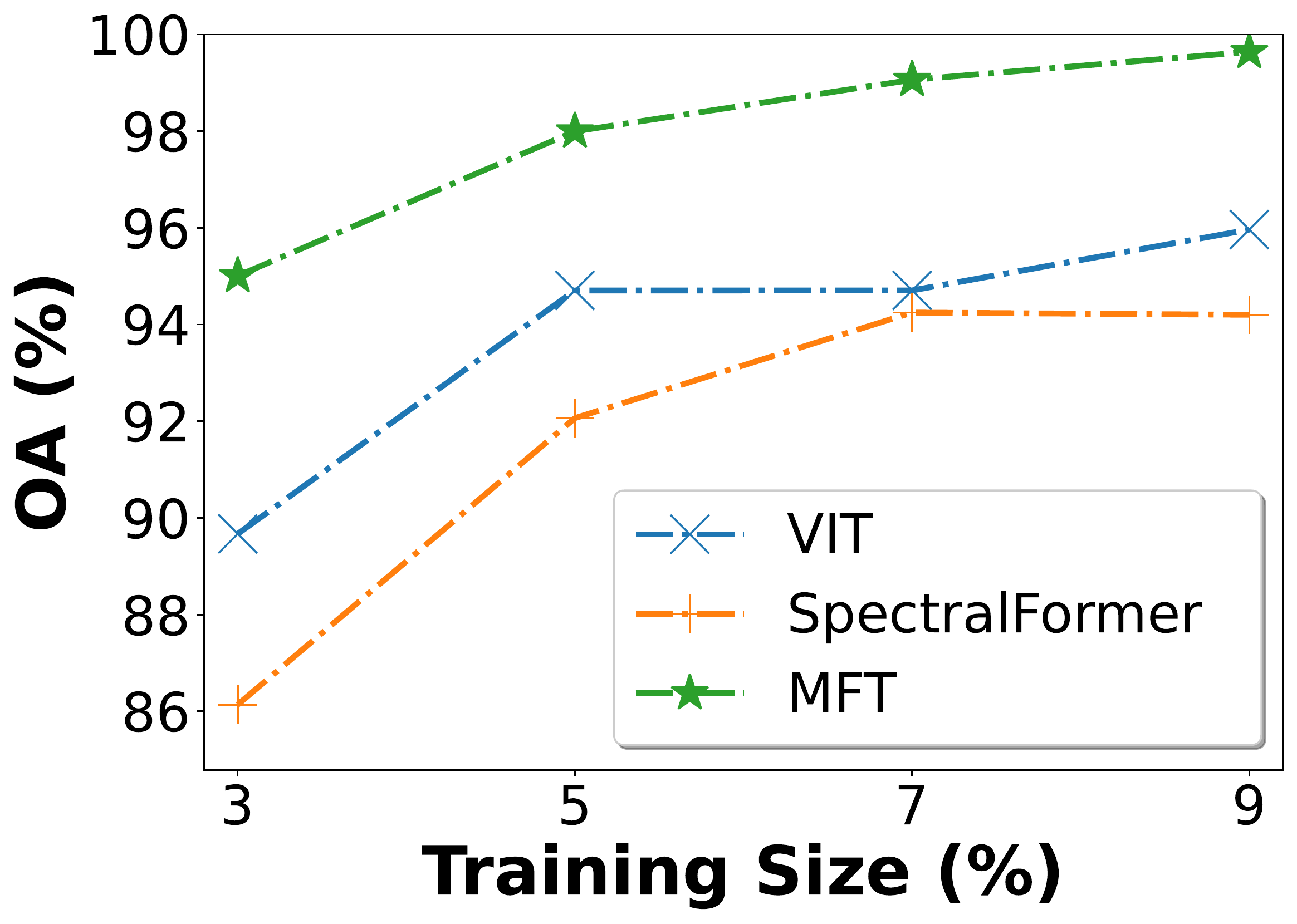}
		\centering
		\caption{UH (H+L)} 
		% \label{Fig1C}
	\end{subfigure}
	\begin{subfigure}{0.24\textwidth}
		\includegraphics[width=0.99\textwidth]{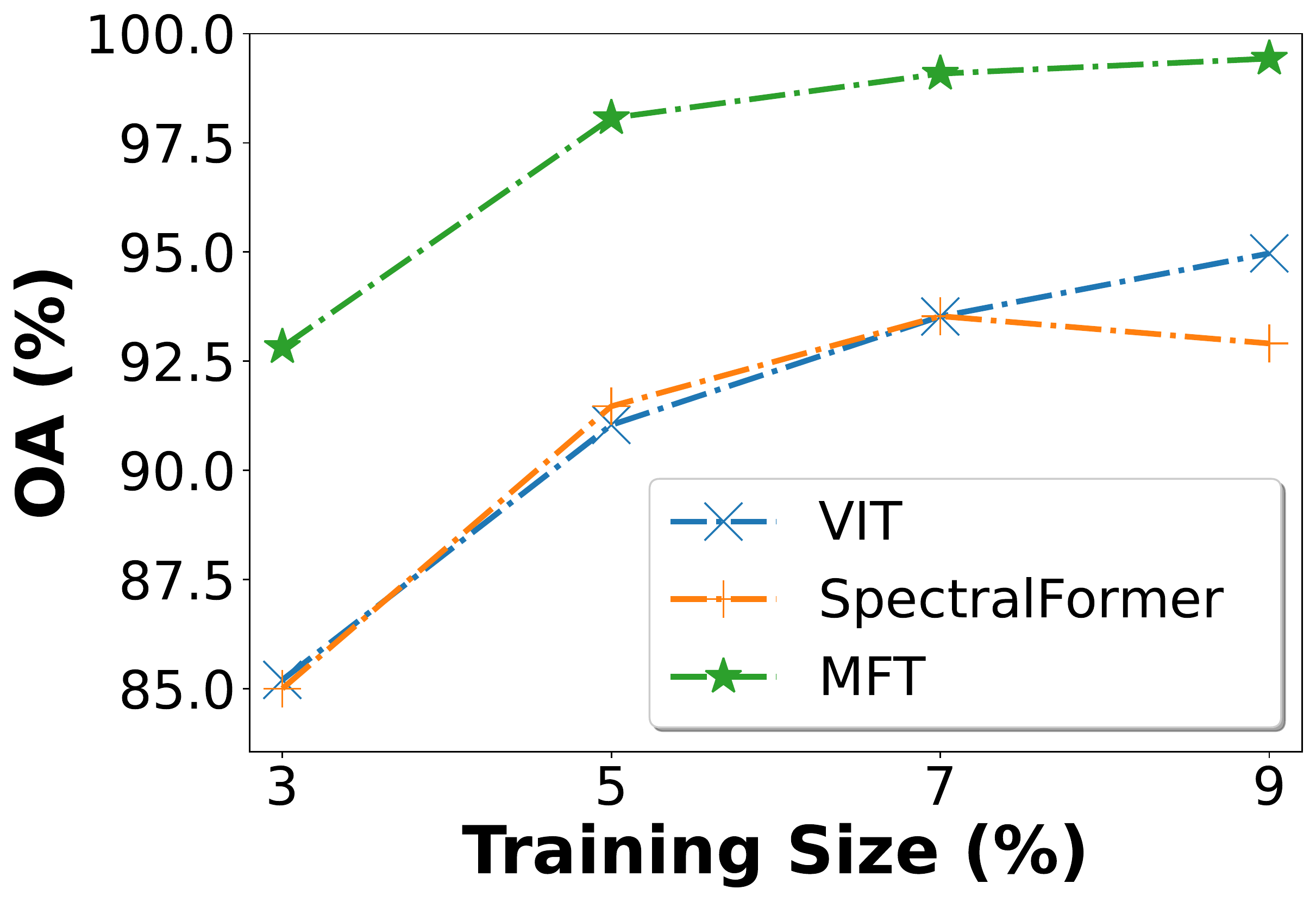}
		\centering
		\caption{UH (H+M)} 
		% \label{Fig1D}
	\end{subfigure}
	\begin{subfigure}{0.24\textwidth}
		\includegraphics[width=0.99\textwidth]{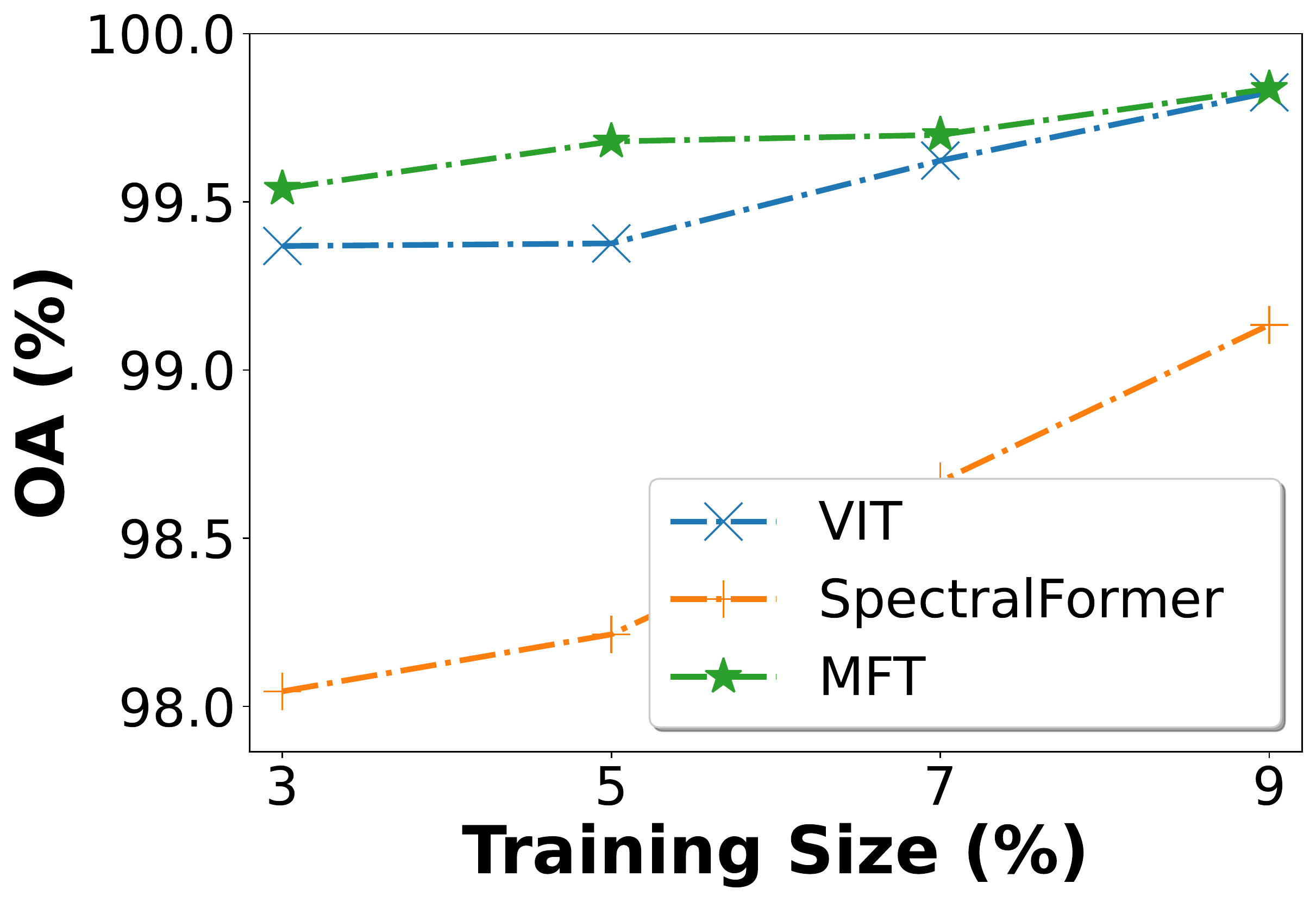}
		\centering
		\caption{UT}
		% \label{Fig1E}
	\end{subfigure}
	\begin{subfigure}{0.24\textwidth}
		\includegraphics[width=0.99\textwidth]{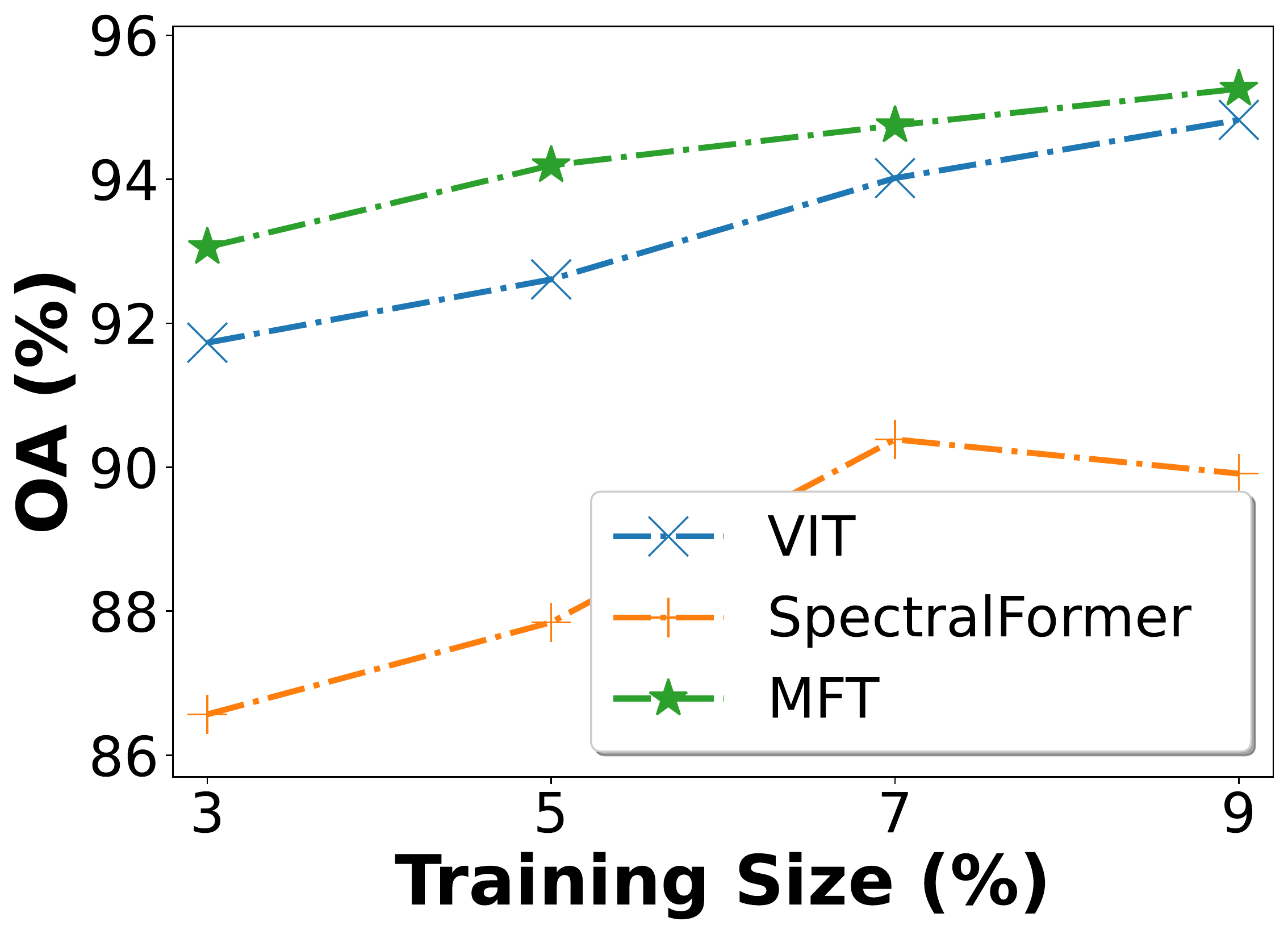}
		\centering
		\caption{MUUFL}
		% \label{Fig1E}
	\end{subfigure}
	\begin{subfigure}{0.24\textwidth}
		\includegraphics[width=0.99\textwidth]{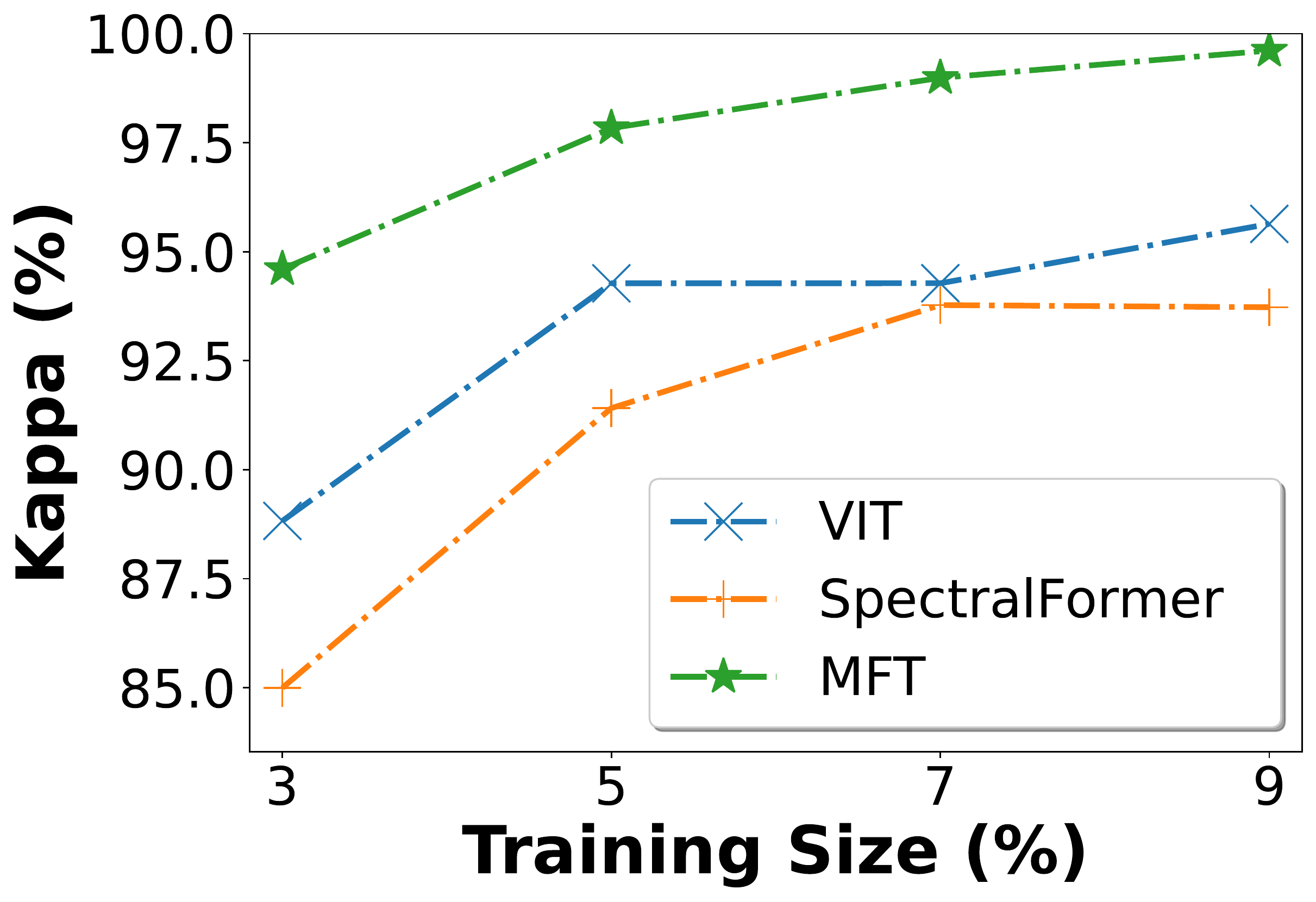}
		\centering
		\caption{UH (H+L)} 
		% \label{Fig1C}
	\end{subfigure}
	\begin{subfigure}{0.24\textwidth}
		\includegraphics[width=0.99\textwidth]{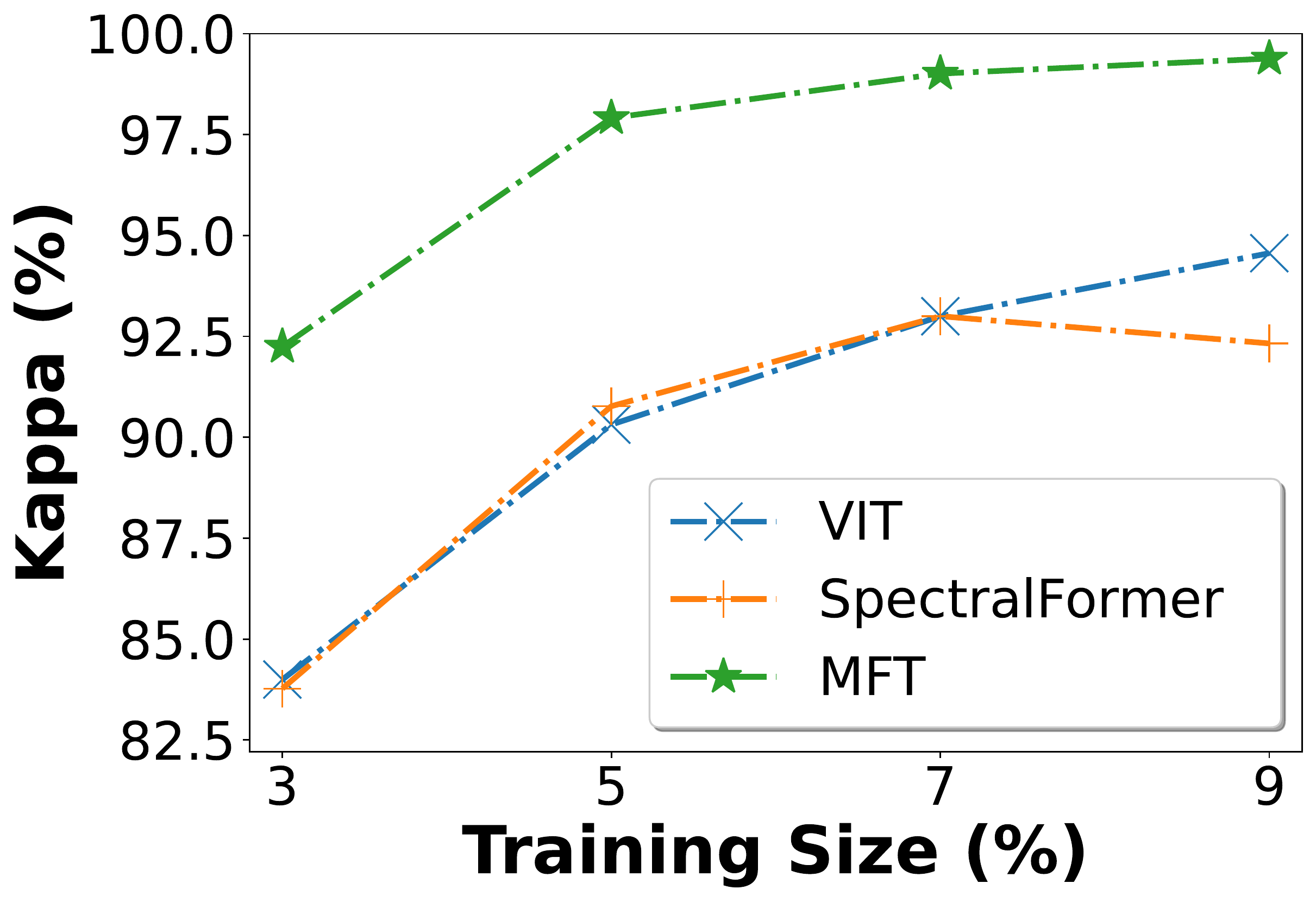}
		\centering
		\caption{UH (H+M)} 
		% \label{Fig1D}
	\end{subfigure}
	\begin{subfigure}{0.24\textwidth}
		\includegraphics[width=0.99\textwidth]{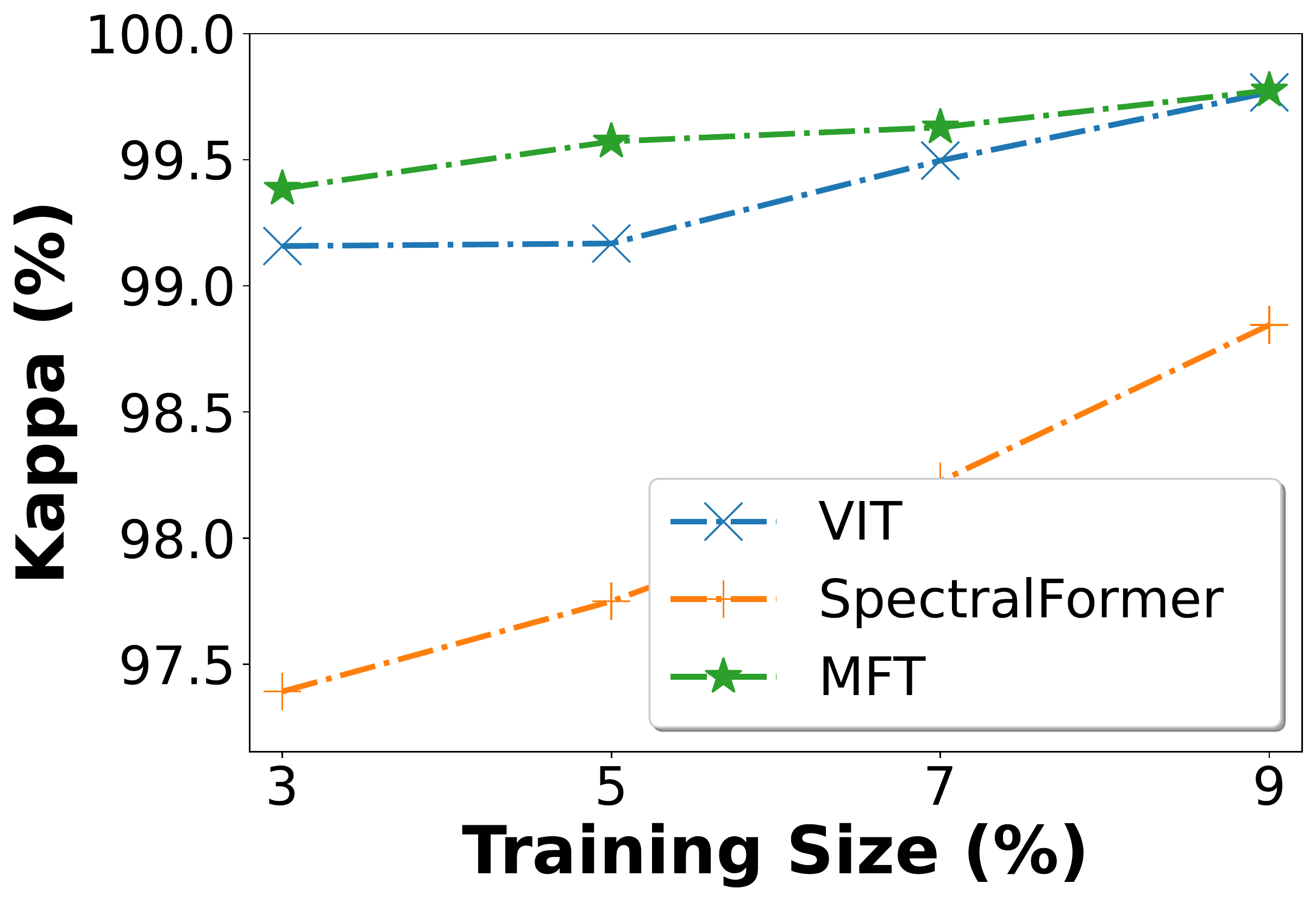}
		\centering
		\caption{UT}
		% \label{Fig1E}
	\end{subfigure}
	\begin{subfigure}{0.24\textwidth}
		\includegraphics[width=0.99\textwidth]{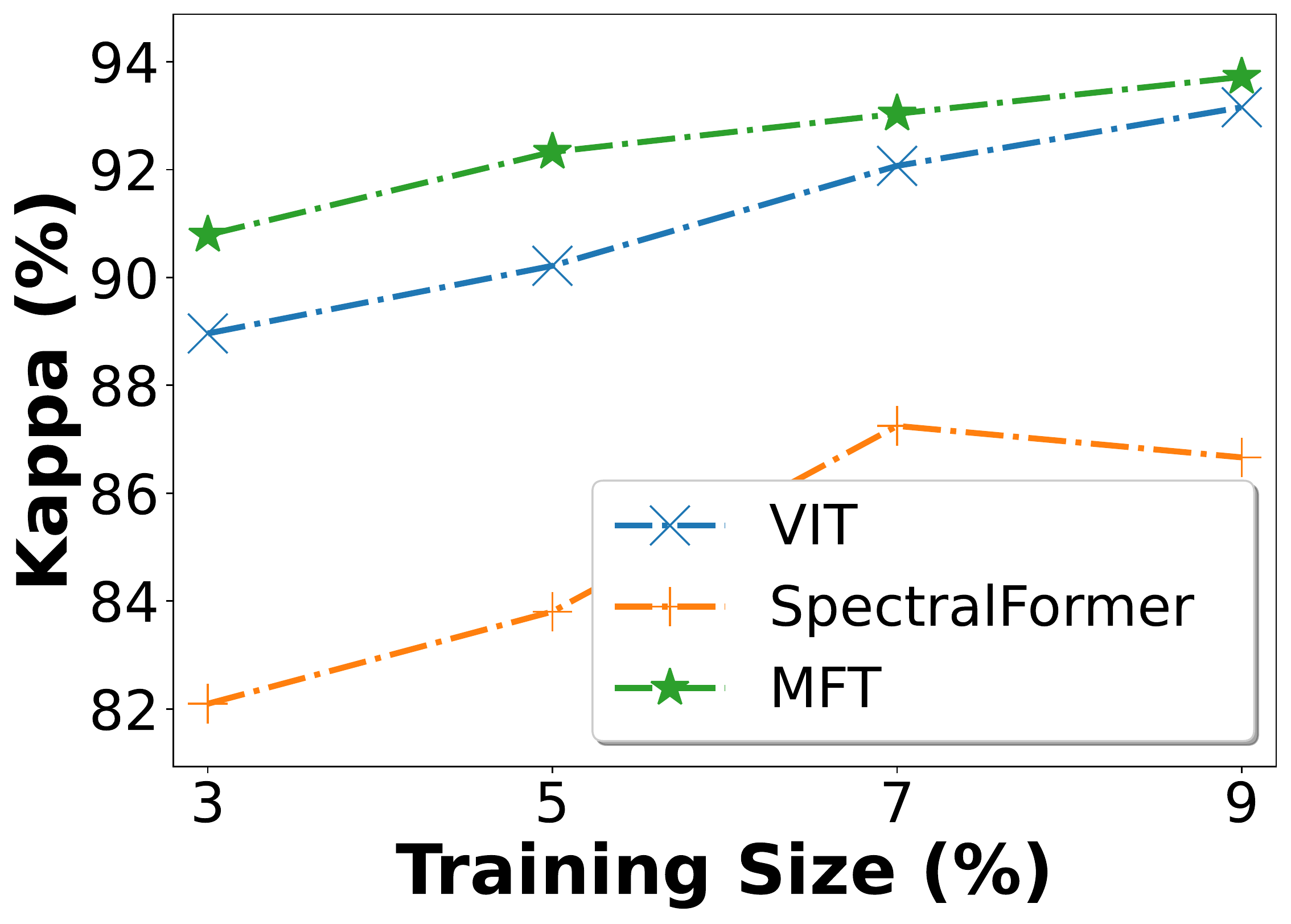}
		\centering
		\caption{MUUFL}
		% \label{Fig1E}
	\end{subfigure}
\caption{Average accuracy (AA), overall accuracy (OA) and kappa coefficient ($\kappa$) achieved by different methods with varying training sample sizes which are randomly taken from (a) UH (HSI+LiDAR) (b) UH (HSI+MS) (c) Trento (HSI+LiDAR) and (d) MUUFL (HSI+LiDAR) datasets.}
\label{fig:aa_oa_kappa}
\end{figure*}

%%%%%%%%%%%%%%%%%%%%%%%%%%%%%%%%%%%%%%%%%%%%%%%%%

\subsection{Visual Comparison}
We make a qualitative evaluation by visualizing the classification maps obtained by different methods. Figs. \ref{fig:comparativeUH(H+L)}, \ref{fig:comparativeUH(H+M)}, \ref{fig:comparativeUT}, \ref{fig:comparativeMUUFL}, \ref{fig:comparativeAugSAR}, and \ref{fig:comparativeAugDSM} depict the obtained classification maps for the hyperspectral datasets i.e., Houston, Trento, MUUFL and Augsburg along with their corresponding other modalities like LiDAR, MS image, SAR and DSM data, respectively. Although the conventional classifiers, i.e., KNN, RF, and SVM, produce rich classification maps, the maps still have salt and pepper noise in the boundary regions of the land-cover classes since they only rely on the spectral information from HS image data. The imperfections in the generated classification maps can be slightly reduced by considering the complementary information from other sources of modalities. However, the DL models successfully capture the nonlinear relationship between the input and output feature maps due to their learnable feature extraction capabilities. Hence, CNN1D, CNN2D, and CNN3D produce relatively smooth classification maps where land-use and land-cover class boundaries are well separable. ViT was found effective for HS classification due to its ability to extract high-level sequential abstract representations from HS data resulting. Therefore, classification maps are of better visual quality compared to classical networks. By enhancing the neighboring spectral information and conveying positional information across layers more effectively, the proposed MFT obtains highly desirable classification maps, especially in terms of texture and edge details compared to ViT, SpectralFormer. As observed from the generated classification maps, distinct class boundaries and remarkable improvements in the visual quality are obtained by considering the other sources of modalities during feature extraction and learning of all the networks. More noticeable improvements can be seen for the proposed MFT network due to the better exchange of information between HS patch and other modalities. The learning of complementary information helps find strong nonlinear sequential relations from the HS image data. Hence, the generated classification maps contain more realistic and finer details inside the land-use and land-cover region. %On the other hand, it is true that a fraction of noise still exists in the maps but \texttt{modalFormer} reduces it to a great extent.

%%%%%%%%%%%%%%%%%%%%%%%%%%%%%%%%%%%%
\begin{figure*}[!htbp]
\centering
	\begin{subfigure}{0.24\textwidth}
		\includegraphics[width=0.99\textwidth]{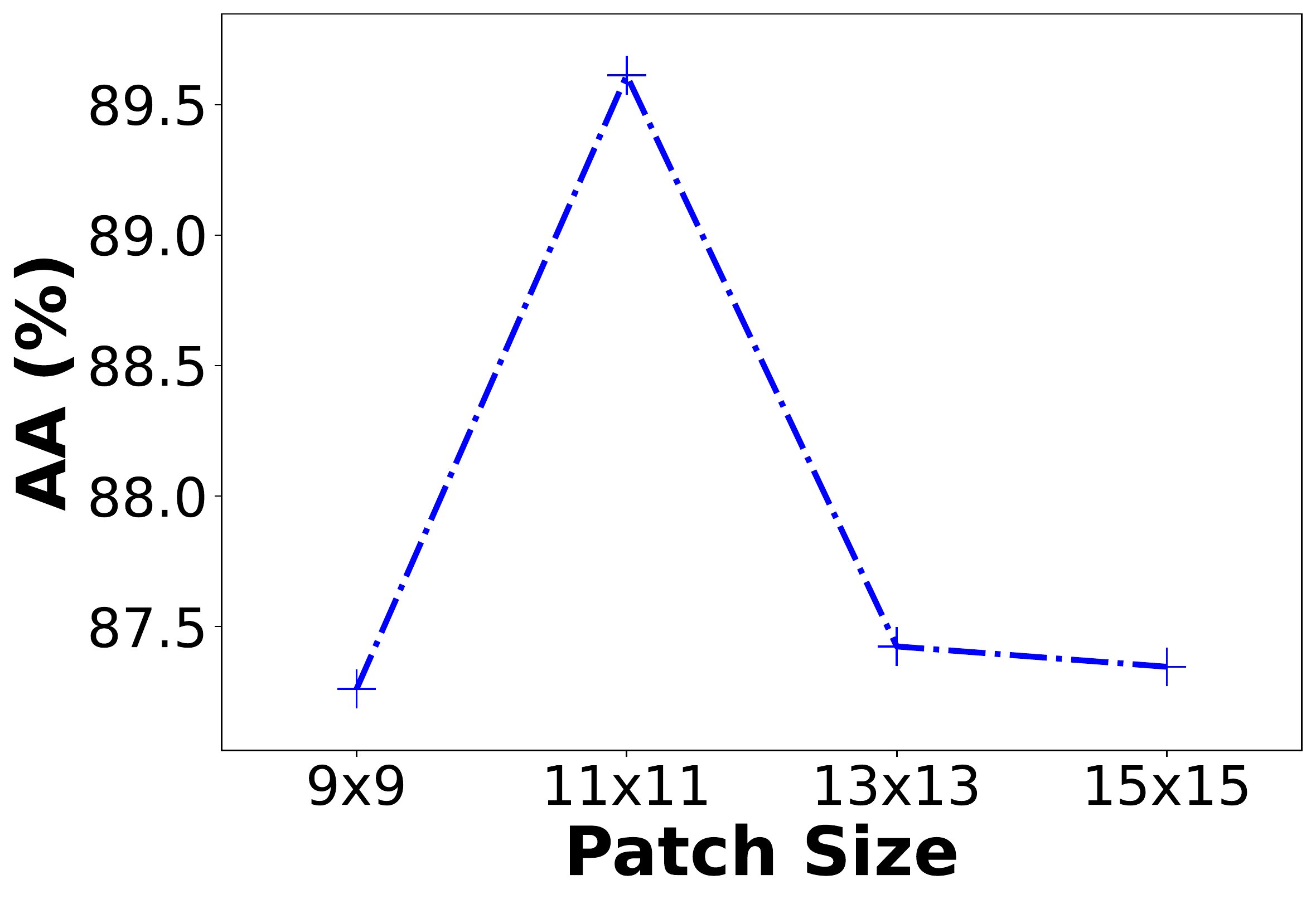}
		\centering
		\caption{UH (H+L)} 
		% \label{Fig1C}
	\end{subfigure}
	\begin{subfigure}{0.24\textwidth}
		\includegraphics[width=0.99\textwidth]{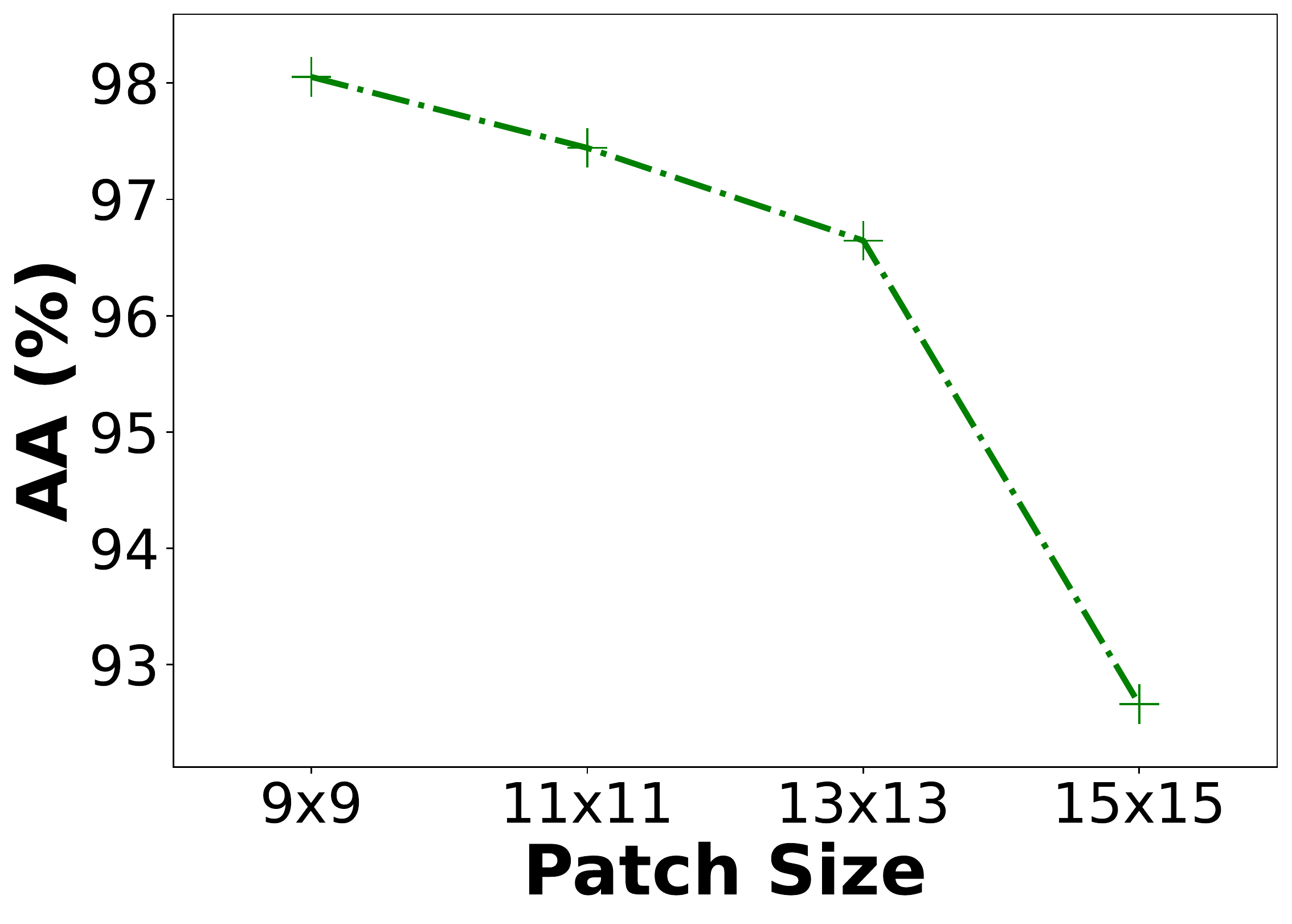}
		\centering
		\caption{UT (H+L)} 
		% \label{Fig1D}
	\end{subfigure}
	\begin{subfigure}{0.24\textwidth}
		\includegraphics[width=0.99\textwidth]{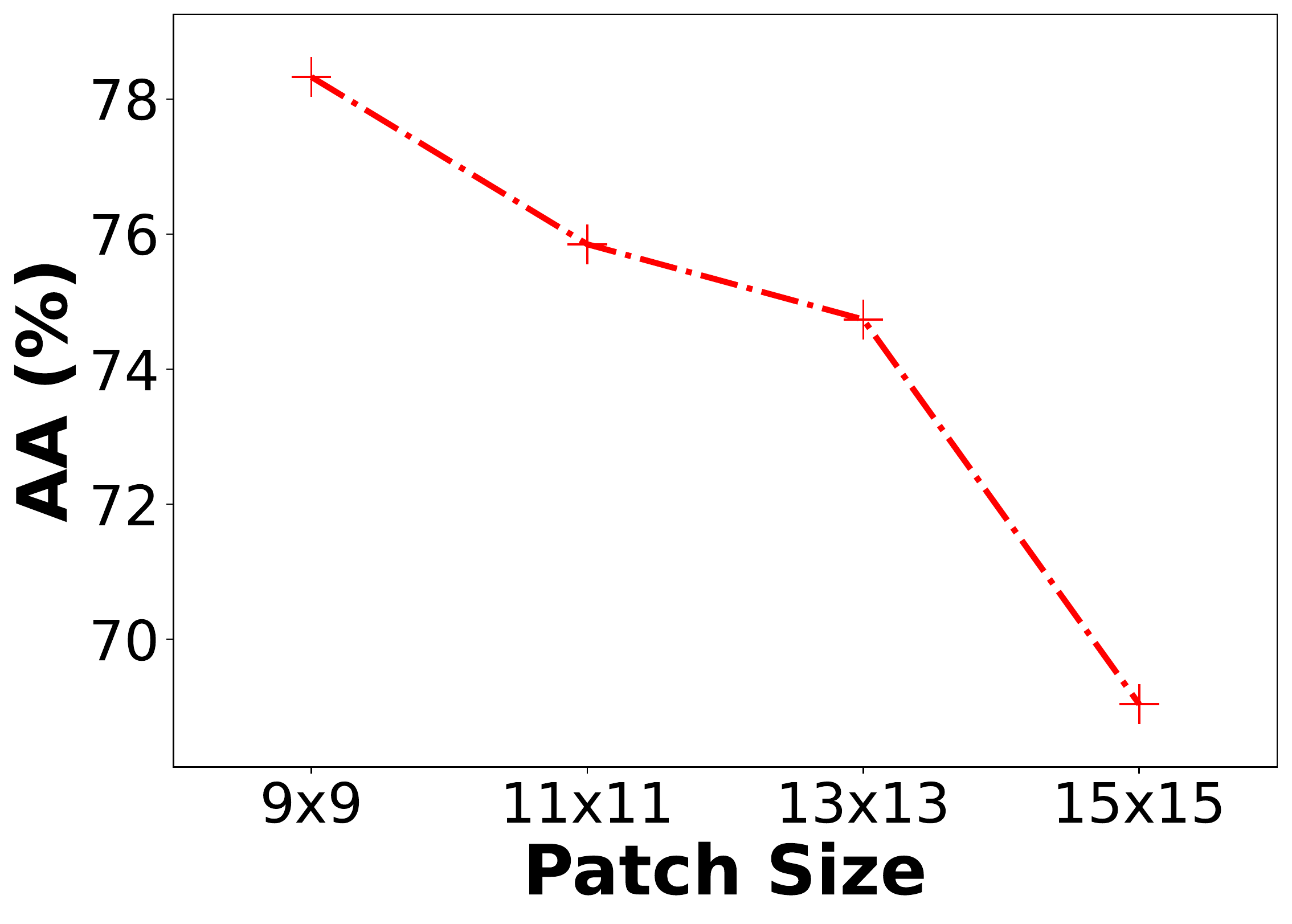}
		\centering
		\caption{MUUFL (H+L)}
		% \label{Fig1E}
	\end{subfigure}
	\begin{subfigure}{0.24\textwidth}
		\includegraphics[width=0.99\textwidth]{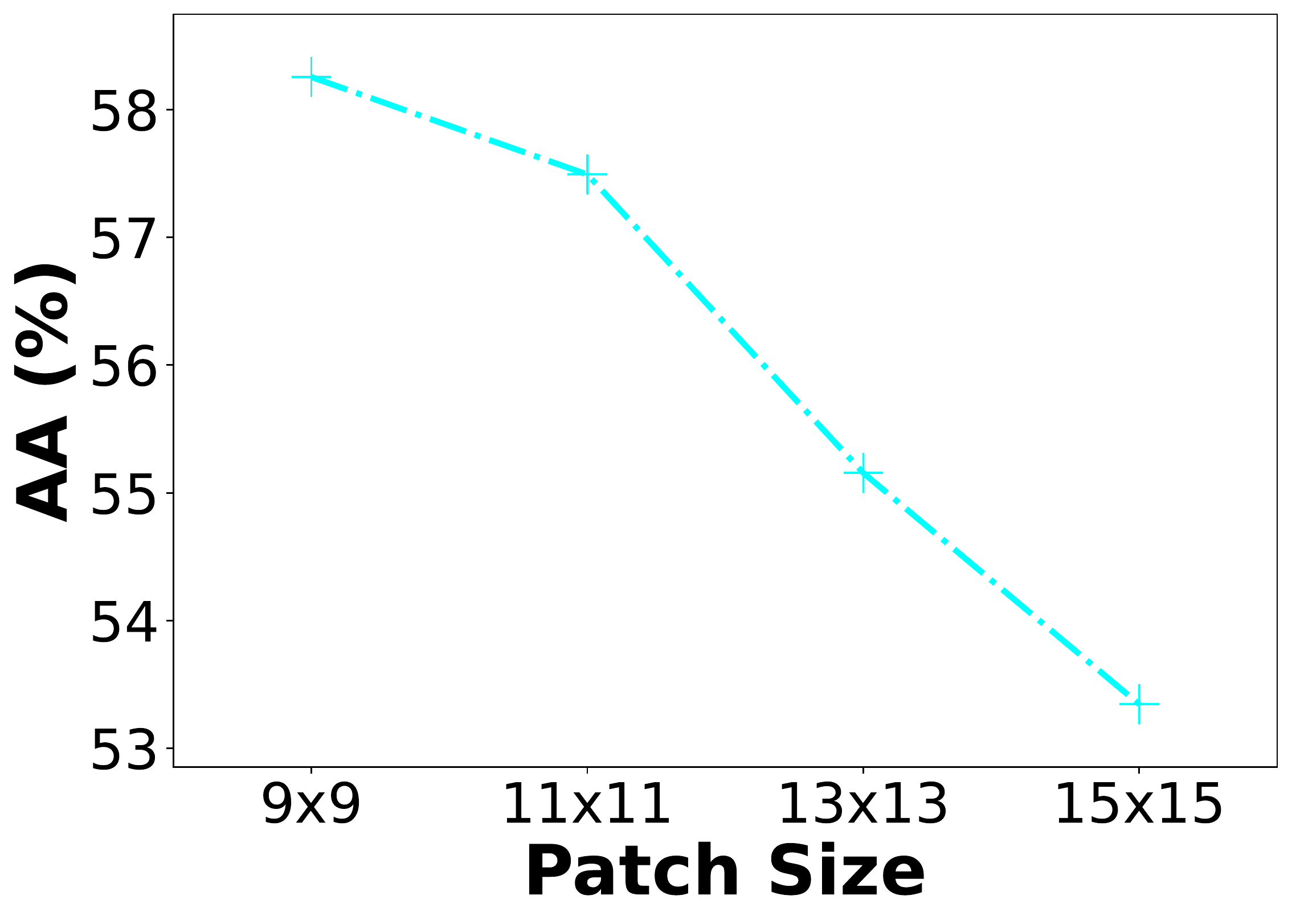}
		\centering
		\caption{Augsburg (H+S)}
		% \label{Fig1E}
	\end{subfigure}	
	\begin{subfigure}{0.24\textwidth}
		\includegraphics[width=0.99\textwidth]{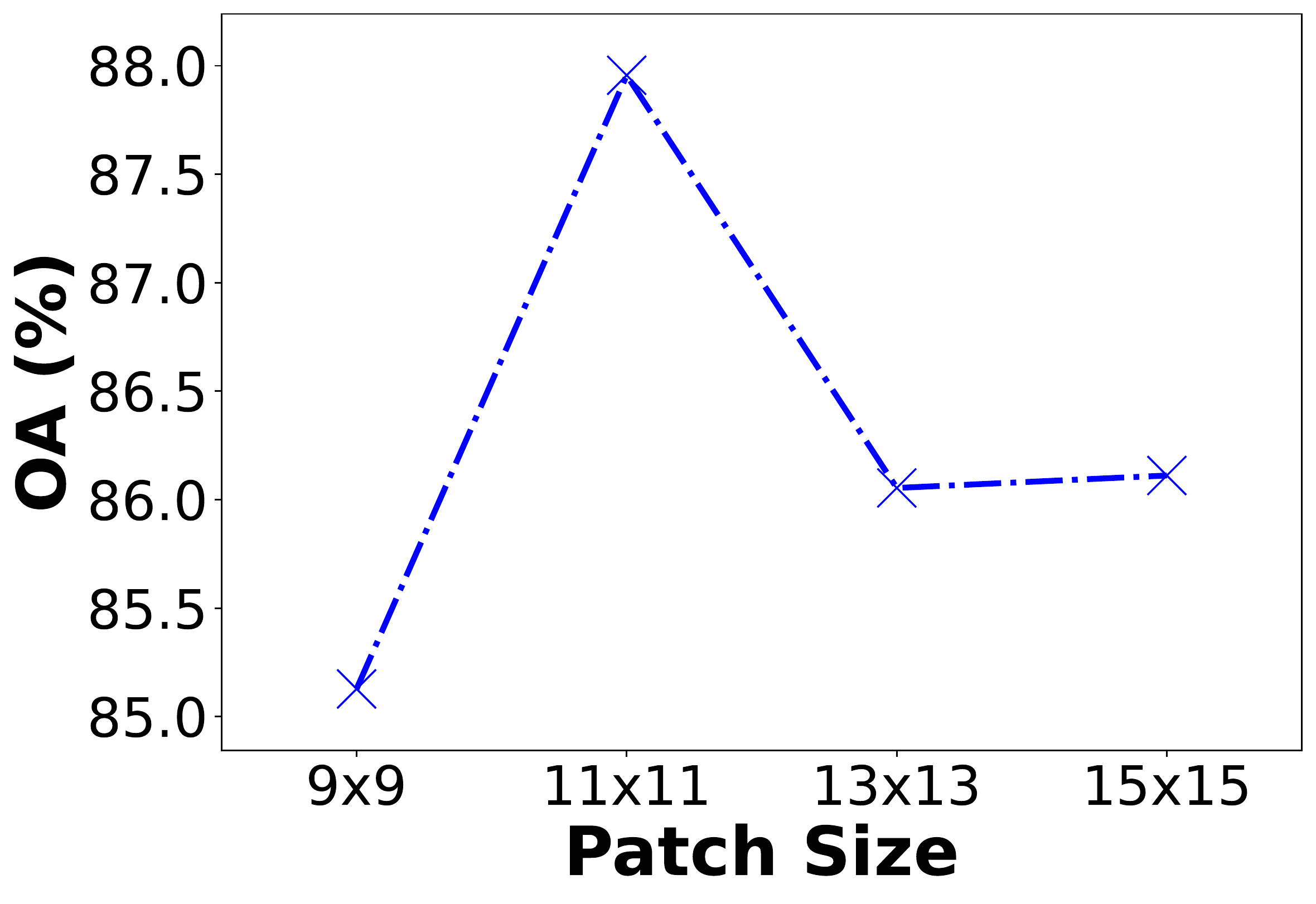}
		\centering
		\caption{UH (H+L)} 
		% \label{Fig1C}
	\end{subfigure}
	\begin{subfigure}{0.24\textwidth}
		\includegraphics[width=0.99\textwidth]{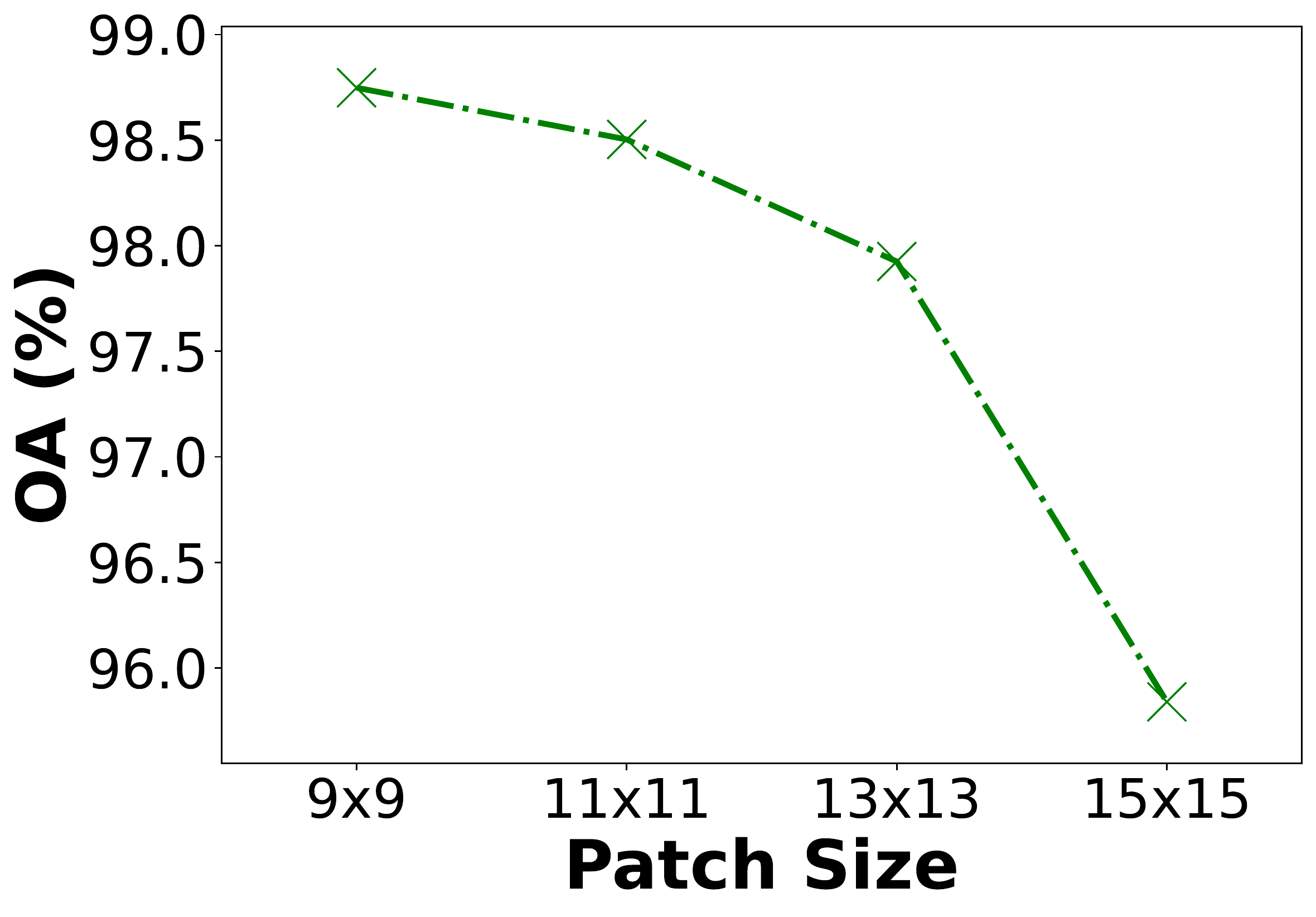}
		\centering
		\caption{UT (H+L)} 
		% \label{Fig1D}
	\end{subfigure}
	\begin{subfigure}{0.24\textwidth}
		\includegraphics[width=0.99\textwidth]{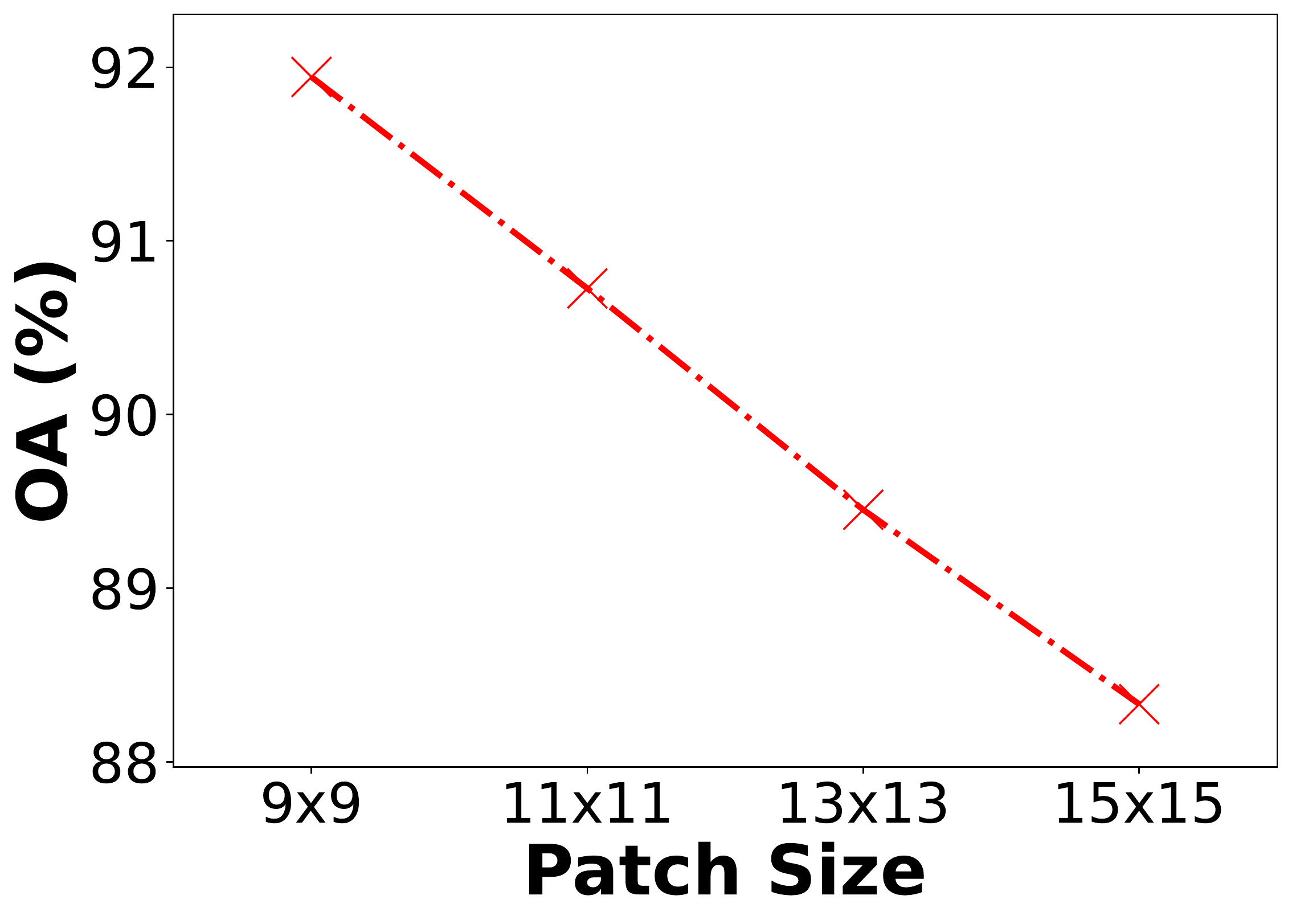}
		\centering
		\caption{MUUFL (H+L)}
		% \label{Fig1E}
	\end{subfigure}
	\begin{subfigure}{0.24\textwidth}
		\includegraphics[width=0.99\textwidth]{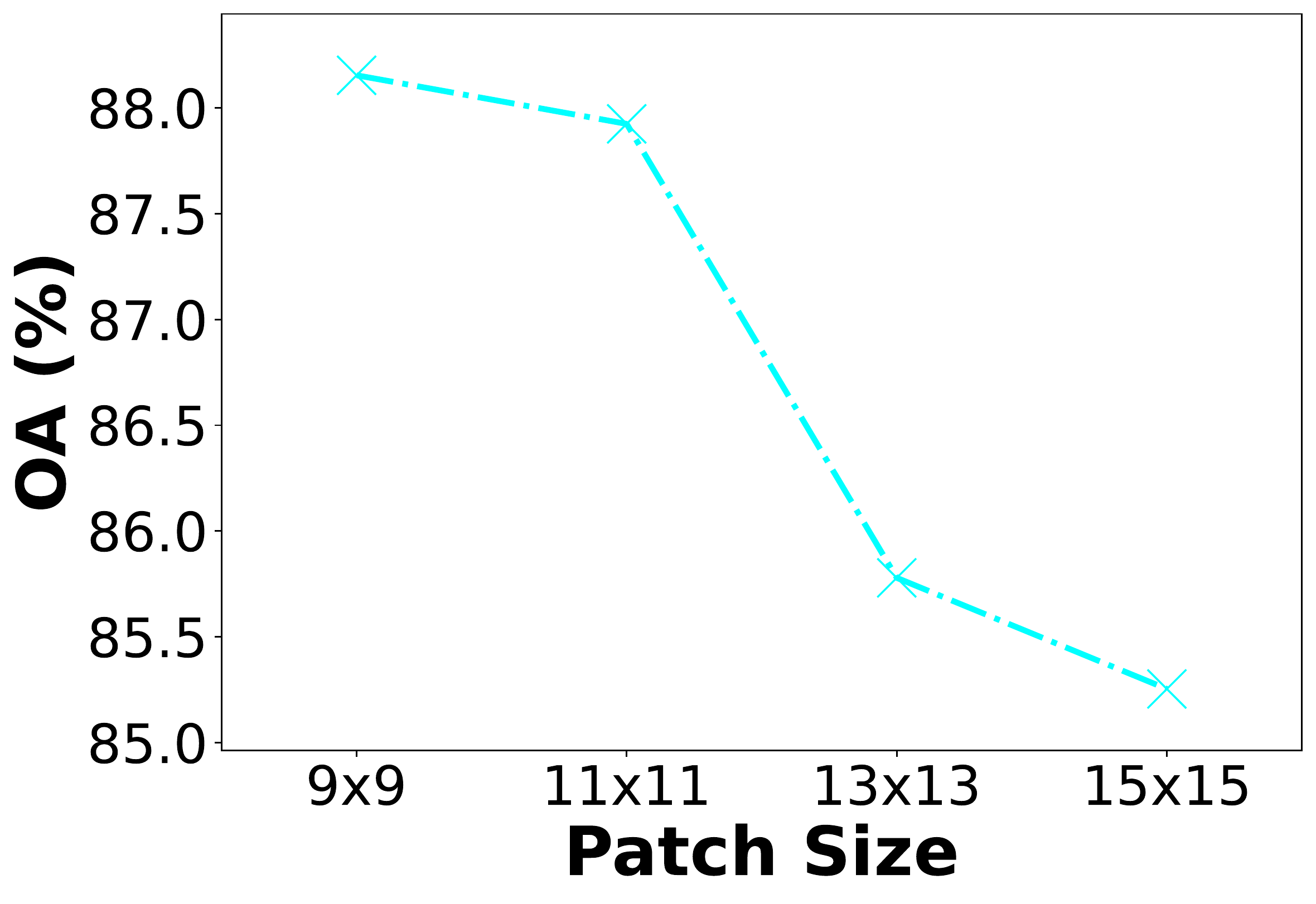}
		\centering
		\caption{Augsburg (H+S)}
		% \label{Fig1E}
	\end{subfigure}
	\begin{subfigure}{0.24\textwidth}
		\includegraphics[width=0.99\textwidth]{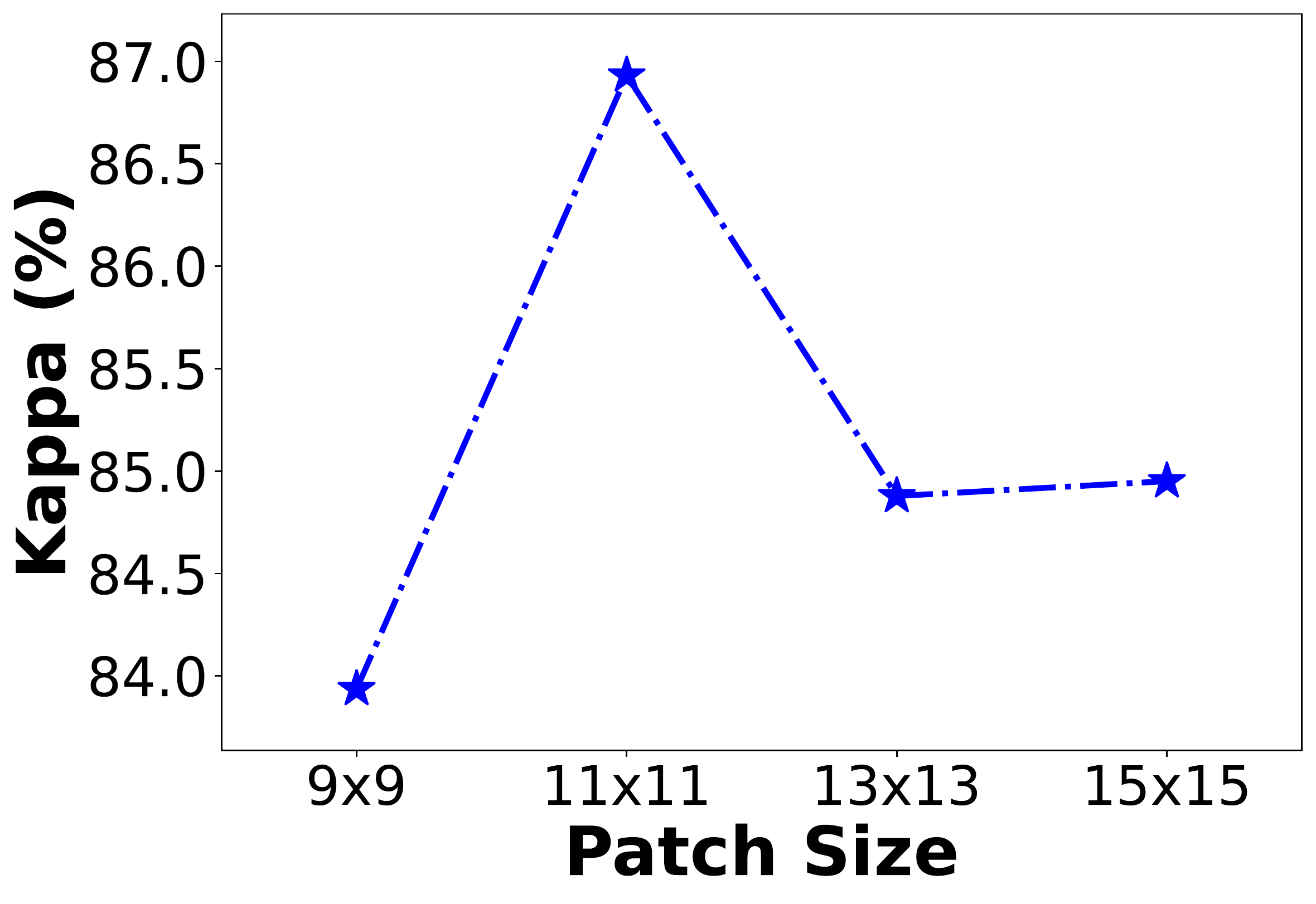}
		\centering
		\caption{UH (H+L)} 
		% \label{Fig1C}
	\end{subfigure}
	\begin{subfigure}{0.24\textwidth}
		\includegraphics[width=0.99\textwidth]{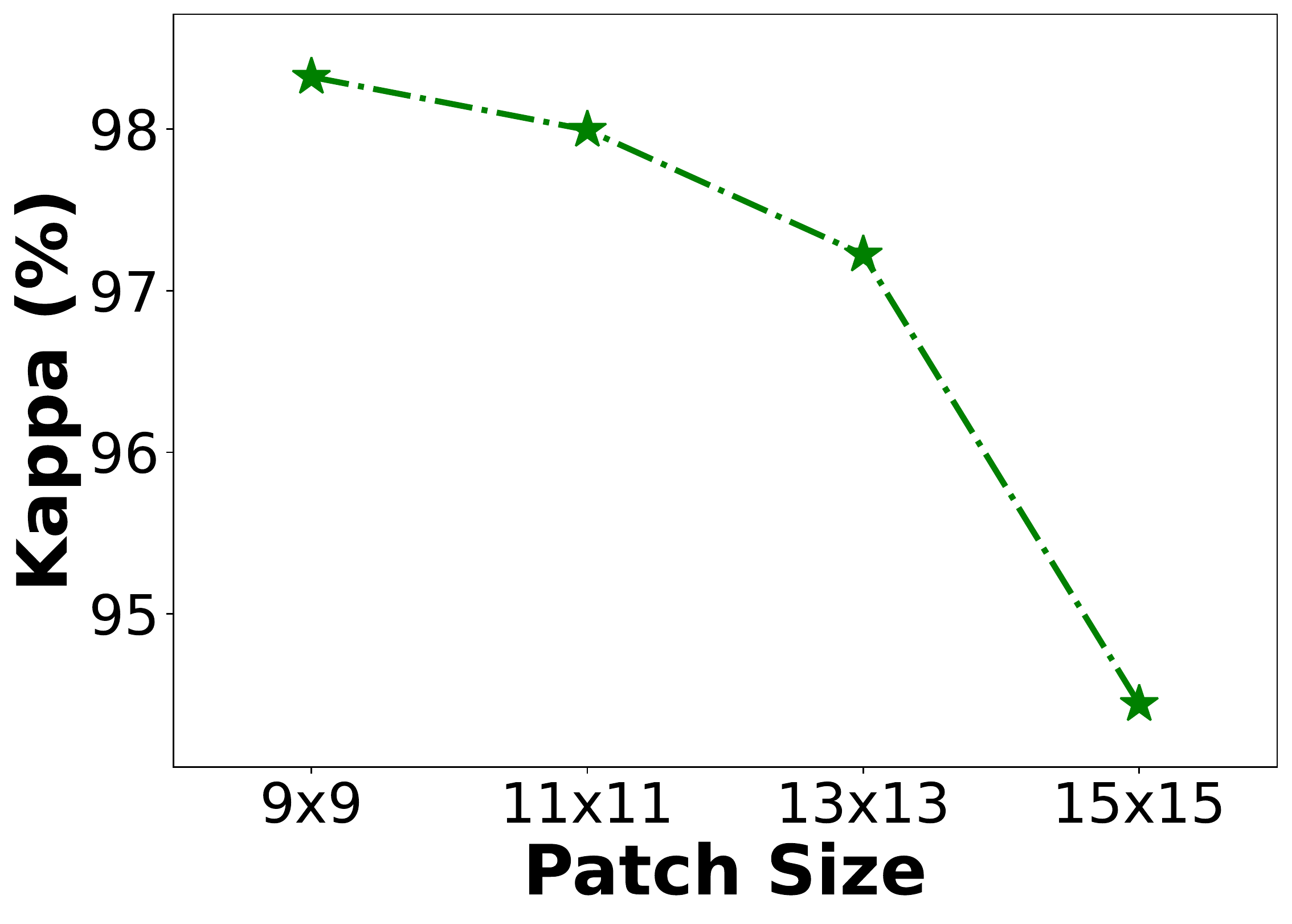}
		\centering
		\caption{UT (H+L)} 
		% \label{Fig1D}
	\end{subfigure}
	\begin{subfigure}{0.24\textwidth}
		\includegraphics[width=0.99\textwidth]{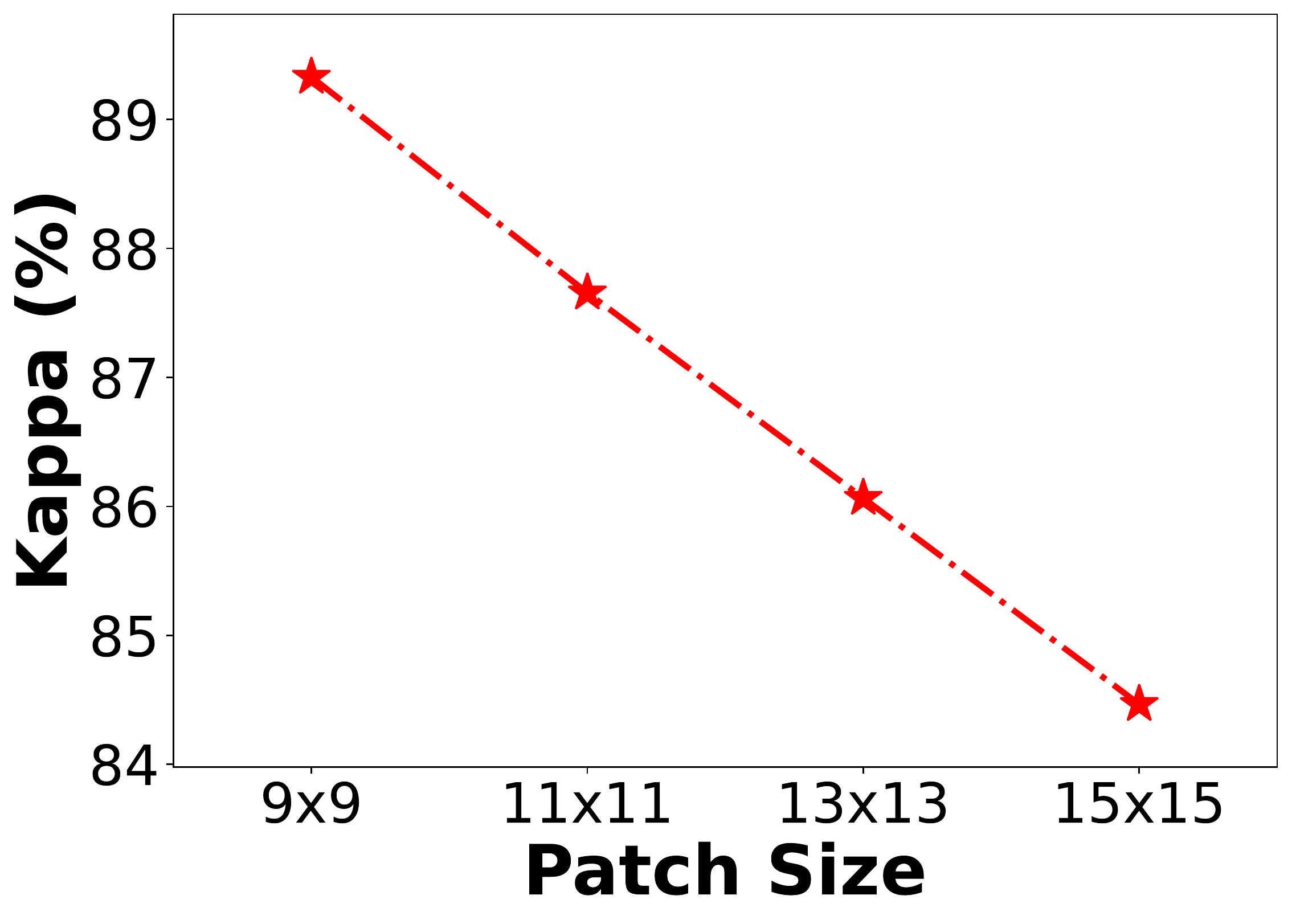}
		\centering
		\caption{MUUFL (H+L)}
		% \label{Fig1E}
	\end{subfigure}
	\begin{subfigure}{0.24\textwidth}
		\includegraphics[width=0.99\textwidth]{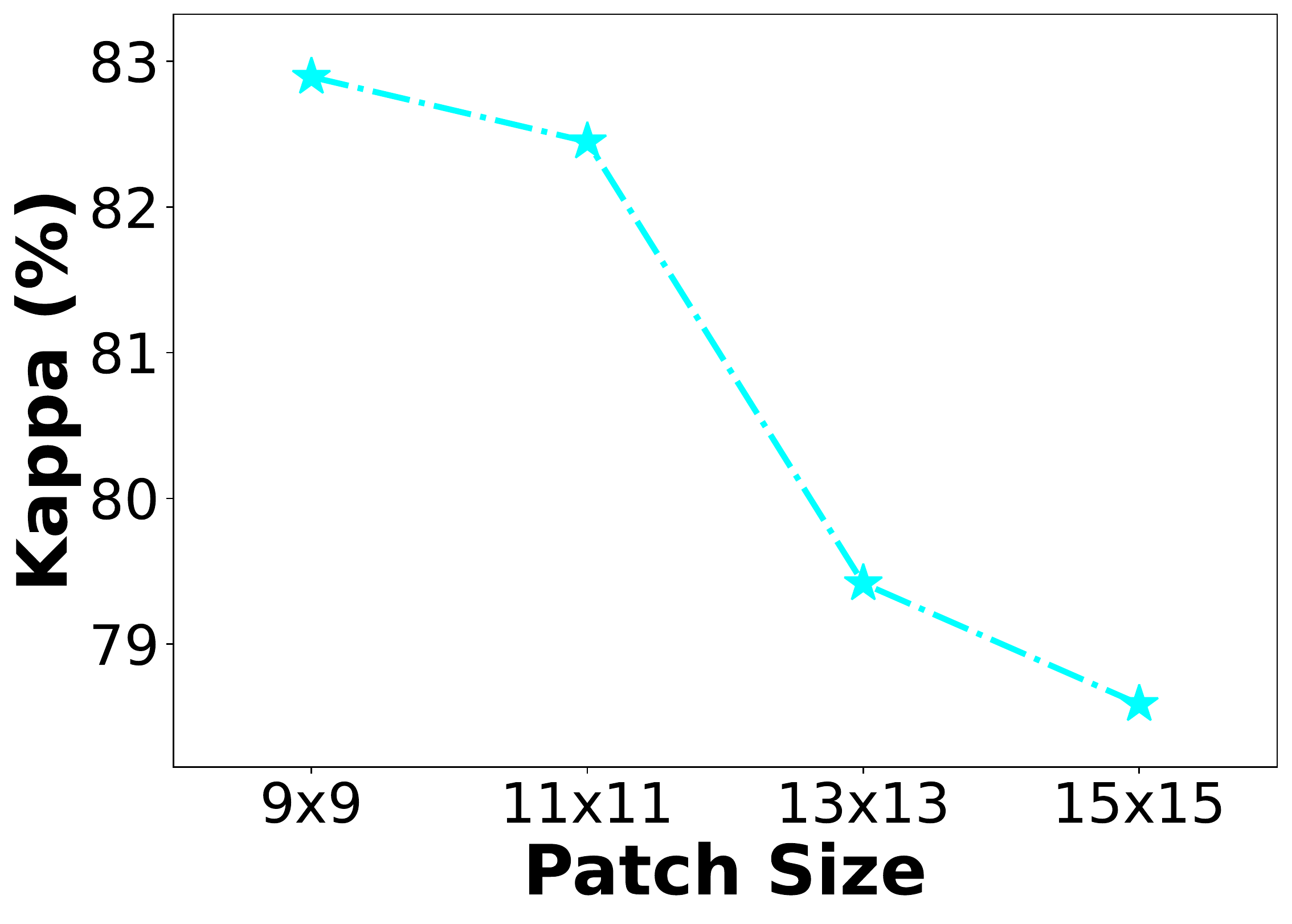}
		\centering
		\caption{Augsburg (H+S)}
		% \label{Fig1E}
	\end{subfigure}
\caption{\textcolor{black}{Average accuracy (AA), overall accuracy (OA) and kappa coefficient ($\kappa$) achieved by the MFT network with varying patch sizes which are taken from (a) UH (HSI+LiDAR) (b) Trento (HSI+LiDAR) (c) MUUFL (HSI+LiDAR) and (d) Augsburg (HSI+SAR) datasets.}}
\label{fig:aa_oa_kappa_patch}
\end{figure*}
%%%%%%%%%%%%%%%%%%%%%%%%%%%%%%%%%%%%%%%%%%%%%%%%%

\subsection{Performance Over Varying Training Ratio}

Here, we evaluate the performance of the proposed model with respect to the number of labeled training samples. In the case of inadequate or unreliable training samples, under-fitting (also called the Hughes Phenomena) or over-fitting (non-reliable) issues can occur. Therefore, selecting the appropriate number of training samples is another crucial factor for classification performance. The purpose of this section is to analyze the experimental results on different percentages of training samples where 3\%, 5\%, 7\%, and 9\% training samples are randomly selected to train the models (both the one presented in this paper and the comparative methods), and the remaining samples are used for testing. Other parameters remain the same, as discussed in the previous sections.

Figs. \ref{fig:aa_oa_kappa}(a)-(d), \ref{fig:aa_oa_kappa}(e)-(h), and \ref{fig:aa_oa_kappa}(i)-(l) compare the detailed classification performances of the transformer models using AA, OA, and $\kappa$ metrics respectively, on four datasets, i.e., Houston (HSI + LiDAR), Trento (HSI + LiDAR), MUUFL (HSI + LiDAR) and Houston (HSI + MS), with different percentages ($3$\%, $5$\%, $7$\% and $9$\%) of randomly selected training samples. The y-axes of the graphs show the metrics (OA, AA or $\kappa$) while the x-axes show the training sample size in percent. Each model is denoted with a different marker and color which are consistent throughout Fig. \ref{fig:aa_oa_kappa}. The proposed MFT is shown in green. 

The graphs illustrate that the experimental results obtained with $3$\% training samples are inferior to those obtained with $9$\% training samples for all four datasets. In Houston, (HSI + LiDAR) and Houston (HSI + MS) the proposed MFT performs approximately $4$\% and $6$\% better than the second best model i.e. ViT, respectively, in all $3$ accuracies for all percentages of training samples. In other datasets, though the proposed MFT performs better than ViT, the improvement is not much for larger training sample sizes. It is quite evident that even with the decrease in training samples, the performance of the proposed MFT is significantly better than the other transformer networks.
%%%%%%%%%%%%%%%%%%%%%%%%%%%%%%%%%%%%
\begin{figure*}[!t]
\centering
	\begin{subfigure}{0.30\textwidth}
		\includegraphics[width=0.9\textwidth]{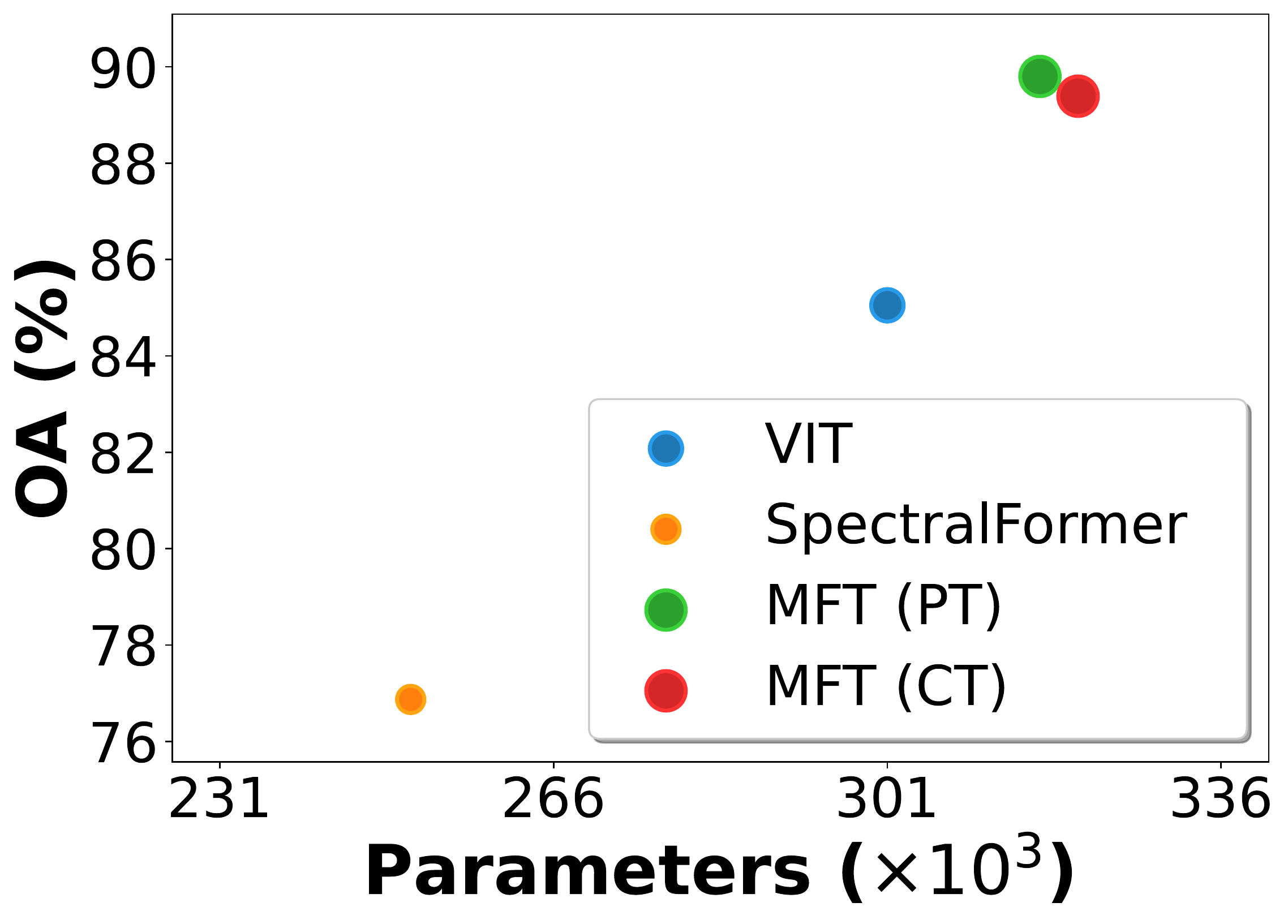}
		\centering
		\caption{UH (H+L)} 
		% \label{Fig1C}
	\end{subfigure}
	\begin{subfigure}{0.30\textwidth}
		\includegraphics[width=0.99\textwidth]{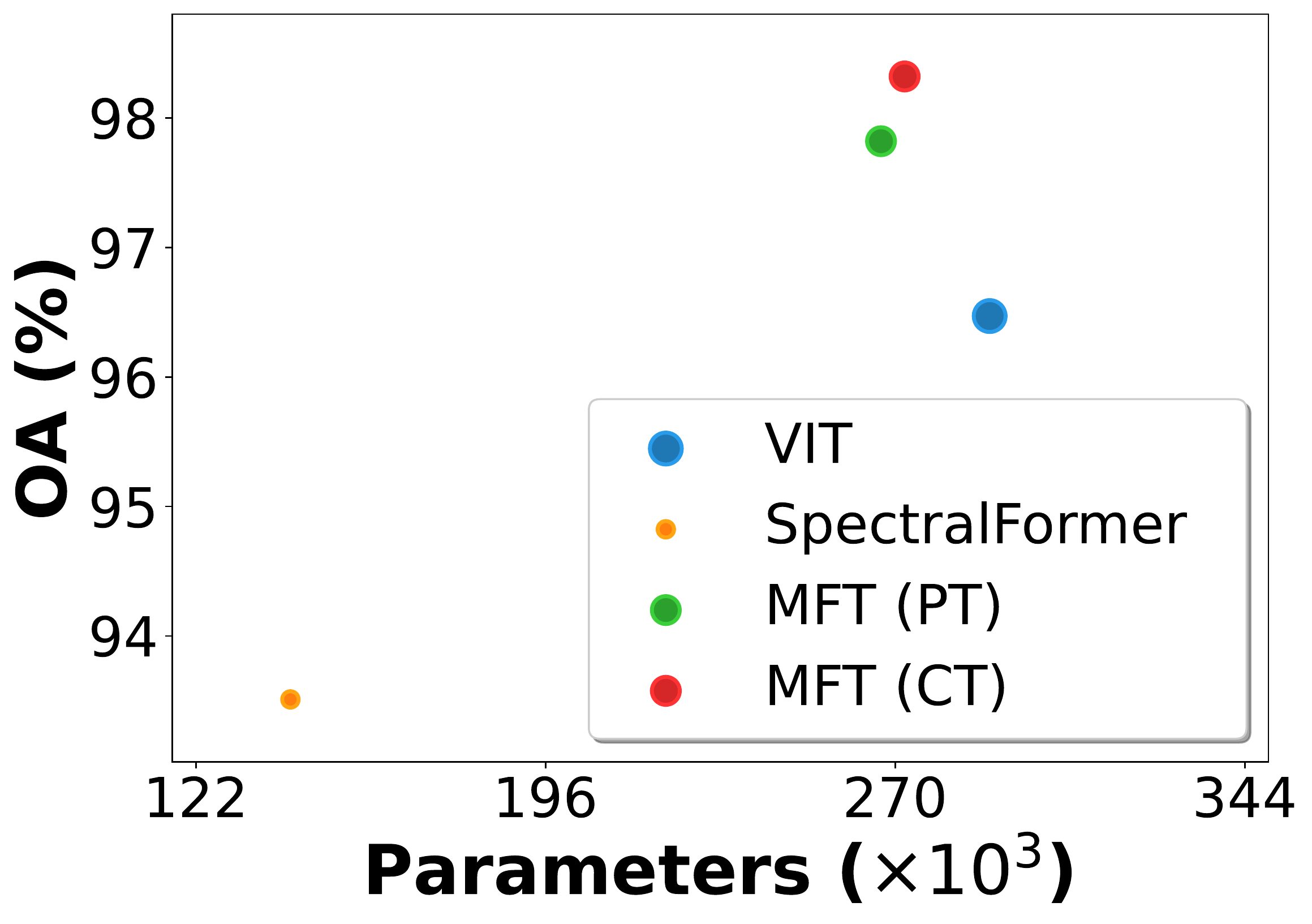}
		\centering
		\caption{UT (H+L)}
		% \label{Fig1E}
	\end{subfigure}
	\begin{subfigure}{0.30\textwidth}
		\includegraphics[width=0.99\textwidth]{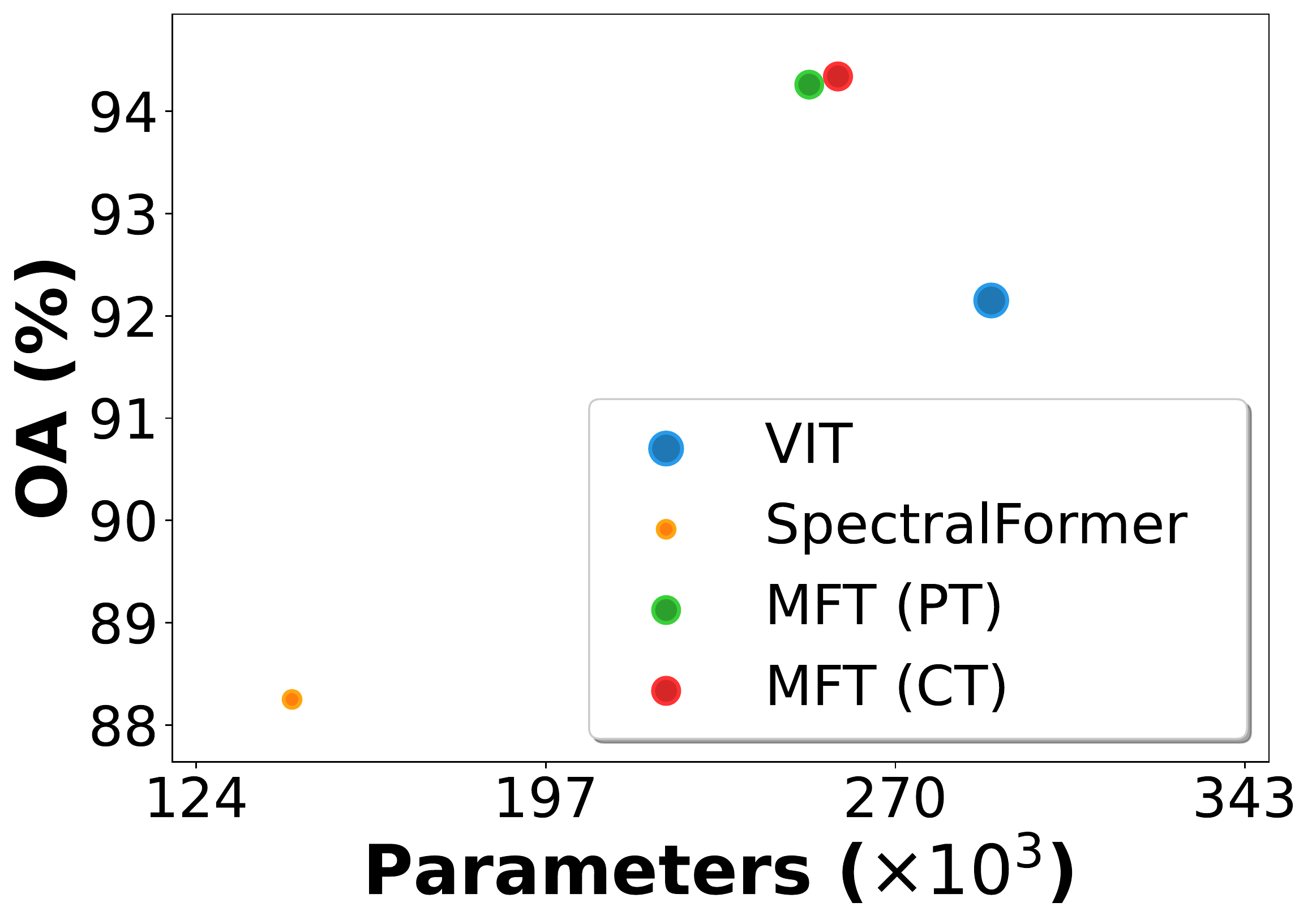}
		\centering
		\caption{MUUFL (H+L)}
		% \label{Fig1E}
	\end{subfigure}
	\begin{subfigure}{0.30\textwidth}
		\includegraphics[width=0.99\textwidth]{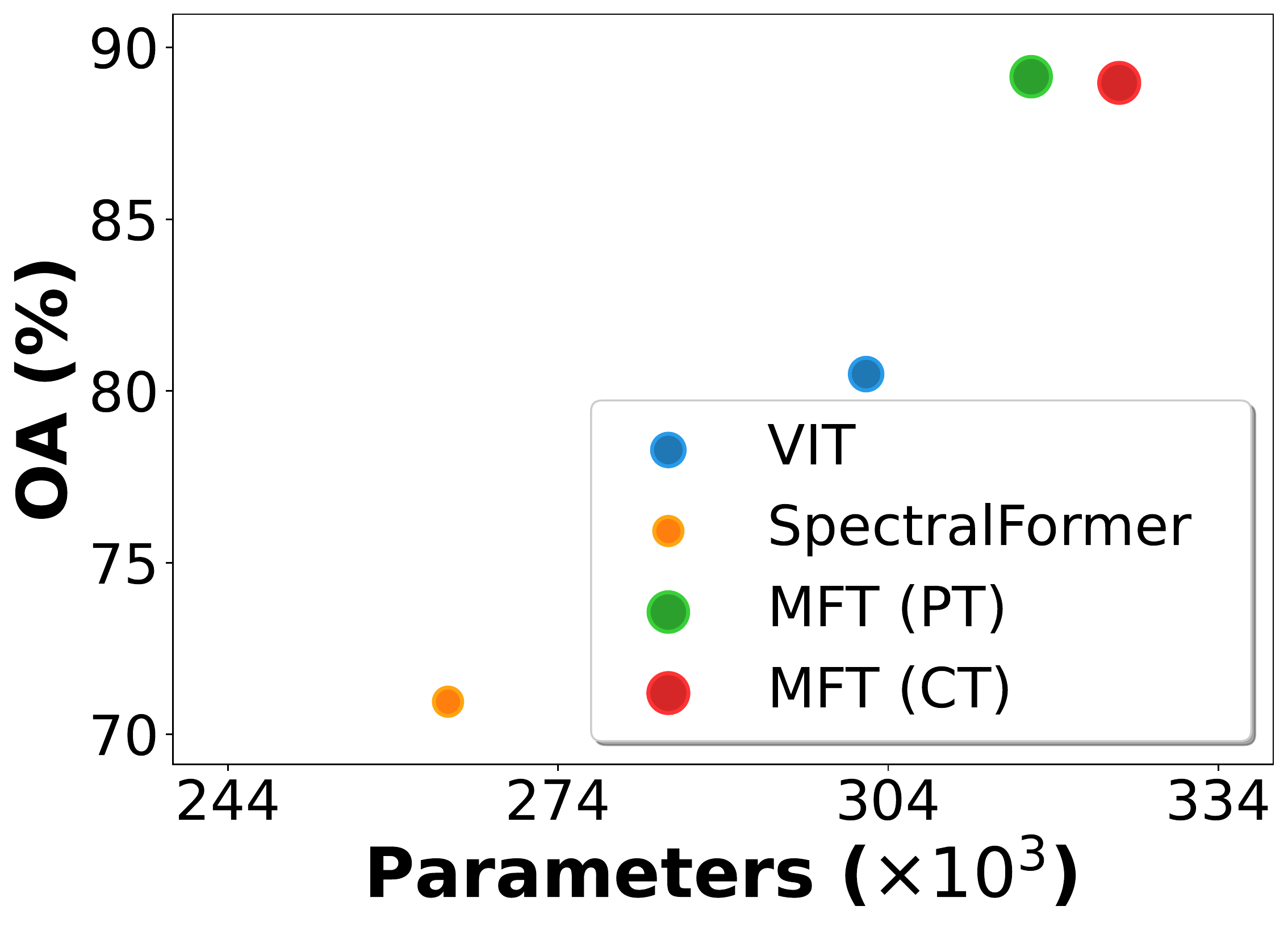}
		\centering
		\caption{UH (H+M)} 
		% \label{Fig1D}
	\end{subfigure}
	\begin{subfigure}{0.30\textwidth}
		\includegraphics[width=0.99\textwidth]{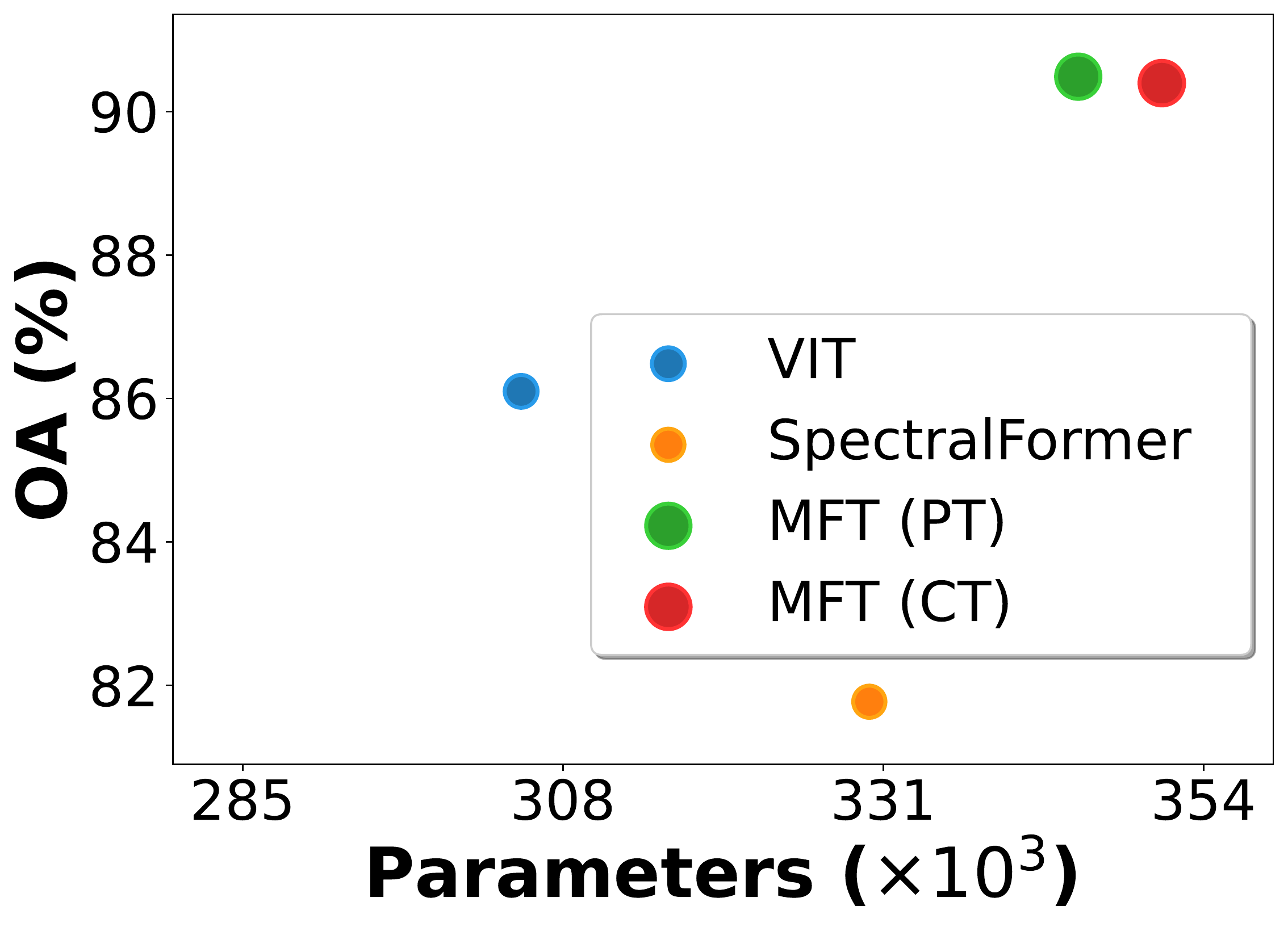}
		\centering
		\caption{Augsburg (H+SAR)}
		% \label{Fig1E}
	\end{subfigure}
	\begin{subfigure}{0.30\textwidth}
		\includegraphics[width=0.99\textwidth]{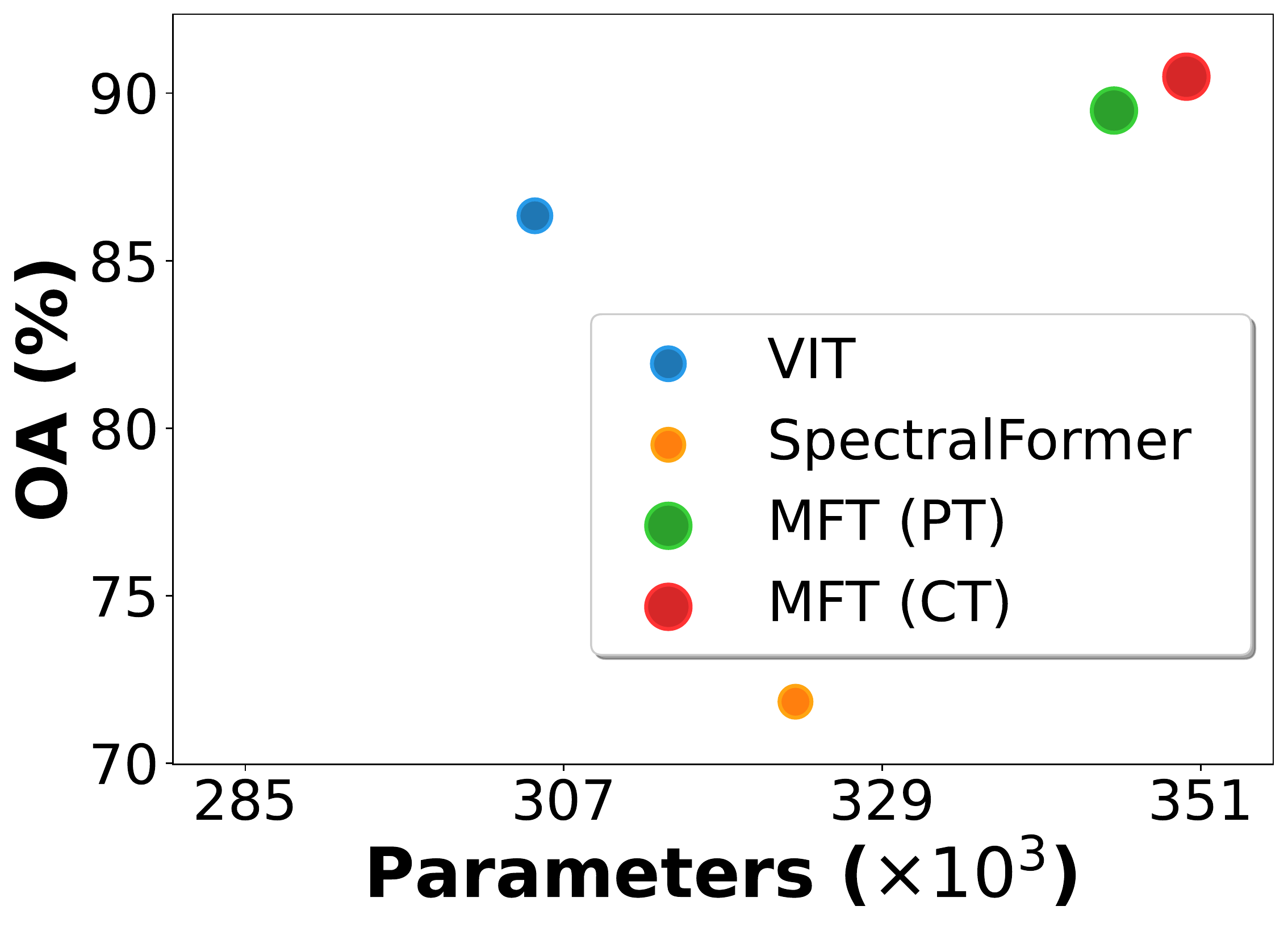}
		\centering
		\caption{Augsburg (H+DSM)}
		% \label{Fig1E}
	\end{subfigure}
\caption{Comparison of the transformer networks in terms of overall classification accuracy, network parameters, and FLOPs, indicated by radii of circles taken from (a) UH (HSI+LiDAR) (b) UT (HSI+LiDAR) (c) MUUFL (HSI+LiDAR) (d) UH (HSI+MS) (e) Augsburg (HSI+SAR) and (f) Augsburg (HSI+DSM) datasets.}
\label{fig:complexity}
\end{figure*}

%%%%%%%%%%%%%%%%%%%%%%%%%%%%%%%%%%%%%%%%%%%%%%%%%

\subsection{Model Stability Analysis}

1) \textbf{Ablation Study:} Regardless of the source of multimodal data, the dimension to tokenize plays a crucial role in the final classification performance and should be fine-tuned in addition to learning parameters and hyper-parameters. Therefore, it is essential to compare the model's performance with respect to pixel tokenization and channel tokenization. Hence, we investigate the classification performance of both variations on all four datasets, along with their respective combined multimodal data. Table \ref{tab:Ablation} shows the comparison of the classification accuracies of both methods in terms of OA, AA, and $\kappa$ for all the datasets. The optimal values are marked in bold. Overall, the performances of both methods are quite similar, but channel tokenization narrowly exceeds pixel tokenization. The reason may be that the channels represent the features better than pixels after convolution operations.

2) \textbf{Impact of Patch Size:} \textcolor{black}{The performance of the proposed MFT network is also influenced by the spatial size of the input HSI patches. As shown in Fig.~\ref{fig:aa_oa_kappa_patch}, the best and  more stable AAs, OAs and Kappa's are achieved by the proposed model using a
window of size ($11\times{11}$) for all the experimentally evaluated datasets. In all these experiments, disjointed train/test samples are utilized as explained in Section~\ref{subsec:hsi_data}. It can also be observed from the figures that, even the higher values in terms of AAs, OAs and Kappa's can be achieved over the UT (H+L), MUUFL (H+L), and Augsburg (H+S) datasets using a patch size of ($9\times{9}$), but the performance is not satisfactory for the UH (H+L) multi-modalities. The results for patches of size $13\times{13}$ and $15\times{15}$ shows that performance can significantly be reduced when increasing the patch size. It is worth mentioning that increasing the spatial window size will lead to an increase in redundancy. Hence we have considered $11\times{11}$ HSI patches for the entire experimental evaluation.}

3) \textbf{Hyperparameter Sensitivity Analysis:} The proposed MFT is not only effective but also fairly efficient in terms of computational complexity. The parameters and computations of the proposed MFT against other transformer networks are compared in Fig. \ref{fig:complexity}(a)-(f). They illustrate the overall accuracies, parameters, and computations (FLOPs) for UH (HSI+LiDAR), UT (HSI+LiDAR), MUUFL (HSI+LiDAR), UH (HSI+MS), Augsburg (HSI+SAR), and Augsburg (HSI+DSM) datasets, respectively. The radii of circles indicate the computations (FLOPs). The efficiency of the proposed MFT is quite evident in Trento (HSI+LiDAR) and MUUFL (HSI+LiDAR), where it uses fewer parameters and FLOPS than ViT (the model with second-best accuracies). In other cases, though, the parameters and FLOPS of the proposed MFT are higher than others, the increase in performance makes up for it. As we can see with Houston (H + MS), the proposed MFT exhibits an impressive $8.66\%$ increase in OA than the next best model, i.e., ViT. Such a massive gain in classification accuracy makes the parameter trade-off justified.

\section{Conclusions}\label{sec:conc}
\textcolor{black}{In this paper, we propose a new multimodal fusion transformer network based on a multihead cross patch attention (\texttt{mCrossPA}) module to fuse HSI and other sources of multimodal data for land cover classification. Instead of using conventional {feature} fusion {techniques}, other sources of multimodal data are used as an external classification (\texttt{CLS}) token that is incorporated into the \texttt{mCrossPA} block in the transformer encoder, which helps learning long-range feature dependencies. As the \texttt{CLS} token captures information that is complementary to that of the HSI data, it also helps in achieving a better generalization of the model and improves classification accuracy. The \texttt{CLS} token can be generated from multiple sources of easily available data, such as MSIs, SAR, DSM, and LiDAR. As a result, the model's performance is significantly increased with minimal effort.}

\textcolor{black}{Fusing multimodal data (such as HSIs, MSIs, SAR, DSMs, and Lidar) using recently developed transformers is challenging and requires significant effort. However, the extensive experimental results presented in this work confirm that the proposed MFT can successfully fuse multimodal data using transformers, which are inherently better at classification tasks than classical convolutional models. The proposed MFT performs better than all the other tested models with all the considered datasets. Therefore, we believe that the proposed MFT exhibits the capacity to perform multimodal fusion tasks easily in remote sensing and Earth Observation due to its superior fusion capabilities.}

\bibliographystyle{ieeetr}

\bibliography{reference}

\end{document}